%% file: arxiv.tex
\documentclass{article}

\usepackage[main, final]{neurips_2026}

\usepackage[utf8]{inputenc} 
\usepackage[T1]{fontenc}    
\usepackage{hyperref}       
\usepackage{url}            
\usepackage{booktabs}       
\usepackage{amsfonts}       
\usepackage{nicefrac}       
\usepackage{microtype}      
\usepackage{xcolor}         
\usepackage{amssymb}
\usepackage{amsmath}
\usepackage{amsthm}
\usepackage{graphicx}
\usepackage{wrapfig}
\usepackage{pgfplots}
\usepgfplotslibrary{groupplots}
\pgfplotsset{compat=1.18}

\DeclareMathOperator{\E}{\mathbb{E}}

\newcommand{\Mx}[1]{\mathbf{#1}}
\newcommand{\A}{\Mx{A}}
\newcommand{\I}{\Mx{I}}
\renewcommand{\S}{\Mx{S}}
\newcommand{\U}{\Mx{U}}
\newcommand{\V}{\Mx{V}}

\renewcommand{\H}{\Mx{H}}

\newcommand{\Vc}[1]{\mathbf{#1}}
\newcommand{\x}{\Vc{x}}
\renewcommand{\b}{\Vc{b}}
\newcommand{\s}{\Vc{s}}
\renewcommand{\u}{\Vc{u}}
\renewcommand{\v}{\Vc{v}}
\newcommand{\z}{\Vc{z}}
\newcommand{\w}{\Vc{w}}
\newcommand{\e}{\Vc{e}}

\newcommand{\X}{\Mx{X}}

\newtheorem{assumption}{Assumption}
\newtheorem{theorem}{Theorem}[section]
\newtheorem{corollary}{Corollary}[theorem]
\newtheorem{proposition}{Proposition}[theorem]
\newtheorem{lemma}[theorem]{Lemma}

\usepackage{pgfplots}
\pgfplotsset{compat=1.18}
\title{Debiasing Sketched Ridge Regression: A Functional Estimation Perspective}

\author{%
Yucong Liu \\
  School of Mathematics\\
  Georgia Institute of Technology\\
  Atlanta, GA 30332 \\
  \texttt{yucongliu@gatech.edu} \\
\And
Florian Sch\"afer \\
  Courant Institute of Mathematical Sciences\\
  New York University\\
  New York, NY 10012 \\
  \texttt{florian.schaefer@nyu.edu} \\
}

\begin{document}

\maketitle

\begin{abstract}
We study sketched ridge regression through the lens of functional estimation. The sketched solution is a plug-in estimator of a nonlinear function of a second-moment/covariance matrix.
Within this framework, we analyze two classical bias-reduction principles---iterative bootstrap and linear aggregation across multiple estimators.
We show that these resampling-based procedures cancel low-order bias terms and yield higher-order bias decay in the sketch size without increasing the order of the computational complexity. 
Concretely, let $\A\in\mathbb{R}^{n\times r}$ have full column rank $r$ and let the sketch size be $s$.
For Gaussian sketching, a $k$-step iterative bootstrap estimator achieves bias of order $(\sqrt{r/s})^{k+2}$.
For linear aggregation over $m$ plug-in estimators with different sketch sizes, we obtain bias of order $(\sqrt{r/s})^{m+1}$ for Gaussian sketching and corresponding result for common discrete sketches, including uniform and leverage score sampling.
\end{abstract}

\section{Introduction}
Solving linear systems $\A \x = \b$ and related linear problems, including regression—particularly ridge regression—is fundamental to a wide range of scientific and engineering applications.
Randomized Numerical Linear Algebra (RNLA) \citep{10.1561/2200000035, woodruff2014, drineas2016randnla, kannan2017, martinsson2020, murray2023randomized} provides an alternative approach via multiplication with a random matrix $\S$. Such methods enable randomized dimension reduction and consequently reduce computational cost. At the same time, this randomness incurs additional error and bias as the price paid for computational efficiency.

More precisely, a standard approach in sketching is the following: for a tall matrix $\A \in \mathbb{R}^{n \times r}$, one reduces the computational cost of ridge regression by replacing the original problem \mbox{$\min_{\x \in \mathbb{R}^r} \|\A\x - \b\|_2^2 + \gamma\|\x\|_2^2$} with the sketched problem \mbox{$\min_{\x \in \mathbb{R}^r} \|\S\A\x - \S\b\|_2^2 + \gamma\|\x\|_2^2$}. This random sketch matrix $\S \in \mathbb{R}^{s \times n}$ projects the data into a lower-dimensional space and reduces the problem size from $n$ to $s$. While this compression can substantially improve computational efficiency, it also introduces approximation error. Controlling this error is therefore essential for ensuring that sketching methods are both efficient and robust.

This work studies the bias of sketched solution of ridge regression via a functional estimation framework. In particular, the sketched solution can be viewed as a plug-in estimator of a certain function of the covariance matrix. We provide the bias bound for the sketched solution and show the bias reduction phenomenon by iterative bootstrap method and linear aggregation method. Both methods cancel out low-order bias term and achieve a high-order bias with respect to the sketch size. Moreover, with a fixed iteration number, both methods have same computational complexity as one sketched solution. 

\subsection{Motivation}
Given the optimal solution $\x^*$ and a sketched estimator $\hat \x$ of a ridge regression problem, the bias $\E \hat \x - \x^* $ is a persistent distortion in the estimator. While variance measures the spread of $\hat{\x}$ around its mean, the bias captures whether this mean is systematically shifted away from the true solution $\x^*$. Therefore, we aim to understand the bias with respect to sketch size.


Existing theoretical results provide only high-probability error bounds, valid when the sketch size is sufficiently large as a function of the probability tolerance $\delta$ and the error tolerance $\varepsilon$. More precisely, \citet{Wang2018} showed that 
\[
\left(\|\A\hat \x - \b\|_2^2 + \gamma\|\hat \x\|_2^2\right) - \left(\|\A\x^* - \b\|_2^2 + \gamma\|\x^*\|_2^2 \right) \le \varepsilon C \left(\|\A\x^* - \b\|_2^2 + \gamma\|\x^*\|_2^2 \right),
\]
with probability at least $1-\delta$, provided that the sketch size exceeds $s(\varepsilon,\delta)$. Here, $C$ is a constant depending on the matrix $\A$ and the parameter $\gamma$. However, this requirement on the sketch size $s$ makes it technically difficult to study the bias $\E\hat \x -\x^*$ directly as a function of $s$. In particular, such high-probability bounds do not yield an explicit bias decay rate of the form $\| \E\hat \x-\x^*\| = O(s^{-\alpha})$ for some $\alpha>0$. 

To overcome this technical difficulty, we develop an alternative perspective on the sketched solution by viewing it as a plug-in estimator for a functional estimation problem. Within this framework, we establish a bias decay rate of order $O(s^{-1})$. Moreover, both the iterative bootstrap and the linear aggregation method achieve faster bias decay rates while maintaining the same time complexity.

Another advantage of these two resampling-based iterative methods is that they offer an alternative mechanism for bias reduction through additional iterations. Although increasing the sketch size can also reduce the bias, doing so requires more memory and may be impractical in large-scale or distributed settings. In particular, under distributed computation, the memory available at the central server is often limited, so it may not be feasible to construct or process an overly large sketch in a single step. In contrast, iterative resampling allows us to work with sketches of moderate size while progressively reducing the bias across iterations. Therefore, repeating computations with a reasonable sketch size provides a practical and memory-efficient strategy for improving statistical accuracy.

Moreover, the matrix $(\A^\intercal \A)^{-1}$ plays a central role in ordinary least squares and its statistical analysis. When it is approximated by the sketch-based quantity $[(\S\A)^\intercal(\S\A)]^{-1}$, one encounters an inversion bias, since in general $\E\left[\left((\S\A)^\intercal(\S\A)\right)^{-1}\right] \neq (\A^\intercal \A)^{-1}$.
Our work focuses on the ridge regression setting and provides an analogous perspective for the regularized inversion $(\A^\intercal \A + \gamma \I)^{-1}$.

\subsection{Related work}
\citet{Wang2018} provided a comprehensive study of sketched ridge regression, proposing a model averaging approach and establishing several statistical guarantees. Their method averages multiple sketched solutions constructed with the same sketch size. While this averaging strategy can effectively reduce variance, it does not affect the bias and therefore does not correct the systematic distortion introduced by sketching.  \cite{dobriban2019asymptotics, Liu2020Ridge} analyzed the asymptotic properties of sketched estimators. Conditional central limit theorem has been studied by~\citet{Ahfock2021}. bootstrap-based methods for error estimation in randomized matrix multiplication and sketched SVD were further developed in \cite{Lopes2019, Lopes2020}.
Bias has been studied from related perspectives. \citet{derezinski2017unbiased, Derezinski2022} developed unbiased estimation methods for linear regression under volume sampling and random-design settings, while \citet{derezinski2021sparse, niu2025fundamental} investigated the inversion bias of $(\A^{\intercal}\A)^{-1}$ via a $(\varepsilon, \delta)$-unbiased estimator.

Functional estimation is a statistical problem with a long history. The question is how to estimate $f(\theta)$ for some function $f$ and parameter $\theta$ of a high-dimensional or even infinite-dimensional model~\citep{Donoho1991, Birge1995, Laurent1996, Lepski1999, Robins2008}.  Recently, \cite{Jiao2020} studied the estimation of one dimensional smooth function $f(\theta)$ based on binomial observations. They summarized three different methods: bootstrap, linear aggregation via several estimators and using Taylor expansion. Bootstrap, introduced by~\citet{efron1979}, then generalized by~\citet{MR995110}, is a classic resample technique, aiming to iteratively reduce bias. The linear aggregation method is a general version of Jackknife. The original Jackknife was initially introduced by~\cite{MR32176,MR81040} and \cite{tukey1958bias}. This method uses delete-one-sample estimator, inspiring using several different estimators to cancel bias.

Our work is inspired by recent developments about estimation for functionals of covariance operators. This line of research originated with a sharp characterization of covariance operator estimation in high-dimensional and infinite-dimensional settings~\citep{Koltchinskii_Lounici}. These techniques were subsequently extended to the estimation of bilinear forms of an unknown covariance operator~\citep{Koltchinskii_Lounici_bilinear} and linear functionals of its eigenvectors~\citep{Koltchinskii_Nickl}. More recently, the estimation of real-valued Hölder-smooth functionals of covariance operators has been systematically studied \citep{Koltchinskii2018, Koltchinskii2021, Koltchinskii_Zhilova, Koltchinskii2025}. In addition, \citet{koltchinskii2026trace} investigated the trace functionals of covariance operators, an important class of orthogonally invariant functionals.

\subsection{Notation}
\paragraph{Matrices and vectors.}
We denote the cone of $r \times r$ symmetric positive semi-definite matrices as $\mathbb{S}_+^r$ and the space of positive definite matrices as $\mathbb{S}_{++}^r$.
For a matrix $\X \in \mathbb{S}_+^r$, $\|\X\|$ denotes its operator norm. For a vector $\Vc{v} = (v_1, v_2, \cdots, v_n)\in\mathbb{R}^n$, we use $\|\Vc{v}\|_2 =(\sum_{i=1}^n v_i^2)^{1/2}$ for the Euclidean norm and $\|\Vc{v}\|_{\infty} = \max_{1\le i\le n}|v_i|$ for the infinity norm.
We write $\e_i$ for the $i$-th standard basis vector with the $i$-th entry equal to 1 and all others equal to 0.
For $\x, \Vc{y} \in \mathbb{R}^n$, $\langle \x, \Vc{y} \rangle = \x^{\intercal} \Vc{y}$. For $\A, \Mx B \in \mathbb{S}_+^r$, $\langle \A, \Mx B \rangle = \text{tr}(\A^\intercal \Mx B)$.
\paragraph{Asymptotic notation.}
For nonnegative $A,B$, we write $A \lesssim B$ if $A \le CB$ for a constant $C>0$.
If the constant $C$ depends on a parameter $m$, we write $A \lesssim_m B$.
We write $A \gtrsim B$ if $B \lesssim A$, and $A \asymp B$ if both $A \lesssim B$ and $A \gtrsim B$.
\paragraph{Orlicz norms.}
Let $\psi: \mathbb{R} \to \mathbb{R}_{+}$ be convex with $\psi(0)=0$ and increasing on $\mathbb{R}_+$. For a random variable $\xi$, we define its Orlicz norm with respect to $\psi$ as
\[
\|\xi\|_{\psi} = \inf \left\{c > 0 : \E \psi \left( \frac{|\xi|}{c} \right) \le 1 \right\}.
\]
In particular, if $\psi(u)=|u|^p$ for $p\ge 1$, then $\|\xi\|_{\psi}=\|\xi\|_{L_p}$.
Denote $\psi_{\alpha}(u) = e^{u^\alpha} - 1$, $\alpha \ge 1$. Then the $\psi_\alpha$ describes various type of decay of the tails of random variables. For example, $\|\xi\|_{\psi_2}$ is the sub-Gaussian norm. If $\xi$ is a random vector, $\|\xi\|_{\psi_2} := \sup_{u:\|u\|_2 = 1} \|\langle \xi, u \rangle\|_{\psi_2}$

\subsection{Outline and contributions}
\begin{enumerate}
\item We introduce the sketched ridge regression in Section~\ref{sec:sketch}. We provide a new insight of sketched ridge regression by formulating it as a functional estimation problem in Section~\ref{subsec:plugin}. In particular, both the classic and Hessian sketch solutions can be viewed as plug-in estimators of nonlinear functions of a second-moment (covariance) matrix. Under this framework, we show a bias bound in Section~\ref{subsec:bias_bound}.

\item
We study two principled debiasing strategies: iterative bootstrap in Section~\ref{sec:bootstrap} and linear aggregation method in Section~\ref{sec:jackknife}.
These procedures cancel low-order bias terms.
At the same time, the overall computational complexity remains of the same order as a single sketched solve.
We show that, in the Gaussian case, a $k$-step bootstrap-corrected estimator achieves bias of order $(\sqrt{r/s})^{k+2}$. For linear aggregation over $m$ plug-in estimators, we obtain bias of order $(\sqrt{r/s})^{m+1}$ even beyond the Gaussian setting.

\item
We complement the theoretical analysis with experiments on three real-world datasets, evaluating bias reduction, relative objective gaps, and runtime across sketching schemes.  Experimental results, implementation details, and proofs are provided in the appendix.

\end{enumerate}

\section{Sketching as a Functional Estimation Problem}\label{sec:sketch}
\subsection{Sketched Ridge Regression}
Consider a ridge regression problem, i.e. a least square problem with $\ell_2$ regularization term:
\begin{equation}\label{eq:ridge}
\min_{\x \in \mathbb{R}^r} \|\A\x - \b\|_2^2 + \gamma\|\x\|_2^2,
\end{equation}
where $\A \in \mathbb{R}^{n \times r}$ with full column rank and $\b \in \mathbb{R}^n$, $\gamma > 0$. 

The optimal solution of the ridge regression problem~\eqref{eq:ridge} is given by
\begin{equation}\label{eq:true_solution}
\x^* = (\A^\intercal \A + \gamma \I)^{-1} \A^\intercal \b
\end{equation}

Given a sketch matrix $\S \in \mathbb{R}^{s \times n}$, the corresponding sketching problem is 
\begin{equation}\label{eq:sketch_problem}
\min_{\x \in \mathbb{R}^r} \|\S\A\x - \S\b\|_2^2 + \gamma\|\x\|_2^2,
\end{equation}
with classic sketch solution
\begin{equation}\label{eq:classic_sketch}
    \hat \x_c = (\A^\intercal \S^\intercal \S \A + \gamma \I)^{-1} \A^\intercal \S^\intercal \S \b.
\end{equation}
There is an alternative sketch solution called Hessian sketch solution, defined as
\begin{equation}\label{eq:hessian_sketch}
    \hat \x_h = (\A^\intercal \S^\intercal \S \A + \gamma \I)^{-1} \A^\intercal  \b.
\end{equation}
Sketch solutions achieve faster computation with time complexity $O(sr^2) + T_s(\A) + T_s(\b)$ for classic sketch solution~\eqref{eq:classic_sketch} and $O(sr^2 + nr) + T_s(\A)$ for Hessian sketch solution~\eqref{eq:hessian_sketch}, compared with $O(nr^2)$ for the optimal solution~\eqref{eq:true_solution}. Here $T_s(\A)$ denotes the time complexity to form $\S \A$ and $T_s(\b)$ denotes the time complexity to form $\S \b$ as in Equation~\eqref{eq:sketch_problem}. One main difference between the classic solution~\eqref{eq:classic_sketch} and the Hessian sketch solution~\eqref{eq:hessian_sketch} is that the Hessian sketch only sketches the Hessian matrix, with no effect on the target vector $\b$. 
This inspires an iterative Hessian sketch method to iteratively decrease the residual \citep{Pilanci2016, wang2017sketching}.

\subsection{Sketched solution is a plug-in estimator}\label{subsec:plugin}
We now re-formulate the sketch solution as an estimation problem. 

Consider the compact SVD decomposition of $\A = \U \Mx{\Lambda} \V^\intercal$ with $\U \in \mathbb{R}^{n \times r}$, $\Mx{\Lambda}, \V \in \mathbb{R}^{r \times r}$. Denote the singular value of $\A$ as $\sigma_1 \ge \sigma_2 \ge \cdots \sigma_r > 0$. To construct a random sketch matrix $\S \in \mathbb{R}^{s \times n}$, we view its rows $\s_1^\intercal,\ldots,\s_s^\intercal\in\mathbb{R}^n$ as i.i.d. samples from a distribution $P$ on $\mathbb{R}^n$. 
Define the compressed sample $\z_i = \sqrt{s} \U^\intercal \s_i \in \mathbb{R}^r$ for each index $i$.
One can directly get
\[
\U^\intercal \S^\intercal \S\U = \frac{1}{s} \sum_{i=1}^s \z_i \z_i^\intercal.
\]
We emphasize that the compressed sample provides a dimension reduction from $n$ to $r$, which captures the intrinsic dimension. The sample $\z_i$ is only used for analytical purpose, we do not explicitly compute it to get sketch solution or use it in the iterative debiasing steps.

Equivalently, we can view sketching as generating i.i.d.
``compressed'' random variables $\z_i \in \mathbb{R}^r$ from a random vector $\z$.
We call $\z$ an isotropic random vector if its second-moment matrix satisfies
\[
\Mx{\Sigma} := \E \z\z^\intercal = \I.
\]
We say $\S$ is an isotropic sketch if it satisfies the above condition.
An  estimator of $\Mx{\Sigma}$ is the sample second-moment matrix
\[
\hat{\Mx{\Sigma}}_s := \frac{1}{s} \sum_{i=1}^s \z_i \z_i^\intercal = \U^\intercal \S^\intercal \S\U.
\]
If $\E\z = 0$, then $\Mx{\Sigma}$ coincides with the covariance matrix and $\hat{\Mx{\Sigma}}_s$ is called the sample covariance matrix.
It is clear that $\hat{\Mx{\Sigma}}_s$ is an unbiased estimator of $\Mx{\Sigma}$, namely, $\E\hat{\Mx{\Sigma}}_s = \Mx{\Sigma}$.

We begin with the following theorem, which quantifies the closeness of $\hat{\Mx\Sigma}_s$ to $\Mx\Sigma$.
\begin{theorem}\label{thm:operator_norm}
    If $\S$ is an isotropic sketch matrix with sketch size $s$, then for every $p \ge 1$,
    \begin{equation}
        \E^{1/p} \|\hat{\Mx{\Sigma}}_s - \Mx\Sigma\|^p \lesssim_p \kappa \left(\sqrt{\frac{r}{s}} \vee \frac{r}{s} \right)
    \end{equation}
    Here $\kappa = \|\z\|^2_{\psi_2}$ depends on the distribution.
\end{theorem}

We first connect the matrix $\Mx\Sigma$ and its estimator $\hat{\Mx\Sigma}_s$ with the optimal solution $x^*$ and the Hessian sketch solution $\hat x_h$. Define a vector valued function from $\mathbb{S}_+^r$ to $\mathbb{R}^r$ with $\gamma > 0$:
\begin{equation}
f_h(\X) = (\V\Mx{\Lambda}^\intercal \X \Mx{\Lambda} \V^\intercal + \gamma \I_r)^{-1} \A^\intercal \b .
\end{equation}
Note that under the isotropic condition, we have $\Mx\Sigma = \E \z \z^\intercal = \I_r$. Consequently, by Equation~\ref{eq:true_solution} and the SVD of $\A$, the optimal solution $x^*$ can be expressed as a function of the second-order moment matrix $\Mx\Sigma$:
\[
\x^* = (\V\Mx{\Lambda}^\intercal \U^\intercal \U \Mx{\Lambda} \V^\intercal + \gamma \I_r)^{-1} \A^\intercal \b = f_h(\I_r) = f_h(\Mx\Sigma) = f_h(\E\z\z^\intercal).
\]
Moreover, the Hessian sketch solution $\hat x_h$ can be expressed as the same function of the $\hat{\Mx\Sigma}_s$:
\[
\hat \x_h 
= (\A^\intercal \S^\intercal \S \A + \gamma \I_r)^{-1} \A^\intercal  \b 
= \left[\V\Mx{\Lambda}^\intercal (\U^\intercal \S^\intercal) (\S \U)  \Mx{\Lambda} \V^\intercal + \gamma \I\right]^{-1} \A^\intercal  \b 
= f_h(\hat{\Mx{\Sigma}}_s).
\]
As a result, the Hessian sketch solution is a plug-in estimator $f_h(\hat{\Mx\Sigma}_s)$ of $f_h(\Mx\Sigma)$ given unbiased estimator $\hat{\Mx{\Sigma}}_s$ of $\Mx{\Sigma}$.

Similarly, the classic sketch solution can also be formulated as a functional estimation problem involving an additional term $\A^\intercal \S^\intercal \S \b$. Consider the SVD of the augmented matrix $[\A\ \b] = \U_+ \Mx{\Lambda}_+\V_+^\intercal$. Here $\U_+ \in \mathbb{R}^{n \times r_+}$, $\Mx{\Lambda}_+ \in \mathbb{R}^{r_+ \times r_+}, \V_+ \in \mathbb{R}^{(r+1) \times r_+}$, where $r_+ = r$ if the vector $\b$ lies in the column space of $\A$, and $r_+=r+1$ otherwise. 

Define the compressed samples $\z_{+,i}:=\sqrt{s}\U_+^\intercal\s_i\in\mathbb{R}^{r_+}$. For the classic sketch, we require isotropy on this augmented space, namely,
\[
\Mx{\Sigma}_+ = \E\z_{+,i}\z_{+,i}^\intercal = \I_{r_+}.
\]
The corresponding sample second-moment matrix is
\[
\hat{\Mx{\Sigma}}_{+,s}=\frac{1}{s}\sum_{i=1}^s
\z_{+,i}\z_{+,i}^\intercal = \U_+^\intercal\S^\intercal\S\U_+.
\]

To express the classic sketch solution, define the vector-valued function
$f_c:\mathbb{S}_+^{r_+}\to\mathbb{R}^r$ by
\[
f_c(\X)
= (\V_+\Mx{\Lambda}_+^\intercal \X \Mx{\Lambda}_+ \V_+^\intercal + \gamma \I)^{-1} \V_+\Mx{\Lambda}_+^\intercal \X \U^\intercal_+ \b.
\]
Since $\b$ lies in the column space of $\U_+$, we obtain a representation of the optimal solution:
\[
\x^*
= (\A^\intercal\A+\gamma\I_r)^{-1}\A^\intercal\b 
= f_c(\I_{r_+})
= f_c(\Mx{\Sigma}_+).
\]
Moreover, the classic sketch solution satisfies
\[
\hat{\x}_c
=
(\A^\intercal\S^\intercal\S\A+\gamma\I_r)^{-1}
\A^\intercal\S^\intercal\S\b
=
f_c(\hat{\Mx{\Sigma}}_{+,s}).
\]
Thus, the classic sketch solution is a plug-in estimator of
$f_c(\Mx{\Sigma}_+)=\x^*$, based on the unbiased estimator
$\hat{\Mx{\Sigma}}_{+,s}$ of $\Mx{\Sigma}_+$.

Note here, the isotropic condition is common in sketch problems. In Appendix~\ref{appx:notations}, we introduce uniform sampling, leverage score sampling and Gaussian scheme, all of them satisfy this condition.

For notational convenience, we henceforth suppress the subscript $+$ on the compressed samples and sample second-moment matrices. Accordingly, $\Mx\Sigma$ and $\hat{\Mx\Sigma}_s$ denote $r\times r$ matrices when used with $f_h$, and $r_+\times r_+$ matrices when used with $f_c$.

\subsection{Bias of plug-in estimator without tail probability}\label{subsec:bias_bound}
The plug-in estimator causes bias in functional estimation problem in general, such that 
\[
\E f(\hat{\Mx{\Sigma}}_s) \neq f({\Mx{\Sigma}}),
\] 
for functional $f:\mathbb{S}_+^r \to \mathbb{R}$. One well-known example is the log-determinant of the covariance matrix in a Gaussian model, for which the plug-in estimator has an explicit bias term~\citep{MR3332804}.
In this work, we aim to study the bias in sketched ridge regression, i.e the difference between approximated solution $\hat\x_c, \hat\x_h$ and the optimal solution  $\x^*$ under expectation:
\[
\E f_c(\hat{\Mx{\Sigma}}_s) - f_c({\Mx{\Sigma}}) \quad \text{and} \quad \E f_h(\hat{\Mx{\Sigma}}_s) - f_h({\Mx{\Sigma}}).
\]
This bias is intrinsic due to the non-linearity of $f_c$ and $f_h$. Unlike variance, which can often be reduced by repeating the same randomized experiment and averaging, bias cannot in general be removed simply by averaging multiple plug-in estimators. Given $k$ independent estimators $f_c(\hat{\Mx\Sigma}_{(i)})$ or $f_h(\hat{\Mx\Sigma}_{(i)})$, $1\le i \le k$, with same sample size, the bias remains unchanged after averaging:
\[
\E f_c(\hat{\Mx{\Sigma}}) - f_c({\Mx{\Sigma}}) = \E \frac{1}{k} \sum_{i=1}^k f_c(\hat{\Mx{\Sigma}}_{(i)}) - f_c({\Mx{\Sigma}}),  \quad \E f_h(\hat{\Mx{\Sigma}}) - f_h({\Mx{\Sigma}}) = \E \frac{1}{k} \sum_{i=1}^k f_h(\hat{\Mx{\Sigma}}_{(i)}) - f_h({\Mx{\Sigma}}).
\]
This limitation motivates a direct analysis of the bias itself. We state the corresponding bias bound below.

\begin{theorem}\label{thm:bias_bound}
    If $\S$ is an isotropic sketch matrix with sketch size $s$, then for a Hessian sketch solution, we have
    \[
    \left\|\E f_h(\hat{\Mx{\Sigma}}_s) - f_h({\Mx{\Sigma}}) \right\|_{\infty}
    \lesssim_{\sigma_1, \gamma, \|\b\|} \kappa^2 \left[\frac{r}{s} \vee \left(\frac{r}{s} \right)^{2} \right],
    \]
    Similarly, for the classic sketch solution, we have
    \[
    \left\| \E f_c(\hat{\Mx{\Sigma}}_s) - f_c({\Mx{\Sigma}}) \right\|_\infty 
    \lesssim_{\sigma_1, \gamma, \|\b\|} \kappa^2 \left[\frac{r_+}{s} \vee \left(\frac{r_+}{s} \right)^{2} \right].
    \]
    In particular, if $\S$ is a Gaussian sketch matrix, the classic sketch bias admits the sharper bound:
    \[
    \left\| \E f_c(\hat{\Mx{\Sigma}}_s) - f_c({\Mx{\Sigma}}) \right\|_\infty 
    \lesssim_{\sigma_1, \gamma, \|\b\|} \left[\frac{r}{s} \vee \left(\frac{r}{s} \right)^{2} \right].
    \]
    Here $\kappa = \|\z\|^2_{\psi_2}$ depends on the distribution. 
\end{theorem}

The sharper dependence on $r$ under Gaussian sketching arises from independence along orthogonal directions. Write
\[
\b=\U\U^\intercal\b+\b_\perp,
\qquad
\b_\perp:=(\I_n-\U\U^\intercal)\b.
\]
The classic sketch solution then decomposes as
\begin{align*}
\hat{\x}_c
=
(\A^\intercal\S^\intercal\S\A+\gamma\I_r)^{-1}
\A^\intercal\S^\intercal\S\U\U^\intercal\b
+
(\A^\intercal\S^\intercal\S\A+\gamma\I_r)^{-1}
\A^\intercal\S^\intercal\S\b_\perp.
\end{align*}
For a Gaussian sketch, $\S\A$ and $\S\b_\perp$ are independent, and $\S\b_\perp$ has mean zero. Then the second term has zero expectation. The first term is precisely the classic sketch solution with response $\U\U^\intercal\b$. Consequently, the bias coincides with that of the projected-response problem, whose analysis involves only $r$ dimensions. The orthogonal component $\b_\perp$ increase only variance without contributing to the bias. For general sketches, isotropy alone does not guarantee this cancellation, and our general bound uses the augmented dimension $r_+$.

When the sketch size $s$ satisfies $s \gtrsim r$, Theorem~\ref{thm:bias_bound} indicates a bias decay of rate $O(s^{-1})$. Both covariance estimation error bound and sketch bias bound rely on the sub-Gaussian norm, which indicates the tail decay. In the Gaussian case, $\kappa \asymp 1$ and in the sub-Gaussian case, $\kappa \lesssim 1$. It's known that if an isotropic random vector supported in a finite set has constant sub-gaussian norm, then the cardinality of the support set must be exponentially large; see~\citep{Vershynin2018}. Therefore, discrete sampling schemes necessarily have relatively large sub-Gaussian norms. In Appendix~\ref{appx:sketch_dist} we provide upper bounds for the uniform sketch and leverage score sampling. Therein, we also discuss when the upper bound is attained.

\section{Iterative Bootstrap}\label{sec:bootstrap}
We study the bootstrap in the Gaussian case. Let $\hat{\Mx{\Sigma}}$ be the estimator based on observations $\z_1, \dots, \z_s \sim \mathcal{N}(0, \Mx\Sigma), \Mx\Sigma \in \mathbb{S}_{++}^r$ and $f(\hat{\Mx{\Sigma}})$ be the plug-in estimator of $f(\Mx{\Sigma})$ for some functional $f$.
Then the first order bias can be described as 
\[
\E f(\hat{\Mx{\Sigma}}) - f(\Mx{\Sigma}) := (\mathcal{B}f)(\Mx{\Sigma}).
\]
The first order bias correction yields a new estimator
$
f(\hat{\Mx{\Sigma}}) - (\mathcal{B}f)(\hat{\Mx{\Sigma}}),
$
with a second order bias
\[
\E f(\hat{\Mx{\Sigma}}) - \E (\mathcal{B}f)(\hat{\Mx{\Sigma}}) - f(\Mx\Sigma) = (\mathcal{B}f)(\Mx\Sigma) - \E(\mathcal{B}f)(\hat{\Mx{\Sigma}}) = -(\mathcal{B}^2f)(\Mx{\Sigma}).
\]
Then the second order bias correction yields an estimator
\[
f(\hat{\Mx{\Sigma}}) - (\mathcal{B}f)(\hat{\Mx{\Sigma}}) + (\mathcal{B}^2f)(\hat{\Mx{\Sigma}}),
\]
and higher-order corrections can be obtained in a similar fashion:
\[
\sum_{j=0}^k(-1)^j\left(\mathcal{B}^j f\right)(\Mx\Sigma)=f(\Mx\Sigma)+\sum_{j=1}^k(-1)^j\left(\mathcal{B}^j f\right)(\Mx\Sigma).
\]

This procedure yields the following estimator
\begin{equation}\label{eq:bootstrap_def}
f_k(\hat{\Mx\Sigma}) = \sum_{i=1}^{k+1} (-1)^{i+1} \binom{k+1}{i} f(\hat{\Mx\Sigma}_{(i)})
\end{equation}
Here $\hat{\Mx\Sigma}_{(i)}$ is the covariance estimator based on an i.i.d. sample $\z_{1}^{(i-1)}, \ldots, \z_{s}^{(i-1)}$ drawn from $\mathcal{N}(0,\hat{\Mx\Sigma}_{(i-1)})$. We initialize the recursion with $\hat{\Mx\Sigma}_{(1)}=\hat{\Mx\Sigma}$, $\z_i^{(0)}=\z_i$.

In practice, the iterative bootstrap is implemented in an alternative way by repeatedly resampling from existing samples, i.e resample $s$ samples from $\z_{1}^{(i-1)}, \cdots, \z_s^{(i-1)}$ with replacement to construct $\z_{1}^{(i)}, \cdots, \z_s^{(i)}$. As mentioned, we don't calculate covariance estimator and plug into target function $f_c$ and $f_h$. One can resample from the observed sketch, i.e uniformly sampling $s$ rows with replacement from $\S$. This step has a computational cost of $O(s)$, which can be ignored in the total computational cost. Also the re-computing plug-in estimators via solving the linear system remain the same complexity as the original sketch solution. As a result, for a fixed iteration number, the iterative bootstrap will not increase the computational complexity.  

It's known that, if a function $f$ is $k$-th time continuously differentiable, its $k$-th order forward difference yields the following property: as $h \to 0$,
\[
\Delta_{h}^{k}f(x) = \sum_{i=0}^{k} (-1)^{k-i} \binom{k}{i} f(x+ih) = f^{(k)}h^k + o(h^k).
\]
Denote $\hat{\Mx\Sigma}_{(0)} = \Mx\Sigma$, the error of iterative bootstrap~\eqref{eq:bootstrap_def} has a similar structure:
\begin{align*}
&
f(\Mx\Sigma) - f_k(\hat{\Mx\Sigma}) =\sum_{i=0}^{k+1} (-1)^{i} \binom{k+1}{i} f(\hat{\Mx\Sigma}_{(i)}) \\
&= \sum_{i=0}^{k+1} (-1)^{i} \binom{k+1}{i} f\left(\hat{\Mx\Sigma}_{(0)} + \sum_{j=1}^{i}\left[\hat{\Mx\Sigma}_{(j)} - \hat{\Mx\Sigma}_{(j-1)}\right]\right). 
\end{align*}
\citet{Koltchinskii_Zhilova} developed a technique called random homotopy, to control the step $\hat{\Mx\Sigma}_{(j)} - \hat{\Mx\Sigma}_{(j-1)}$, which is the difference between one covariance and its estimator. Equipped with that, one can derive similar bias bound as the forward difference of continuously differentiable functions.

Let $f_{c,k}(\hat{\Mx\Sigma})$ be the $k$-step bootstrap estimator of $f_c(\Mx\Sigma)$ and $f_{h,k}(\hat{\Mx\Sigma})$ be the $k$-step bootstrap estimator of $f_h(\Mx\Sigma)$.
We now state our main bias reduction result for sketched ridge regression under Gaussian sketching. As discussed above, sketching methods produce plug-in estimators whose bias decays at rate of $1/s$. By exploiting higher-order smoothness of the underlying functionals and applying iterated bias correction, we show that one can substantially accelerate the decay of the bias with respect to the sketch dimension. In particular, when the sketching matrix is Gaussian, the proposed 
$k$-th order bias-corrected estimators achieve a bias of order $(\sqrt{r/s})^{k+2}$, improving upon the baseline $r/s$ rate. When $k = 0$, these estimators coincide with the classic sketch solution~\eqref{eq:classic_sketch} and Hessian sketch solution~\eqref{eq:hessian_sketch}, staying the same bias order.

\begin{theorem}\label{thm:bootstrap_bias}
Let $\S$ be a Gaussian sketch matrix and $s \ge Cr$ for some constant $C$, For the $k$-step parametric Gaussian bootstrap estimators and arbitrary $\b\in\mathbb{R}^n$, the Hessian sketch bias satisfies
\[
\left\| \E f_{h,k}(\hat{\Mx\Sigma}_s)-f_h(\Mx\Sigma) \right\|_\infty
\lesssim_{k,\sigma_1,\gamma,\|\b\|}
\left(\sqrt{\frac{r}{s}}\right)^{k+2}.
\]
Similarly, the classic sketch bias satisfies
\[
\left\|\E f_{c,k}(\hat{\Mx\Sigma}_s)-f_c(\Mx\Sigma)\right\|_\infty
\lesssim_{k,\sigma_1,\gamma,\|\b\|}
\left(\sqrt{\frac{r}{s}}\right)^{k+2}.
\]
\end{theorem}

While the preceding theorem establishes improved bias decay for the proposed bias-corrected estimators, controlling the bias alone is not sufficient to characterize statistical performance. In particular, a useful estimator must balance bias reduction with variance inflation induced by the correction procedure. We therefore complement the bias analysis with $L_p$-risk bounds, which capture both variance and bias in a unified manner. The following theorem shows that, for Gaussian sketching, each coordinate of the $k$-step bootstrap-corrected estimator admits an $L_p$ error bound with two components: a plug-in estimator variance term of order $1/\sqrt{s}$ and a bias term of order $(\sqrt{r/s})^{k+2}$.

\begin{theorem}\label{thm:bootstrap_lp}
Let $\S\in\mathbb{R}^{s\times n}$ be a Gaussian sketch
matrix, $s \ge Cr$ for some constant $C$ and consider the $k$-step parametric Gaussian bootstrap estimators.
For arbitrary $\b\in\mathbb{R}^n$, every $p\ge1$,
and every $1\le i\le r$,
the Hessian sketch estimator satisfies
\[
\left\|\left\langle
f_{h,k}(\hat{\Mx\Sigma}_s)-f_h(\Mx\Sigma),\e_i
\right\rangle\right\|_{L_p}
\lesssim_{k,\sigma_1,\gamma,\|\b\|,p}
\frac{1}{\sqrt{s}}
+
\left(\sqrt{\frac{r}{s}}\right)^{k+2}.
\]
For the classic sketch, we have
\[
\left\|\left\langle
f_{c,k}(\hat{\Mx\Sigma}_s)-f_c(\Mx\Sigma),\e_i
\right\rangle\right\|_{L_p}
\lesssim_{k,\sigma_1,\gamma,\|\b\|,p}
\frac{1}{\sqrt{s}}
+
\left(\sqrt{\frac{r_+}{s}}\right)^{k+2}.
\]
\end{theorem}

The distinction between the two risk bounds reflects the different ways in which the response enters the sketched estimators. For the Hessian sketch, the response appears only through $\A^\intercal\b$, and $\A^\intercal\b_\perp=0$. Thus, the orthogonal response component has no effect on the estimator. In contrast, the classic sketch contains the additional term
\[
(\A^\intercal\S^\intercal\S\A+\gamma\I_r)^{-1}
\A^\intercal\S^\intercal\S\b_\perp.
\]
Under Gaussian sketching, this term has zero expectation, but generally has nonzero variance. It therefore does not contribute to the bias, while its fluctuations must be accounted for in the $L_p$-risk analysis, including after bootstrap correction. To account for this additional component, we use the $r_+$-dimensional augmented representation in the classic sketch risk analysis, whereas the Hessian sketch involves only the $r$-dimensional representation. Nevertheless, the Gaussian bias bound retains the sharper dependence on $r$.

Asymptotically, when the model dimension grows with the sketch size as $r \asymp s^\alpha$ for $\alpha \in (1/2, 1)$, bias will dominant in the previous $L_p$ risk bound. To be precise, the $L_p$ risk bound is of order $r/s \asymp s^{\alpha - 1}$.
However, with sufficient iteration via bootstrap, one may achieve error of order $s^{-1/2} < s^{\alpha - 1}$. We state it as the following corollary.
\begin{corollary}
    If $\S \in \mathbb{R}^{s \times n}$ is chosen as a Gaussian sketch matrix and $s \ge Cr$ for some constant $C$ and $s^\alpha \asymp r$, for $\alpha \in (1/2, 1)$. Choose $k \ge \lceil 1/(1-\alpha) \rceil - 2$, then for every $1 \le i \le r$,
    $$
    \left\|\langle f_{h,k}(\hat{\Mx\Sigma}_s) - f_h(\Mx\Sigma), \e_i \rangle\right\|_{L_p} \lesssim_{k,\sigma_1, \gamma, \|\b\|,p} s^{-1/2}.
    $$
    and
$$
    \left\|\langle f_{c,k}(\hat{\Mx\Sigma}_s) - f_c(\Mx\Sigma), \e_i \rangle\right\|_{L_p} \lesssim_{k,\sigma_1, \gamma, \|\b\|,p} s^{-1/2}.
    $$
\end{corollary}

The next section develops debiasing strategy based on linear aggregation across multiple estimators, which cancels leading bias terms via an explicit choice of coefficients.

\section{Careful Bias Reduction via Linear Aggregation}\label{sec:jackknife}
In this section, we will introduce the bias reduction via linear aggregation over multiple plug-in estimators. Let $m \ge 2$ be the number of estimators and $s/c \le s_1 < \cdots < s_m = s$ for some constant $c > 1$. Then, the estimators for the classic sketch solution~\eqref{eq:classic_sketch} and Hessian sketch solution~\eqref{eq:hessian_sketch} are designed as 
\[
\hat{T}_{c,m} = \sum_{j=1}^m C_j f_c(\hat{\Mx{\Sigma}}_{s_j}),
\quad
\hat{T}_{h,m} = \sum_{j=1}^m C_j f_h(\hat{\Mx{\Sigma}}_{s_j}),
\]
where the coefficient $C_j$ satisfies the following linear system:
\begin{equation}\label{eq:coefficient}
    \sum_{j=1}^m C_j = 1, \qquad
    \sum_{j=1}^m \frac{C_j}{s_j^\ell} = 0,
\end{equation}
for all integer $1 \le \ell \le m-1$.
This linear system~\eqref{eq:coefficient} has explicit solution such that
\[
C_j = \prod_{i \neq j} \frac{s_j}{s_j-s_i}.
\]

In this study, we only require the samples used to construct each $\hat{\Mx \Sigma}_{s_j}$ to be i.i.d, which means that one may re-use same sample to construct $\hat{\Mx \Sigma}_{s_j}$ for different $j$. This helps control the total sample size. Even if sample reuse is not allowed, the total number of samples remains at most $O(ms)$.
Throughout the remainder of this work, we make the following assumption, which ensures that the aggregation coefficients do not blow up. This assumption can be easily achieved if the sample size is chosen as $s_j = \rho^{j-m} s$ for some fixed $\rho >1$. It is also worth emphasizing that, under this choice of $s_j$, computing $f_c(\hat{\Mx\Sigma}_{s_j})$ and $f_h(\hat{\Mx\Sigma}_{s_j})$ has the same order of computational complexity as computing $f_c(\hat{\Mx\Sigma}_{s})$ and $f_h(\hat{\Mx\Sigma}_{s})$ for each $j$,. Consequently, for a fixed iteration count $m$, there is no increase in the order of the total computational complexity.
\begin{assumption}
    Suppose that $\sum_{j=1}^m |C_j| \lesssim 1$.
\end{assumption}

The bias term admits a Taylor expansion with respect to the sketch size and  the choice of coefficients~\ref{eq:coefficient} cancels out bias terms of order $s^{-\ell}$ explicitly for $1 \le \ell \le m-1$ as:
\[
\E f(\hat{\Mx{\Sigma}}_s) - f(\Mx{\Sigma}) \approx \sum_{\ell=1}^{m-1} \frac{\beta_{\ell,m}(\Mx{\Sigma})}{s^\ell} \text{ and } \sum_{j=1}^m C_j \frac{\beta_{\ell,m}(\Mx{\Sigma})}{s_j^\ell} = \beta_{\ell,m}(\Mx{\Sigma})\sum_{j=1}^m  \frac{C_j}{s_j^\ell} = 0,
\]
with $\beta$ independent to the sketch size. For details, we refer to Appendix~\ref{appx:linear_aggragation}.
We now specialize our bias bound in Theorem~\ref{thm:jk_sketch_bias} and $L_p$ risk bound in Theorem~\ref{thm:jk_lprisk}. The $L_p$ bound only work for Gaussian case, due to a technical limitation via Gaussian concentration property, while the bias bound works in general.

\begin{theorem}\label{thm:jk_sketch_bias}
Under the preceding assumptions on the sketch sizes and
aggregation coefficients, let $\S \in \mathbb{R}^{s \times n}$ be an isotropic sketch matrix. For arbitrary $\b\in\mathbb{R}^n$, the Hessian sketch estimator satisfies
\begin{equation}
\left\|
\E\hat T_{h,m}-f_h(\Mx\Sigma)
\right\|_\infty
\lesssim_{m,c,\gamma,\sigma_1,\|\b\|}
\left[
\kappa\left(
\sqrt{\frac{r}{s}}\vee\frac{r}{s}
\right)
\right]^{m+1}.
\end{equation}
Similarly, the classic sketch estimator satisfies
\begin{equation}
\left\|
\E\hat T_{c,m}-f_c(\Mx\Sigma)
\right\|_\infty
\lesssim_{m,c,\gamma,\sigma_1,\|\b\|}
\left[
\kappa\left(
\sqrt{\frac{r_+}{s}}\vee\frac{r_+}{s}
\right)
\right]^{m+1}.
\end{equation}
Here $\kappa = \|\z\|^2_{\psi_2}$ depends on the distribution. Under Gaussian sketching, $r_+$ in the classic sketch
bias bound can be replaced by $r$.
\end{theorem}

\begin{theorem}\label{thm:jk_lprisk}
Under the preceding assumptions on the sketch sizes and aggregation coefficients, suppose that the sketches are Gaussian. For arbitrary $\b\in\mathbb{R}^n$, every $1\le p<\infty$, and every $1\le i\le r$, the Hessian sketch estimator
satisfies
\begin{equation}
    \left\|\langle \hat T_{h,m}- f_{h}(\Mx\Sigma), \e_i \rangle \right\|_{L_p}  
    \lesssim_{m,\gamma, \sigma_1,\|\b\|} 
    \left( \sqrt{\frac{p}{s}} \left( \frac{r}{s} \vee 1 \right) + \frac{p}{s} \right) + \left(\sqrt{\frac{r}{s}} \vee \frac{r}{s} \right)^{m+1}.
\end{equation}
Similarly, the classic sketch satisfies that
\begin{equation}
        \left\|\langle \hat T_{c,m}- f_c(\Mx\Sigma), \e_i  \rangle\right\|_{L_p} 
        \lesssim_{m,\gamma, \sigma_1,\|\b\|} 
        \left( \sqrt{\frac{p}{s}} \left( \frac{r_+}{s} \vee 1 \right) + \frac{p}{s} \right) + \left(\sqrt{\frac{r}{s}} \vee \frac{r}{s} \right)^{m+1}.
\end{equation}
\end{theorem}

The two terms in each bound control the random error around the estimator's expectation and its bias, respectively. For the classic sketch, $\b_\perp$ only contribute random error but no bias, under Gaussian setting. As a consequence, we account for this additional source of random error using the augmented representation of dimension $r_+$ while the bias part within dimension $r$. The Hessian sketch is not affected and both terms are expressed using $r$.

Similar to the bootstrap, the asymptotic conclusion also holds for the linear aggregation estimator. If $r \asymp s^\alpha$, increase $m$ makes the bias decay faster, achieving a final order $s^{-1/2}$. However, without debiasing, original sketch methods only achieve order of $s^{\alpha - 1}$.
\begin{corollary}
    If $\S \in \mathbb{R}^{s \times n}$ is chosen as a Gaussian sketch matrix and $s^\alpha \asymp r$, for $\alpha \in (1/2, 1)$. Choose $m \ge \lceil 1/(1-\alpha) \rceil - 1$, then for every $1 \le i \le r$,
    $$
    \left\|\langle \hat{T}_{h,m} - f_h(\Mx\Sigma), \e_i \rangle\right\|_{L_p} \lesssim_{m,\sigma_1, \gamma, \|\b\|,p} s^{-1/2}, \quad
    \left\|\langle \hat{T}_{c,m} - f_c(\Mx\Sigma), \e_i \rangle\right\|_{L_p} \lesssim_{m,\sigma_1, \gamma, \|\b\|,p} s^{-1/2}.
    $$
\end{corollary}

The bootstrap and linear aggregation are similar in spirit, but different. One difference here is the index $m$ for the linear aggregation method is the number of plug-in estimators, starting at $1$, while the index $k$ for bootstrap method is the times of resampling, starting at $0$. When $m = 1$ and $k = 0$, both are the single plug-in estimator, i.e the original sketch solution. 
Furthermore if $s > r$, and we pick $m = k + 1$, both methods yield same bias bound of order
$(\sqrt{r/s} )^{m+1}$ and same $L_p$ risk of order $(\sqrt{1/s} + (\sqrt{r/s})^{m+1})$ in the Gaussian case. In particular, both approaches are linear combinations of $m = k+1$ different estimators.

\section{Conclusions and Open Questions}
This work established theoretical guarantees for two resampling-based debiasing methods, iterative bootstrap and linear aggregation, for sketched ridge regression. We showed that these methods achieve higher-order bias reduction, and for Gaussian sketches we further derived $L_p$ risk bounds.

Several questions remain open. In particular, $L_p$ risk bounds for non-Gaussian sketches are still unknown, and a principled analysis of bootstrap debiasing under discrete sketching distributions is not yet available. In addition, our results provide only upper bounds, while matching lower bounds remain open. More broadly, it would be of interest to extend the analysis beyond Gaussian models and obtain sharper distribution-dependent risk guarantees.

\begin{ack}
The authors gratefully acknowledge support from the Office of Naval Research under award number N00014-23-1-2545 (Untangling Computation). FS was also supported by the Alfred P. Sloan Foundation via a Sloan Research Fellowship in Mathematics.
\end{ack}

\bibliography{ref}

\appendix



\section{Organization of the appendices}
Appendix~\ref{appx:notations} introduces additional notation and some well used sketch methods with their sub-Gaussian norm bound. Appendix~\ref{appx:smoothness} collects derivative and smoothness bounds for the target functionals $f_c$ and $f_h$. Appendix~\ref{appx:bias} provides auxiliary results, as well as the proofs of Theorems~\ref{thm:operator_norm} and \ref{thm:bias_bound}. Appendix~\ref{appx:bootstrap} contains the proof for the iterative bootstrap method, while Appendix~\ref{appx:linear_aggragation} contains the proof for the linear aggregation method. Additional experiments are presented in Appendix~\ref{appx:experiment}. We evaluate bias reduction across sketching methods and sampling distributions, compare iterative methods with larger single sketches, contrast empirical and parametric bootstrap, and examine bias decay and the influence of estimator variance.


\section{Additional Notations and 
Definitions}\label{appx:notations}
A centered random variable $X$ in a Banach space $E$ is called pre-Gaussian iff there exist a centered Gaussian random variable $Y \in E$ with the same covariance operator. Additionally, a centered random variable $X \in E$ is called sub-Gaussian \citep{MR1102015} iff for all $u \in E^*$,
\[
\left\|\langle X, u \rangle\right\|_{\psi_2} \lesssim \left\|\langle X, u \rangle \right\|_{L_2}.
\]

We introduce definitions of functional smoothness on Banach spaces, which will be used only in Appendix~\ref{appx:bootstrap}.
Let $E,F$ be two Banach spaces and let $U \subset E$. For a function $g: U \mapsto F$, the $L_{\infty}$ norm and Lipschitz norm are defined as follows
\[
\|g\|_{L_{\infty}(U)}:=\sup _{x \in U}\|g(x)\|, \quad
\|g\|_{\operatorname{Lip}(U)}:=\sup _{x, x^{\prime} \in U, x \neq x^{\prime}} \frac{\left\|g(x)-g\left(x^{\prime}\right)\right\|}{\left\|x-x^{\prime}\right\|},
\]
and, for $\rho \in(0,1]$,
\[
\|g\|_{\operatorname{Lip}_\rho(U)}:=\sup _{x, x^{\prime} \in U, x \neq x^{\prime}} \frac{\left\|g(x)-g\left(x^{\prime}\right)\right\|}{\left\|x-x^{\prime}\right\|^\rho}.
\]

Let $s=m+\rho, m \geq 0, \rho \in(0,1]$. For an $m$-times Fréchet differentiable function $g$ from $U$ into $F$, define
\[
\|g\|_{C^s(U ; F)}:=\max \left(\|g\|_{L_{\infty}(U)}, \max _{0 \leq j \leq m-1}\left\|g^{(j)}\right\|_{\operatorname{Lip}(U)},\left\|g^{(m)}\right\|_{\operatorname{Lip}_\rho(U)}\right),
\]
We will omit $F$ if $F = \mathbb{R}$.

\subsection{Some well-used sketch scheme}\label{appx:sketch_dist}
We introduce three standard sketching distributions that are widely used in randomized least squares and related problems.

\paragraph{Uniform Sampling.}
Each row of $\S$ selects a single standard basis vector.
A probabilistic description is
\[
\mathbb{P}\left(\s = \sqrt{\frac{n}{s}}\e_i\right)=\frac{1}{n},\quad 1\le i\le n.
\]
Equivalently, $\S\A$ consists of $s$ rows of $\A$ sampled uniformly at random with replacement and scaled by $\sqrt{n/s}$ so that $\E \S^\intercal\S=\I_n$ consequently.
This choice is attractive when the rows of $\A$ have roughly comparable ``importance'', and it is computationally cheap since forming $\S\A$ reduces to selecting and rescaling rows.
However, uniform sampling may require a larger sketch to achieve the same accuracy as other distributions.

\paragraph{Leverage Score Sampling.}
Leverage score sampling is also a sparse row-sampling sketch, but it adapts to the geometry of $\A$.
Let $\A=\U\Mx{\Lambda}\V^\intercal$ be the compact SVD with $i$-th row in $\U$ to be $\u_i^\intercal \in \mathbb{R}^r$. Define the $i$-th leverage score of matrix $\A$ as
\[
\ell_i:=\|\u_{i}\|_2^2,\qquad 1\le i\le n.
\]
The leverage scores satisfy that $\sum_{i=1}^n\ell_i=r$.
Leverage score sampling chooses row indices with probabilities proportional to $\ell_i$, i.e
\[
\mathbb{P}\left(\s=\sqrt{\frac{1}{sp_i}}\e_i\right)=p_i,\quad p_i:=\frac{\ell_i}{r}.
\]
The scaling $\sqrt{1/(sp_i)}$ here again ensures $\E \S^\intercal\S=\I_n$.
Intuitively, rows with larger leverage score contribute more, so sampling them more frequently. In practice, exact leverage scores require the computation of $\U$; many implementations use approximate leverage scores~\citep{drineas2012fast} or shrinkage leverage scores~\citep{ma2015statistical}.

\paragraph{Gaussian Sampling.}
Each row is an independent Gaussian vector
\[
\s\sim\mathcal{N}\left(\Vc{0},\frac{1}{s}\I_n\right),
\]
or equivalently, $\S_{ij}\stackrel{\text{i.i.d.}}{\sim}\mathcal{N}(0,1/s)$.
This sketch is rotationally invariant and typically yields clean theoretical guarantees. They are closely related to Johnson--Lindenstrauss embeddings~\citep{Johnson_Lindenstrauss}.
Its main drawback is computational. Forming $\S\A$ is more expensive than sparse sampling sketches because $\S$ is dense.

We highlight here, all these three sketch scheme satisfy the isotropic condition.
\begin{lemma}
If $\S$ is chosen to be uniform, leverange score or Gaussian sketch matrix, then $\S$ is an isotropic sketch matrix.
\end{lemma}
\begin{proof}
Since $\U$ has orthonormal columns, it sufficient to check $\E \s_i \s_i^\intercal = \I_n / s$, which implies that
\[
\E \z_i \z_i^\intercal = s \U^\intercal\E \left[\s_i \s_i^\intercal \right] \U
\]
It's clear here, when $\S$ is uniform sketch,
\[
\E \s_i \s_i^\intercal = \frac{1}{n} \sum_{i=1}^n \frac{n}{s} \Mx{E}_i  = \frac{1}{s}  \I_n ,
\]
when $\S$ is leverage score sketch,
\[
\E \s_i \s_i^\intercal = \sum_{i=1}^n \frac{1}{s p_i}\Mx{E}_i \cdot p_i = \frac{1}{s} \I_n ,
\]
and when $\S$ is Gaussian, it's the covariance matrix:
\[
\E \s_i \s_i^\intercal = \frac{1}{s}  \I_n .
\]
\end{proof}

\begin{lemma}
    If $\S$ is chosen as uniform sampling, then $\|\z\|_{\psi_2} \lesssim \sqrt{\frac{n}{\log n}}$. If $\S$ is chosen as leverage score sampling, then $\|\z\|_{\psi_2} \lesssim \sqrt{\frac{r}{\log r}}$.
\end{lemma}
\begin{proof}
For the Uniform case, 
\[
\|\z\|_{\psi_2} = \sup_{\v \in S^{r-1}}\|\langle \z, \v \rangle\|_{\psi_2}.
\]
Given $\v \in S^{r-1}$,
\[
\|\langle \z, \v \rangle\|_{\psi_2} = \inf \left \{t > 0: \frac{1}{n} \sum_{i=1}^{n} \exp \left(\frac{n |\langle \u_i, \v\rangle|^2}{t^2}\right ) \le 2 \right \}.
\]
Notice here $\U^{\intercal}\U = \I_r$,
\[
\sum_{i=1}^{n} |\langle \u_i, \v\rangle|^2 = \|\v\|_2^2 = 1.
\]
Then by the convexity of exponential function, we have
\[
\frac{1}{n} \sum_{i=1}^{n} \exp \left(\frac{n |\langle \u_i, \v\rangle|^2}{t^2}\right ) \le \frac{n-1}{n} + \frac{1}{n} \exp(n/t^2),
\]
implying $\|\z\|_{\psi_2} \lesssim \sqrt{\frac{n}{\log n}}$.

Now we move to the leverage score sampling.
Similarly, 
\[
\|\langle \z, \v \rangle\|_{\psi_2} = \inf \left \{t > 0: \sum_{i=1}^{n} \frac{\ell_i}{r}  \exp \left(\frac{r |\langle \u_i, \v\rangle|^2}{\ell_i t^2}\right ) \le 2 \right \}.
\]
For each fixed unit vector
$\v \in \mathbb{R}^r$, choose an orthonormal basis
$q_1,\ldots,q_r$ with $q_1=\v$. Write
\[
\u_i=\sum_{j=1}^r w_{ij}q_j,
\qquad
\v=\sum_{j=1}^r k_jq_j,
\qquad
k_1=1,\quad k_j=0 \quad (j\ge2).
\]
Then
\[
|\langle \u_i,\v\rangle|^2
=w_{i1}^2
=\sum_{j=1}^r w_{ij}^2k_j^2.
\]
Moreover, orthonormality and $\U^\intercal\U=\I_r$ imply
\[
\ell_i=\sum_{j=1}^r w_{ij}^2,
\qquad
\sum_{i=1}^n w_{ij}^2=1
\quad \text{for every }j.
\]
Rows with $\ell_i=0$ have zero sampling probability and are
omitted from the following sums. For every $i$ with $\ell_i>0$
and every $t>0$, convexity of the exponential gives
\[
\exp\left(
    \frac{r|\langle \u_i,\v\rangle|^2}{\ell_i t^2}
\right)
=
\exp\left(
    \sum_{j=1}^r \frac{w_{ij}^2}{\ell_i}
    \frac{rk_j^2}{t^2}
\right)
\le
\sum_{j=1}^r \frac{w_{ij}^2}{\ell_i}
\exp\left(\frac{rk_j^2}{t^2}\right).
\]
Multiplying by $\ell_i/r$ and summing over $i$ yields
\begin{align*}
\E\exp\left(\frac{|\langle \z,\v\rangle|^2}{t^2}\right)
&=
\sum_{i:\ell_i>0}\frac{\ell_i}{r}
\exp\left(
    \frac{r|\langle \u_i,\v\rangle|^2}{\ell_i t^2}
\right) \\
&\le
\frac{1}{r}\sum_{j=1}^r
\exp\left(\frac{rk_j^2}{t^2}\right) \\
&=
\frac{\exp(r/t^2)+r-1}{r}.
\end{align*}
Choosing $t^2=r/\log(1+r)$ makes the last expression equal to $2$.
Therefore, by the definition of the $\psi_2$ norm,
\[
\|\z\|_{\psi_2}
=
\sup_{\|\v\|_2=1}\|\langle \z,\v\rangle\|_{\psi_2}
\le
\sqrt{\frac{r}{\log(1+r)}}.
\]
\end{proof}

Thu uniform sketch has larger sub-Gaussian norm $\|\z\|_{\psi_2}$ since the random vector has scale $\sqrt{n}$ compared with leverage score sampling. Note here the sub-Gaussian norm here depends on the structure of matrix $\U$ since the rows of $\U$ may reduce the scale of the random vector $\z$ compared with $\s$ in discrete case. For example, when $\U = \begin{bmatrix}
    \I_r \\ 
    0
\end{bmatrix}$,
then for uniform sampling, $\z$ has probability $1/n$ to be $\sqrt{n} \e_i$, for that case, the upper bound on sub-Gaussian norm is attained. Similarly, for the same $\U$, it has $r$ leverage score to be $1$ and $n-r$ to be 0, then leverage score sampling is to draw $\sqrt{r}\e_i$ uniformly, with upper bound attained. But when $\U$ has evenly leverage scores, multiplying $\U$ would scale $\z$ by $\sqrt{r/n}$ from $\s$, achieve smaller sub-Gaussian norm. 

\section{Smoothness of Target Functional}\label{appx:smoothness}
Our theory highly relies on the smoothness of the functional $f_c$ and $f_h$. Due to regularization term, the derivatives can be upper bounded as below. We remind the reader that, the dimension is of $r$ for Hessian sketch and $r_+$ for classic sketch.

\begin{lemma}\label{lemma:derivative_bounds}
Let $\A\in\mathbb R^{n\times r}$ have full column rank, let $\b\in\mathbb R^n$, let $\gamma>0$, and set $\sigma_1=\|\A\|$. Write the compact SVD as $\A=\U\Mx{\Lambda}\V^\intercal$. Let $\U_+$ have $r_+$ orthonormal columns forming a basis for the column space of $[\A\ \b]$. For $1\le i\le r$, define
\[
g_{h,i}(\X)=\langle f_h(\X),\e_i\rangle,
\quad
g_{c,i}(\X)=\langle f_c(\X),\e_i\rangle.
\]
Here derivative norms are multilinear operator norms with the matrix operator norm on each symmetric argument; $L_\infty$ denotes the supremum of these norms over the indicated domain.

For every integer $j\ge1$,
\[
\|D^j g_{h,i}\|_{L_\infty(\mathbb S_+^r)}
\le
j!\frac{\sigma_1^{2j}}{\gamma^{j+1}}
\|\A^\intercal\b\|_2.
\]
For every $\X\in\mathbb S_+^{r_+}$,
\[
\|D^j g_{c,i}(\X)\|
\le
j!\frac{\sigma_1^{2j-1}}{\gamma^j}
\|\b-\A f_c(\X)\|_2.
\]
In particular,
\[
\|D^j g_{c,i}(\I_{r_+})\|
\le
j!\frac{\sigma_1^{2j-1}}{\gamma^j}\|\b\|_2.
\]
\end{lemma}

\begin{proof}
Recall the definitions of the two functionals:
\[
f_h(\X)=M_h(\X)^{-1}\A^\intercal\b,
\qquad
f_c(\X)=M_c(\X)^{-1}
\A^\intercal(\U_+\X\U_+^\intercal)\b,
\]
where
\begin{align*}
M_h(\X)
&=
\V\Mx{\Lambda}^\intercal\X\Mx{\Lambda}\V^\intercal
+\gamma\I_r,
\\
M_c(\X)
&=
\A^\intercal(\U_+\X\U_+^\intercal)\A
+\gamma\I_r.
\end{align*}
The data, the bases, and $\gamma$ are fixed throughout
differentiation.
For either $q\in\{h,c\}$ and positive-semidefinite $\X$
in the corresponding domain,
\[
M_q(\X)\succeq\gamma\I_r,
\qquad
\|M_q(\X)^{-1}\|\le\gamma^{-1},
\qquad
\|DM_q(\X)[\H]\|\le\sigma_1^2\|\H\|.
\]
The same formulas define smooth functions on an open neighborhood of each such $\X$ in the space of symmetric matrices, where $M_q$ remains positive definite.

For the Hessian functional, differentiate $M_h(\X)f_h(\X)=\A^\intercal\b$. Since $M_h$ is affine, for arbitrary symmetric directions $\H_1,\ldots,\H_j$ and every $j\ge1$,
\begin{align*}
D^j f_h(\X)[\H_1,\ldots,\H_j]
=&-M_h(\X)^{-1}
\sum_{\ell=1}^j
\Bigl(DM_h(\X)[\H_\ell]
\cdot D^{j-1}f_h(\X)
[\H_1,\ldots,\widehat{\H_\ell},\ldots,\H_j]\Bigr),
\end{align*}
where the hat denotes omission and
$D^0f_h(\X)=f_h(\X)$.
Using the Euclidean norm on vector-valued outputs gives
\[
\|D^j f_h(\X)\|
\le
\frac{j\sigma_1^2}{\gamma}
\|D^{j-1}f_h(\X)\|.
\]
Starting from
$\|f_h(\X)\|_2\le\gamma^{-1}\|\A^\intercal\b\|_2$,
induction yields
\[
\|D^j f_h(\X)\|
\le
j!\frac{\sigma_1^{2j}}{\gamma^{j+1}}
\|\A^\intercal\b\|_2.
\]

For the classic functional, recall
\[
M_c(\X)f_c(\X)
=
\A^\intercal\U_+\X\U_+^\intercal\b,
\qquad
M_c(\X)=\A^\intercal\U_+\X\U_+^\intercal\A+\gamma\I_r.
\]
Differentiating in direction $\H$ gives
\[
DM_c(\X)[\H]\,f_c(\X)
+
M_c(\X)Df_c(\X)[\H]
=
\A^\intercal\U_+\H\U_+^\intercal\b.
\]
Since
\[
DM_c(\X)[\H]
=
\A^\intercal\U_+\H\U_+^\intercal\A,
\]
we obtain
\[
\begin{aligned}
M_c(\X)Df_c(\X)[\H]
&=
\A^\intercal\U_+\H\U_+^\intercal\b
-
\A^\intercal\U_+\H\U_+^\intercal\A f_c(\X)\\
&=
\A^\intercal\U_+\H\U_+^\intercal
\bigl(\b-\A f_c(\X)\bigr).
\end{aligned}
\]
As a result,
\[
Df_c(\X)[\H]
=
M_c(\X)^{-1}\A^\intercal\U_+\H\U_+^\intercal
\bigl(\b-\A f_c(\X)\bigr).
\]
Hence
\[
\|Df_c(\X)\|
\le
\frac{\sigma_1}{\gamma}
\|\b-\A f_c(\X)\|_2.
\]
The right-hand side of
\[
M_c(\X)f_c(\X)
=
\A^\intercal(\U_+\X\U_+^\intercal)\b
\]
is affine in $\X$.
Thus, for $j\ge2$, the same differentiation argument gives
\[
\|D^j f_c(\X)\|
\le
\frac{j\sigma_1^2}{\gamma}
\|D^{j-1}f_c(\X)\|.
\]
Induction at the fixed matrix $\X$, starting from the
first-derivative bound, therefore gives
\[
\|D^j f_c(\X)\|
\le
j!\frac{\sigma_1^{2j-1}}{\gamma^j}
\|\b-\A f_c(\X)\|_2.
\]

For each coordinate,
\[
D^j g_{q,i}(\X)[\H_1,\ldots,\H_j]
=
\left\langle
D^j f_q(\X)[\H_1,\ldots,\H_j],\e_i
\right\rangle.
\]
Since $\|\e_i\|_2=1$, the same bounds hold for
$D^j g_{q,i}(\X)$.

Finally, since $\U_+\U_+^\intercal\A=\A$ and
$\U_+\U_+^\intercal\b=\b$,
\[
f_c(\I_{r_+})
=
\x^*
=
(\A^\intercal\A+\gamma\I_r)^{-1}\A^\intercal\b.
\]
Ridge optimality gives
\[
\|\A\x^*-\b\|_2^2+\gamma\|\x^*\|_2^2
\le
\|\b\|_2^2.
\]
Thus $\|\b-\A\x^*\|_2\le\|\b\|_2$, proving the
population-point bound.
\end{proof}

\section{Proof of Theorem~\ref{thm:operator_norm} and Theorem~\ref{thm:bias_bound}}\label{appx:bias}
\begin{lemma}\label{lemma:operator_bound}
    Let $X$ be weakly square integrable centered random vectors in $\mathbb{R}^d$ with covariance matrix $\Sigma$. Assume $X$ is pre-Gaussian and $\left\|\langle X, u \rangle\right\|_{\psi_2} \le K \left\|\langle X, u \rangle \right\|_{L_2}$ for some $K$ and all unit vector $u$. Let $X_1, \ldots, X_n$ be i.i.d. copies of $X$, with covariance matrix estimation $\hat{\Sigma}_n = \frac{1}{n} \sum_{i=1}^n X_i X_i^\intercal$ then there exists a constant $C>0$ such that, for all $t \geq 1$, with probability at least $1-e^{-t}$,
    \[
    \|\hat{\Sigma}_n-\Sigma\| \leq C K^2 \|\Sigma\|\left(\sqrt{\frac{d}{n}} \vee \frac{d}{n} \vee \sqrt{\frac{t}{n}} \vee \frac{t}{n}\right) .
    \]
\end{lemma}
\begin{proof}
The result follows by applying Theorem~8 in \citep{Koltchinskii_Lounici} and repeating the argument used in the proof of Theorem~4 of \citep{Koltchinskii_Lounici}. Our lemma can be treated as a finite-dimensional version of previous result with explicit dependence on $K$. \citet{Koltchinskii_Lounici} only focused on the sub-Gaussian case.
\end{proof}

\begin{proposition}\label{prop:operator_bound}
Under the same assumption as Lemma~\ref{lemma:operator_bound},
\[
\E^{1/p} \|\hat{\Sigma} - \Sigma\|^p \lesssim_p K^2\|\Sigma\| \left(\sqrt{\frac{d}{n}} \vee \frac{d}{n} \right)
\]
\end{proposition}
\begin{proof}
It's sufficient to integral the tail probability. Here,
\[
\E \|\hat{\Sigma} - \Sigma\|^p = p \int_{0}^\infty u^{p-1}\mathbb{P}(\|\hat{\Sigma} - \Sigma\| > u)du.
\]
Set
\[
B_n:=\sqrt{\frac{d}{n}}\vee \frac{d}{n}.
\]
By Lemma~E.1, there exists a constant \(C>0\) such that for every \(t\ge 1\),
\[
\mathbb{P}\!\left(
\|\hat{\Sigma}_n-\Sigma\|
>
C K^2 \|\Sigma\|
\Bigl(
B_n \vee \sqrt{\tfrac{t}{n}} \vee \tfrac{t}{n}
\Bigr)
\right)
\le e^{-t}.
\]
Since \(d\ge 1\), we have \(B_n\ge n^{-1/2}\) and \(B_n\ge n^{-1}\). Hence, for all \(t\ge 1\),
\[
B_n \vee \sqrt{\tfrac{t}{n}} \vee \tfrac{t}{n}
\le
B_n \vee \sqrt{t}\,B_n \vee t B_n
\le t B_n.
\]
Therefore,
\[
\mathbb{P}\!\left(
\|\hat{\Sigma}_n-\Sigma\|
>
C K^2 \|\Sigma\| B_n\, t
\right)
\le e^{-t},\qquad t\ge 1.
\]

Now define
\[
Y:=\frac{\|\hat{\Sigma}_n-\Sigma\|}{C K^2 \|\Sigma\| B_n}.
\]
Then \(\mathbb{P}(Y>t)\le e^{-t}\) for all \(t\ge 1\). Using the tail integration formula,
\[
\mathbb{E}Y^p
=
p\int_0^\infty u^{p-1}\mathbb{P}(Y>u)\,du,
\]
we obtain
\[
\mathbb{E}Y^p
\le
p\int_0^1 u^{p-1}\,du
+
p\int_1^\infty u^{p-1}e^{-u}\,du
\le C_p,
\]
where \(C_p>0\) depends only on \(p\). Taking \(p\)-th roots gives
\[
\bigl(\mathbb{E}\|\hat{\Sigma}_n-\Sigma\|^p\bigr)^{1/p}
\le
C_p K^2 \|\Sigma\| B_n
=
C_p K^2 \|\Sigma\|
\left(\sqrt{\frac{d}{n}}\vee \frac{d}{n}\right).
\]
This proves the claim.
\end{proof}

\begin{proof}[Proof of Theorem~\ref{thm:operator_norm}]
For non-centered sketch method, it's sufficient to consider a new random vector $\s^\prime = \varepsilon \s$ for scalar valued Rademacher random variable $\varepsilon$ which is independent with $\s$. Notice that $\s\s^\intercal$ has the same distribution as $\s^\prime\s^{\prime\intercal}$. Due to the isotropic property, $\|\Mx\Sigma\| = 1$, which implies that  $\left\|\langle \s, \u \rangle \right\|_{L_2} = 1$ for unit vector $\u$. As a result, we can directly apply Proposition~\ref{prop:operator_bound} 
\end{proof}

Before proving Theorem~\ref{thm:bias_bound}, we establish a useful identity for the classic sketch under Gaussian sketching. Write
\[
\b=\U\U^\intercal\b+\b_\perp,
\qquad
\b_\perp=(\I_n-\U\U^\intercal)\b.
\]
Since $\A^\intercal\b_\perp=0$, the jointly Gaussian random matrix $\S\A$ and vector $\S\b_\perp$ have zero cross-covariances and are therefore independent. Consequently,
\[
\E[\S\b_\perp\mid\S\A]=0.
\]
Write $\hat{\x}_c(\S;\v)$ for the classic sketch estimator~\eqref{eq:classic_sketch} with response $\v$. By linearity in the response,
\begin{align*}
\E\!\left[
\hat{\x}_c(\S;\b)
-\hat{\x}_c(\S;\U\U^\intercal\b)
\,\middle|\,\S\A
\right] =
\left((\S\A)^\intercal(\S\A)+\gamma\I_r\right)^{-1}
(\S\A)^\intercal
\E[\S\b_\perp\mid\S\A]
=0.
\end{align*}
These expectations are well defined because $\gamma>0$ and the Gaussian sketch has finite second moments. Taking expectations yields
\begin{equation}\label{eq:gaussian_projection_identity}
\E\hat{\x}_c(\S;\b)
=
\E\hat{\x}_c(\S;\U\U^\intercal\b).
\end{equation}
Moreover, $\A^\intercal\U\U^\intercal\b=\A^\intercal\b$, so projecting the response leaves the exact ridge solution unchanged. Thus, the Gaussian classic-sketch bias coincides with
that of the projected-response problem. The latter estimator depends only on $\S\U$, allowing its bias to be analyzed using $r$-dimensional compressed samples rather than the $r_+$-dimensional augmented samples.

To distinguish the projected-response problem from the augmented formulation, define
$\widetilde f_c:\mathbb S_+^r\to\mathbb R^r$ by
\[
\widetilde f_c(\X)
=
\left(
\V\Mx{\Lambda}^\intercal\X\Mx{\Lambda}\V^\intercal
+\gamma\I_r
\right)^{-1}
\V\Mx{\Lambda}^\intercal\X\U^\intercal\b.
\]
Here $\A=\U\Mx{\Lambda}\V^\intercal$ is the compact SVD of $\A$. Then
\[
\hat{\x}_c(\S;\U\U^\intercal\b)
=
\widetilde f_c(\U^\intercal\S^\intercal\S\U),
\qquad
\widetilde f_c(\I_r)=\x^*.
\]
Under Gaussian sketching, Equation~\eqref{eq:gaussian_projection_identity} therefore gives
\[
\E f_c(\U_+^\intercal\S^\intercal\S\U_+)
-f_c(\I_{r_+})
=
\E\widetilde f_c(\U^\intercal\S^\intercal\S\U)
-\widetilde f_c(\I_r).
\]
Thus, the Gaussian bias of the augmented classic functional equals the bias of the $r$-dimensional projected-response functional $\widetilde f_c$. 

With a slight abuse of notation, we continue to write $f_c$ for both functionals. In the Gaussian bias analysis, $f_c$ denotes the $r$-dimensional projected-response functional $\widetilde f_c$, using the equality of biases established above. In the $L_p$-risk analysis and the non-Gaussian bias analysis, $f_c$ denotes the original augmented functional on
$\mathbb S_+^{r_+}$. The corresponding second-moment matrices are understood to have the appropriate dimensions.

\begin{proof}[Proof of Theorem~\ref{thm:bias_bound}]
It's sufficient to take the Taylor expansion of two solution functions $f_c$ and $f_h$. For simplicity, we only provide the analysis for $f = \langle f_h, \e_1 \rangle$. The proof is exactly the same for $\langle f_c, \e_i \rangle$ and $\langle f_h, \e_i \rangle$ for each $i$.

Notice that, 
$$
f(\hat{\Mx\Sigma}_s) = f(\Mx\Sigma) + \langle f^\prime(\Mx\Sigma) , \hat{\Mx\Sigma}_s - \Mx\Sigma\rangle + R(\Mx\Sigma; \hat{\Mx\Sigma}_s - \Mx\Sigma),
$$
with the remainder term 
$$
\left|R(\Mx\Sigma; \hat{\Mx\Sigma}_s - \Mx\Sigma)\right|\le\frac{1}{2}  \left\|f^{(2)} \right\|_{L_\infty}\left\|\hat{\Mx\Sigma}_s - \Mx\Sigma\right\|^2 .
$$
By taking expectation on both size, we can get a formula of the bias term as 
$$
\E f(\hat{\Mx\Sigma}_s) - f(\Mx\Sigma) = \langle f^\prime(\Mx\Sigma) , \E \hat{\Mx\Sigma}_s - \Mx\Sigma\rangle + \E  R(\Mx\Sigma; \hat{\Mx\Sigma}_s - \Mx\Sigma).
$$
Since  $\hat{\Mx\Sigma}_s$ is unbiased estimator of $\Mx\Sigma$, we have
\begin{align*}
\left|\E f(\hat{\Mx\Sigma}_s) - f(\Mx\Sigma)\right| 
& = \left|\E  R(\Mx\Sigma; \hat{\Mx\Sigma}_s - \Mx\Sigma)\right| \le \E \left|  R(\Mx\Sigma; \hat{\Mx\Sigma}_s - \Mx\Sigma)\right| \\
& \lesssim \left\|f^{(2)} \right\|_{L_\infty} \E \left\|\hat{\Mx\Sigma} - \Mx\Sigma\right\|^2 \lesssim _{\sigma_1, \gamma, \|\w\|}  \kappa^2\left(\sqrt{\frac{r}{s}} \vee \frac{r}{s} \right)^2.
\end{align*}
One can finally conclude the result for every $\langle f_c, \e_i\rangle$. For the Gaussian case, the bias is analyzed with the projected version.
\end{proof}

\section{Proof of Theorem~\ref{thm:bootstrap_bias} and Theorem~\ref{thm:bootstrap_lp}}\label{appx:bootstrap}
One can directly use the following Theorem~\ref{thm:bootstrap_key} to show Theorem~\ref{thm:bootstrap_bias} and Theorem~\ref{thm:bootstrap_lp} by choosing $a = 1$ and pick $f(\Mx\Sigma) = \langle f_c(\Mx\Sigma), \e_i \rangle $ or $f(\Mx\Sigma) = \langle f_h(\Mx\Sigma), \e_i \rangle $ for arbitrary $i$. Note here for the bias analysis, we use the projected functional $\widetilde{f}_c$.

\begin{theorem}\label{thm:bootstrap_key}
Let $X_1, \dots , X_n$ be i.i.d. random variables sampled from a Gaussian distribution $\mathcal{N}(0,\Sigma)$ in $\mathbb{R}^d$ with unknown parameter $\theta =\Sigma \in \Theta:=\mathbb{S}_{++}^d$. Define
$$
\Theta(a;d) = \{\Sigma \in \mathbb{S}_{++}^d: \sigma(\Sigma) \subset [1/a, a]\}, \quad a \ge 1,
$$
where $\sigma(\Sigma)$ is the spectrum of covariance $\Sigma$.

Suppose $f \in C^s(\Theta)$ for some $s=k+1+\rho,k\ge1,\rho \in (0,1]$
and assume that, for some $\epsilon > 0$,
$$
a^{s-1+\epsilon} \sqrt{\frac{d}{n}} \le c_s,
$$
where $c_s \le 1/2$ is a sufficiently small constant. 
Then for all $\theta \in \Theta(a;d)$,
$$
|\E_{\theta}f_k(\hat\theta) - f(\theta)| \lesssim_{s,\epsilon} \|f\|_{C^s(\Theta)}a^{(s-1+\epsilon)k+1+\rho}\left(\sqrt{\frac{d}{n}}\right)^s
$$    
and
$$
\sup _{\theta \in \Theta(a ; d)}\left\|f_k(\hat{\theta})-f(\theta)\right\|_{L_p} \lesssim_{s, \epsilon, p}\|f\|_{C^s(\Theta)}\left(\frac{a}{\sqrt{n}} \vee a^{(s-1+\epsilon) k+1+\rho}\left(\sqrt{\frac{d}{n}}\right)^s\right).
$$
\end{theorem}
This theorem can be shown with same technique used for Theorem 5.1 and Proposition 5.2 in \citep{Koltchinskii_Zhilova}. However, their results do not directly imply the present theorem: \cite{Koltchinskii_Zhilova} study functionals $f$ that depend on both the mean and the covariance under Gaussian case, so the relevant covariance estimator is 
\[
\frac{1}{n-1}\sum_{i=1}^{n}(X_i - \bar X)(X_i - \bar X)^\intercal
\]
which differs from ours due to the additional uncertainty in the mean.

\section{Proof of Theorem~\ref{thm:jk_sketch_bias} and Theorem~\ref{thm:jk_lprisk}}\label{appx:linear_aggragation}

\begin{lemma}\label{lemma:taylor}
Let $f:\mathbb{S}_+^r\to\mathbb{R}$ be a coordinate functional of the sketched solution, i.e., $f(\cdot)=\langle f_c(\cdot),\e_i\rangle$ or $f(\cdot)=\langle f_h(\cdot),\e_i\rangle$.
Let $\hat{\Mx{\Sigma}}_s$ denote the estimator of $\Mx{\Sigma}$ constructed from i.i.d.\ samples $\z_1,\ldots,\z_s$ of a random vector $\z$, where $\z$ is the compressed sample satisfying isotropic condition.
Then,
\begin{equation}
\E f(\hat{\Mx{\Sigma}}_s) - f(\Mx{\Sigma}) = \sum_{\ell=1}^{m-1} \frac{\beta_{\ell,m}(\Mx{\Sigma})}{s^\ell} + R_{s,m}
\end{equation}
with the remainder term
$$
\left| R_{s,m}\right| \lesssim_{m} \left\|f^{(m+1)}\right\|_{L_\infty} \E\left\|\hat{\Mx\Sigma}_s - \Mx\Sigma \right\|^{m+1}.
$$
Here $\beta_{\ell,m} (\Mx\Sigma)$ is a constant coefficient only depends on $f$ and the distribution of $\z$.
\end{lemma}

The expansion in Lemma~\ref{lemma:taylor} makes the mechanism of linear aggregation transparent: by choosing coefficients that cancel the first $m-1$ terms, one eliminates the leading bias terms.
We illustrate this cancellation explicitly in the proof of the following theorem.
\begin{theorem}\label{thm:jk_debias}
Under the assumptions of Lemma~\ref{lemma:taylor}, define the linear aggregation estimator
$$
\hat{T} := \sum_{j=1}^m C_j f(\hat{\Mx\Sigma}_{s_j}),
$$
where the coefficients $\{C_j\}_{j=1}^m$ satisfy \eqref{eq:coefficient}.
Then
\begin{equation}
\left| \E \hat{T} - f(\Mx{\Sigma}) \right|
\lesssim_{m} \left\|f^{(m+1)} \right\|_{L_\infty}
\max_{1 \le j \le m}\E \left\|\hat{\Mx\Sigma}_{s_j} - \Mx\Sigma \right\|^{m+1}.
\end{equation}
\end{theorem}
\begin{proof}
This follows by plugging the expansion in Lemma~\ref{lemma:taylor} into $\E\hat T- f(\Mx\Sigma)$ and using \eqref{eq:coefficient} to cancel the terms $s_j^{-\ell}$ for $1\le \ell\le m-1$.

To be specific, first notice,
\begin{align*}
    \E \hat{T} - f(\Mx\Sigma) = \E\sum_{j=1}^m C_j f(\hat{\Mx{\Sigma}}_{s_j}) - f(\Mx\Sigma) = \sum_{j=1}^m C_j \left[\E f(\hat{\Mx{\Sigma}}_{s_j}) - f(\Mx\Sigma)\right],
\end{align*}
since $\sum_{j=1}^m C_j =1$. Then by Lemma~\ref{lemma:taylor},
\begin{align*}
\E \hat{T} - f(\Mx\Sigma) 
& = \sum_{j=1}^m C_j\left[\sum_{\ell=1}^{m-1} \frac{\beta_{\ell,m}(\Mx{\Sigma})}{s_j^\ell} + R_{s_j} \right] \\
& = \sum_{\ell=1}^{m-1} \beta_{\ell,m}(\Mx{\Sigma})\sum_{j=1}^m \frac{C_j}{s_j^\ell} + \sum_{j=1}^m C_jR_{s_j} \\
& = \sum_{j=1}^m C_jR_{s_j}.
\end{align*}
This imples
\begin{align*}
\left|\E \hat{T} - f(\Mx\Sigma)  \right| 
& = \left| \sum_{j=1}^m C_jR_{s_j} \right| \le \left|\sum_{j=1}^m |C_j| \max_{1 \le j \le m} |R_{s_j}|\right| \\
& \lesssim_{m} \left\|f^{(m+1)}\right\|_{\infty} \max_{1 \le j \le m} \E \left\|\hat{\Mx\Sigma}_{s_j} - \Mx\Sigma\right\|^{m+1}.
\end{align*}
\end{proof}

Now we turn to prove Theorem~\ref{thm:jk_sketch_bias}.
\begin{proof}[Proof of Theorem~\ref{thm:jk_sketch_bias}]

By Theorem~\ref{thm:jk_debias}, it's remaining the check $\|f^{(m+1)}\|_{L_\infty}$ and $\E \|\hat{\Mx\Sigma}_s - \Mx\Sigma\|^{m+1}$, where $f$ is chosen as $f(\cdot)=\langle f_c(\cdot),\e_i\rangle$ or $f(\cdot)=\langle f_h(\cdot),\e_i\rangle$. The derivative bound is provided by Lemma~\ref{lemma:derivative_bounds} and the expectation term is due to Theorem~\ref{thm:operator_norm}.
\end{proof}

\begin{proof}[Proof of Theorem~\ref{thm:jk_lprisk}]

Similar as all other proofs, we can define $f(\cdot)=\langle f_h(\cdot),\e_i\rangle$. It's similar for $\langle f_c(\cdot),\e_i\rangle$. 
It's sufficient to bound the $L_p$ risk for the estimation of $f$, where the linear aggregation estimator of $f$ is $T_f = \sum_{j=1}^m C_j f(\hat{\Mx\Sigma}_{s_j})$.
Notice here, 
\begin{align*}
\left\|T_f - f(\Mx\Sigma)\right\|_{L_p}
& = \left\|T_f - \E T_f + \E T_f - f(\Mx\Sigma)\right\|_{L_p} \\
& \le \sum_{j=1}^m |C_j| \left\|f(\hat{\Mx\Sigma}_{s_j}) - \E f(\hat{\Mx\Sigma}_{s_j})\right\|_{L_p} + |\E T_f - f(\Mx\Sigma)| \\
& \lesssim_m \left\|f(\hat{\Mx\Sigma}_{s}) - \E f(\hat{\Mx\Sigma}_{s})\right\|_{L_p} + |\E T_f - f(\Mx\Sigma)| 
\end{align*}
As a result, we only need to control $\|f(\hat{\Mx\Sigma}_{s}) - \E f(\hat{\Mx\Sigma}_{s})\|_{L_p}$.

Based on Gaussian concentration inequality with local Lipschitz condition, one can show that
\[
\|f(\hat{\Mx\Sigma}_{s}) - \E f(\hat{\Mx\Sigma}_{s})\|_{L_p} \lesssim_{\gamma, \sigma_1,\|w\|} \|\Mx\Sigma\|
\left( \sqrt{\frac{p}{s}} \left( \frac{r}{s} \vee 1 \right) + \frac{p}{s} \right),
\]
where $\|\Mx\Sigma\| = 1$ since $\Mx\Sigma = \I_r$.
\end{proof}

Now we go back to the proof of Lemma~\ref{lemma:taylor}
\begin{proof}[Lemma~\ref{lemma:taylor}]
For simplicity, we write $\E Y = \Mx\Sigma$, $Y_i = \z_i \z_i^\intercal$ and $\Mx\Sigma_s = \bar Y$. Here $\bar Y$ is unbiased estimator with $\E \bar Y = \E Y$.

By taking the Taylor expansion of $f$ at point $\E Y$, we have
\[
f(\bar{Y}) - f(\E Y) = \sum_{j=1}^m \frac{1}{j!} f^{(j)}(\E Y)[\bar Y - \E Y, \cdots, \bar Y - \E Y] + R,
\]
with the remainder term $R$ such that
\[
|R| \lesssim_m \|f^{(m+1}\|_{L_\infty} \|\bar Y - \E Y\|^{m+1}.
\]

By taking the expectation on both side, we have
\[
\E f(\bar{Y}) - f(\E Y) = \sum_{j=2}^{m} \frac{\E f^{(j)}(\E Y)[\bar Y - \E Y, \cdots, \bar Y - \E Y] }{j!} + \E R
\]
with remainder term satisfying that
\[
\E |R| \lesssim_m \|f^{(m+1}\|_{L_\infty} \E \|\bar Y - \E Y\|^{m+1}.
\]
It's sufficient to check the multi-linear form
\begin{align*}
& \E f^{(j)}(\E Y)[\bar Y - \E Y, \cdots, \bar Y - \E Y]\\
& =
s^{-j} \sum_{1\le i_1, \cdots, i_j \le s} \E f^{(j)}(\E Y)[ Y_{i_1} - \E Y, \cdots, Y_{i_j} - \E Y].
\end{align*}
If at least one of the indexes $i_1, \cdots, i_j$ only appear once, by the independence and $\E Y_i - \E Y = 0$
\[
\E f^{(j)}(\E Y)[ Y_{i_1} - \E Y, \cdots, Y_{i_j} - \E Y] = 0
\]
Then we only need to consider the case where every indexes appear at least twice. 
Suppose the index $i_j$ appear $k_j$ times.
It's sufficient to consider 
\[
M_{k_1,\cdots, k_m} = \E f^{(j)}(\E Y)[ Y_{1} - \E Y, \overset{k_1}\cdots, Y_{1} - \E Y, \cdots, Y_{m} - \E Y, \overset{k_m}\cdots, Y_{m} - \E Y]
\]
since $Y_1, \cdots, Y_s$ are identically distributed. Here $m \le j$ denote the number of distinct indexes.
Let
\[
B_m = \sum_{k_1 + \cdots, k_m = j, k_t \ge 2, t = 1, \cdots, m} \frac{j!}{k_1!\cdots k_m!} M_{k_1, \cdots, k_m},
\]
then 
\[
\E f^{(j)}(\E Y)[\bar Y - \E Y, \cdots, \bar Y - \E Y]\ = \sum_{1 \le m \le \frac{j}{2}} \frac{\binom{s}{m}}{s^{j}} B_m
\]
Now write $\binom{s}{m} = \sum_{i=1}^{m}c_{m,i}s^i$, as a consequence,
\[
\E f^{(j)}(\E Y)[\bar Y - \E Y, \cdots, \bar Y - \E Y]\ = \sum_{1 \le i \le \frac{j}{2}} s^{i-j}\sum_{i \le m \le \frac{j}{2}}c_{m,i} B_m 
\]
Note here $B_m$ only depends on $f$ and the distribution and $c_{m,i}$ is a constant, so both are independent to the sketch size.
Finally, combine all the multi-linear form, one can conclude the lemma.
\end{proof}

\section{Experiments}\label{appx:experiment}
We evaluate bias reduction across sketching methods and sampling distributions, compare iterative methods with larger single sketches, contrast empirical and parametric bootstrap, and examine bias decay as the sketch size increases.

\subsection{Experimental setup}
\label{appx:exp-setup}
\paragraph{Datasets and preprocessing.}
We conduct experiments on three datasets,
Ionosphere\footnote{\url{https://archive.ics.uci.edu/dataset/52/ionosphere}},
Abalone\footnote{\url{https://archive.ics.uci.edu/dataset/1/abalone}},
and California Housing\footnote{\url{https://scikit-learn.org/stable/datasets/real_world.html\#california-housing-dataset}}.
The dataset dimensions and preprocessing steps are summarized in Table~\ref{tab:datasets}. For California Housing, the eight features are standardized to have zero mean and unit variance before adding the intercept. The intercept is also ridge-penalized. We perform the experiments on an M2 MacBook.

\begin{table}[t]
\centering
\caption{Dataset dimensions and preprocessing.
Dimensions are reported after preprocessing.
All three design matrices have full column rank.}
\label{tab:datasets}
\small
\setlength{\tabcolsep}{4pt}
\renewcommand{\arraystretch}{1.15}
\begin{tabular}{@{}lrrp{3.5cm}cc@{}}
\toprule
Dataset & $n$ & $d$ & Feature selection
& Standardized & Intercept \\
\midrule
Ionosphere
& 351 & 33
& Remove the identically zero second feature
& No & No \\

Abalone
& 4177 & 7
& Retain the seven numerical features; exclude sex
& No & No \\

California Housing
& 20640 & 9
& Retain all eight original features
& Yes & Yes \\
\bottomrule
\end{tabular}
\end{table}

\paragraph{Estimators and implementation.}
We consider three settings: the classic sketch with the original response $\b$, the classic sketch with the projected response $\U\U^\intercal\b$, and the Hessian sketch with the original response. We do not separately test the projected-response Hessian sketch, since projection leaves $\A^\intercal\b$ and hence the estimator unchanged.

Single denotes the regular sketched estimator. PB simulates the Gaussian bootstrap recursion described in Section~\ref{sec:bootstrap}, whereas EB is empirical bootstrap, an empirical row-resampling variant. At each EB transition, we sample $s$ rows uniformly with replacement from the current sketch. For the classic estimator, the sketched features and response are resampled jointly; for the Hessian estimator, the exact right-hand side $\A^\intercal\b$ remains unchanged. Our theoretical analysis does not cover EB. The suffix $k \in\{2,3\}$ denotes the number of terms. Each simulated bootstrap estimator uses one recursive path with $k-1$ transitions. Single and bootstrap use sketch size $s$.

LA-2 and LA-3 denote linear aggregation of 2 and 3 estimators, such that sizes are chosen as $(s,\lfloor s/2\rfloor)$ and $(s,\lfloor s/2\rfloor,\lfloor s/3\rfloor)$, respectively, with coefficients defined by~\eqref{eq:coefficient}.

\paragraph{Evaluation protocol and metrics.}
We set $\gamma=0.3$ throughout. Each condition uses 20 independent repetitions. Unless stated otherwise, each repetition contains 10000 estimator draws.

Let $\x^*$ denote the exact ridge-regression solution and $\overline{\x}$ the sample mean estimate of $\E\hat{\x}$. The error metrics are evaluated at this sample mean.
Define the ridge objective as
\[
F_\gamma(\x;\b)
=
\|\A\x-\b\|_2^2+\gamma\|\x\|_2^2.
\]
We report the relative bias and relative objective gap:
\begin{align*}
\mathrm{RelBias}
&=
\frac{\|\overline{\x}-\x^*\|_\infty}
     {\|\x^*\|_\infty},
\\
\mathrm{RelObjGap}
&=
\frac{F_\gamma(\overline{\x};\b)-F_\gamma(\x^*;\b)}
     {F_\gamma(\mathbf{0};\b)-F_\gamma(\x^*;\b)}.
\end{align*}
The relative bias measures the discrepancy between the estimated mean and the exact solution, normalized by the size of the exact solution. The relative objective gap measures the excess ridge objective at the estimated mean, normalized by the objective improvement from the zero vector to the exact solution. Runtime is reported in microseconds per estimator as the median across the 20 repetitions. It excludes data preprocessing, response projection.

\subsection{Bias reduction across sketching methods}
\label{appx:exp_bias_reduction}
We compare Single, EB-2, EB-3, LA-2, and LA-3 on all three datasets and in all three estimator--response settings, using Gaussian, uniform, and leverage-score sampling. We use $s=256$ for Ionosphere and Abalone and $s=512$ for California Housing, keeping these sizes fixed across sampling distributions. This comparison fixes the largest component sketch size rather than the total computational budget.

\input{appendix_experiment_tables.tex}

\paragraph{Results.}
Tables~\ref{tab:app-primary-gaussian}--\ref{tab:app-primary-leverage} show that correction can substantially reduce both relative bias and relative objective gap. We highlight the smallest reported value among the tested methods in the tables. At the reported sketch sizes, all four iterative estimators improve both relative error and relative objective gap under Gaussian sketching for the projected-response classic sketch estimator and the Hessian estimator on all three datasets. The same improvement is observed for the projected-response classic sketch estimator under leverage-score sampling. For the classic estimator with the original response, improvements under Gaussian and leverage-score sampling are less consistent. Under uniform sampling with the original response, all four iterative estimators reduce the relative objective gap for both classic and Hessian sketches on all three datasets. Relative bias also decreases.

Using three terms does not consistently outperform using two. For example, LA-2 and LA-3 have nearly identical point estimates in the uniform Hessian experiment on Abalone. The smallest reported bias and objective gap can also correspond to different methods: for the uniform Hessian sketch on California Housing, EB-3 gives the smallest relative bias, whereas LA-3 gives the smallest relative objective gap. Overall, the results illustrate setting-dependent benefits of debiasing at additional computational cost.

In summary, our experiments show that iterative correction can effectively reduce sketching bias. Our theoretical results provide upper bounds rather than exact characterizations of the bias, and these bounds need not be tight in a particular practical setting. Although increasing the correction order yields a faster asymptotic decay rate with respect to the sketch size, the implicit constants in our bounds also depend on the correction order. Consequently, at a fixed sketch size, a higher correction order does not necessarily give a smaller upper bound or a smaller actual bias. 

\subsection{Iterative methods versus larger single sketches}
\label{appx:exp-larger-single}
We compare EB-2, EB-3, LA-2, and LA-3 with larger single sketches on Abalone, using classic Gaussian sketching and the projected response $\U\U^\intercal\b$. The maximum component sketch size is $s\in\{64,128,256\}$.

For each iterative method, the corresponding single-sketch size $s_{\mathrm{S}}$ equals the sum of its component sizes. For EB, this sum includes rows reused through resampling. Thus, the comparison matches the total number of rows entering the component solves.

Unlike the other experiments, this comparison uses 2000 draws per repetition.

\input{appendix_larger_single_sketch_table.tex}

\paragraph{Results.}
Table~\ref{tab:app-larger-single} shows that iterative bootstrap and linear aggregation achieve smaller relative bias and relative objective gap than their corresponding larger single sketches in all 12 tested configurations. The iterative methods retain only $7.2\%$--$47.4\%$ of the single-sketch relative bias and $0.34\%$--$10.46\%$ of its relative objective gap, while requiring $0.99$--$1.30$ times its measured runtime. Thus, in this setting, using several smaller component sketches for iterative debiasing can be more effective than simply increasing the size of a single sketch, at comparable measured runtime.

\subsection{Empirical versus parametric bootstrap}
\label{appx:exp-bootstrap-comparison}

We compare EB-2, PB-2, EB-3, and PB-3 with the single-sketch baseline on Abalone, using Gaussian sketches of size $s=256$ in all three estimator--response settings. 

\input{appendix_bootstrap_comparison_table.tex}

\paragraph{Results.}
Both bootstrap procedures substantially reduce the reported relative bias and relative objective gap for the projected-response classic sketch and the Hessian sketch. Neither bootstrap procedure uniformly outperforms the other across settings and metrics. PB requires more runtime in the current implementation, which additionally constructs a covariance factor and generates Gaussian samples at each transition. EB is relatively cheaper in computation since it resamples existing rows.

\subsection{Bias decay with increasing sketch size}
\include{bias_decay_nine_panels}
\input{mean_variance_nine_panels.tex}
We examine how relative bias changes with sketch size under Gaussian sketching on all three datasets and in all three estimator--response settings. We compare Single, EB-2, PB-2, EB-3, PB-3, LA-2, and LA-3. The sketch sizes double from $16$ to $2048$ for Abalone, from $64$ to $2048$ for Ionosphere, and from $32$ to $2048$ for California Housing. The trace of sample covariance for sketch solution is also reported.

\paragraph{Results.}
Figure~\ref{fig:bias-decay-nine-panels} shows that debiasing can substantially reduce bias and produce a faster observed decrease over parts of the tested sketch-size range.
This behavior is most apparent for the projected-response classic
sketch and the Hessian sketch. For the classic sketch with the original response, the separation between the single-sketch and debiasing methods is less clear.

Some bias decay curves decrease slowly at larger sketch sizes. The high order bias no longer dominant in the error while estimator variance limits the observed rate of decay in this regime. 

\paragraph{Effect of estimator variance.}
Bias reduction can come at the cost of increased estimator variance, measured here by the trace of the covariance matrix. For example, at $s=2048$, the implemented two-term bootstrap estimators have approximately twice the variance of the single sketch, while three-term bootstrap and two-term linear aggregation have approximately six times its variance, as shown in Figure~\ref{fig:mean-variance-nine-panels}. This increased variance makes small residual biases harder to estimate and can obscure their decay.

For the single classic Gaussian sketch, the contribution from $\b_\perp$ has zero expectation but adds variance relative to the projected-response estimator. Comparing the two response settings illustrates how this additional random error can obscure the estimator performance. The effect can be more pronounced for iterative methods, whose weighted combinations can amplify the random terms involving $\b_\perp$.


\bibliographystyle{abbrvnat}

\clearpage
\newpage
\input{checklist.tex}

\end{document}

%% file: appendix_experiment_tables.tex
\begin{table}[htbp]
\centering
\caption{Debiasing comparison under Gaussian sketching.}
\label{tab:app-primary-gaussian}
\begingroup
\small
\setlength{\tabcolsep}{3pt}
\renewcommand{\arraystretch}{1.12}
\resizebox{\linewidth}{!}{%
\begin{tabular}{@{}llrrrrr@{}}
\toprule
Dataset & Metric & Single & EB-2 & EB-3 & LA-2 & LA-3\\
\midrule
\multicolumn{7}{@{}l}{\textit{Classical sketch, original response $\b$}}\\
Ionosphere ($s=256$) & $\mathrm{RelBias}$ & $2.060{\times}10^{-3}$ & $\boldsymbol{1.424{\times}10^{-3}}$ & $2.021{\times}10^{-3}$ & $2.476{\times}10^{-3}$ & $6.516{\times}10^{-3}$\\
 & $\mathrm{RelObjGap}$ & $2.417{\times}10^{-7}$ & $\boldsymbol{2.393{\times}10^{-7}}$ & $7.283{\times}10^{-7}$ & $6.473{\times}10^{-7}$ & $4.601{\times}10^{-6}$\\
 & Time ($\mu$s) & 186.4 & 272.7 & 307.4 & 310.5 & 327.0\\[2pt]
Abalone ($s=256$) & $\mathrm{RelBias}$ & $1.829{\times}10^{-3}$ & $\boldsymbol{6.848{\times}10^{-4}}$ & $2.748{\times}10^{-3}$ & $1.377{\times}10^{-3}$ & $2.533{\times}10^{-3}$\\
 & $\mathrm{RelObjGap}$ & $3.749{\times}10^{-8}$ & $\boldsymbol{7.472{\times}10^{-9}}$ & $6.320{\times}10^{-8}$ & $3.055{\times}10^{-8}$ & $5.424{\times}10^{-8}$\\
 & Time ($\mu$s) & 18.1 & 26.9 & 35.8 & 27.5 & 34.0\\[2pt]
\shortstack[l]{California Housing\\($s=512$)} & $\mathrm{RelBias}$ & $1.417{\times}10^{-4}$ & $\boldsymbol{1.326{\times}10^{-4}}$ & $3.929{\times}10^{-4}$ & $7.417{\times}10^{-4}$ & $1.267{\times}10^{-3}$\\
 & $\mathrm{RelObjGap}$ & $\boldsymbol{9.410{\times}10^{-9}}$ & $1.403{\times}10^{-8}$ & $6.478{\times}10^{-8}$ & $3.517{\times}10^{-8}$ & $3.075{\times}10^{-7}$\\
 & Time ($\mu$s) & 40.7 & 55.4 & 69.9 & 56.1 & 69.9\\
\midrule
\multicolumn{7}{@{}l}{\textit{Classical sketch, projected response $\U\U^\intercal\b$}}\\
Ionosphere ($s=256$) & $\mathrm{RelBias}$ & $1.865{\times}10^{-3}$ & $2.912{\times}10^{-4}$ & $\boldsymbol{9.414{\times}10^{-5}}$ & $6.402{\times}10^{-4}$ & $3.393{\times}10^{-4}$\\
 & $\mathrm{RelObjGap}$ & $1.620{\times}10^{-7}$ & $3.907{\times}10^{-9}$ & $\boldsymbol{4.929{\times}10^{-10}}$ & $1.935{\times}10^{-8}$ & $7.415{\times}10^{-9}$\\
 & Time ($\mu$s) & 98.2 & 372.9 & 665.3 & 164.1 & 189.8\\[2pt]
Abalone ($s=256$) & $\mathrm{RelBias}$ & $1.801{\times}10^{-3}$ & $\boldsymbol{3.934{\times}10^{-5}}$ & $8.469{\times}10^{-5}$ & $3.964{\times}10^{-5}$ & $1.061{\times}10^{-4}$\\
 & $\mathrm{RelObjGap}$ & $2.829{\times}10^{-8}$ & $\boldsymbol{7.857{\times}10^{-12}}$ & $3.840{\times}10^{-11}$ & $2.997{\times}10^{-11}$ & $7.764{\times}10^{-11}$\\
 & Time ($\mu$s) & 16.6 & 31.6 & 45.9 & 25.8 & 32.7\\[2pt]
\shortstack[l]{California Housing\\($s=512$)} & $\mathrm{RelBias}$ & $2.185{\times}10^{-6}$ & $\boldsymbol{2.514{\times}10^{-8}}$ & $1.610{\times}10^{-7}$ & $1.308{\times}10^{-7}$ & $1.913{\times}10^{-7}$\\
 & $\mathrm{RelObjGap}$ & $5.712{\times}10^{-13}$ & $\boldsymbol{2.322{\times}10^{-16}}$ & $1.410{\times}10^{-15}$ & $2.107{\times}10^{-15}$ & $2.737{\times}10^{-15}$\\
 & Time ($\mu$s) & 37.0 & 81.7 & 125.0 & 55.3 & 69.1\\
\midrule
\multicolumn{7}{@{}l}{\textit{Hessian sketch, original response $\b$}}\\
Ionosphere ($s=256$) & $\mathrm{RelBias}$ & $1.526{\times}10^{-1}$ & $2.482{\times}10^{-2}$ & $\boldsymbol{1.205{\times}10^{-2}}$ & $5.143{\times}10^{-2}$ & $5.144{\times}10^{-2}$\\
 & $\mathrm{RelObjGap}$ & $2.291{\times}10^{-2}$ & $6.382{\times}10^{-4}$ & $\boldsymbol{1.043{\times}10^{-4}}$ & $2.863{\times}10^{-3}$ & $1.149{\times}10^{-3}$\\
 & Time ($\mu$s) & 202.4 & 248.7 & 309.4 & 298.7 & 379.0\\[2pt]
Abalone ($s=256$) & $\mathrm{RelBias}$ & $3.002{\times}10^{-2}$ & $7.246{\times}10^{-3}$ & $1.128{\times}10^{-2}$ & $\boldsymbol{5.722{\times}10^{-3}}$ & $2.163{\times}10^{-2}$\\
 & $\mathrm{RelObjGap}$ & $8.613{\times}10^{-4}$ & $\boldsymbol{5.743{\times}10^{-7}}$ & $7.313{\times}10^{-7}$ & $3.707{\times}10^{-6}$ & $6.886{\times}10^{-6}$\\
 & Time ($\mu$s) & 23.3 & 37.3 & 50.9 & 39.0 & 49.0\\[2pt]
\shortstack[l]{California Housing\\($s=512$)} & $\mathrm{RelBias}$ & $1.951{\times}10^{-2}$ & $\boldsymbol{4.759{\times}10^{-4}}$ & $1.672{\times}10^{-3}$ & $9.496{\times}10^{-4}$ & $9.174{\times}10^{-3}$\\
 & $\mathrm{RelObjGap}$ & $3.805{\times}10^{-4}$ & $\boldsymbol{1.686{\times}10^{-7}}$ & $5.171{\times}10^{-7}$ & $7.121{\times}10^{-7}$ & $7.911{\times}10^{-6}$\\
 & Time ($\mu$s) & 78.9 & 134.1 & 187.4 & 131.9 & 185.5\\
\bottomrule
\end{tabular}%
}
\endgroup
\end{table}

\begin{table}[htbp]
\centering
\caption{Debiasing comparison under uniform sampling.}
\label{tab:app-primary-uniform}
\begingroup
\small
\setlength{\tabcolsep}{3pt}
\renewcommand{\arraystretch}{1.12}
\resizebox{\linewidth}{!}{%
\begin{tabular}{@{}llrrrrr@{}}
\toprule
Dataset & Metric & Single & EB-2 & EB-3 & LA-2 & LA-3\\
\midrule
\multicolumn{7}{@{}l}{\textit{Classical sketch, original response $\b$}}\\
Ionosphere ($s=256$) & $\mathrm{RelBias}$ & $9.135{\times}10^{-2}$ & $5.580{\times}10^{-2}$ & $\boldsymbol{2.539{\times}10^{-2}}$ & $1.009{\times}10^{-1}$ & $7.290{\times}10^{-2}$\\
 & $\mathrm{RelObjGap}$ & $1.871{\times}10^{-3}$ & $2.610{\times}10^{-4}$ & $\boldsymbol{6.119{\times}10^{-5}}$ & $5.220{\times}10^{-4}$ & $5.106{\times}10^{-4}$\\
 & Time ($\mu$s) & 39.4 & 286.5 & 534.4 & 65.3 & 85.0\\[2pt]
Abalone ($s=256$) & $\mathrm{RelBias}$ & $4.234{\times}10^{-1}$ & $3.957{\times}10^{-1}$ & $3.831{\times}10^{-1}$ & $3.847{\times}10^{-1}$ & $\boldsymbol{3.581{\times}10^{-1}}$\\
 & $\mathrm{RelObjGap}$ & $4.188{\times}10^{-4}$ & $3.580{\times}10^{-4}$ & $3.350{\times}10^{-4}$ & $3.379{\times}10^{-4}$ & $\boldsymbol{2.933{\times}10^{-4}}$\\
 & Time ($\mu$s) & 10.9 & 26.2 & 41.3 & 16.9 & 20.9\\[2pt]
\shortstack[l]{California Housing\\($s=512$)} & $\mathrm{RelBias}$ & $4.041{\times}10^{-2}$ & $2.905{\times}10^{-2}$ & $2.646{\times}10^{-2}$ & $2.656{\times}10^{-2}$ & $\boldsymbol{2.403{\times}10^{-2}}$\\
 & $\mathrm{RelObjGap}$ & $2.955{\times}10^{-4}$ & $1.292{\times}10^{-4}$ & $1.043{\times}10^{-4}$ & $1.021{\times}10^{-4}$ & $\boldsymbol{8.264{\times}10^{-5}}$\\
 & Time ($\mu$s) & 19.7 & 63.7 & 107.7 & 31.1 & 38.1\\
\midrule
\multicolumn{7}{@{}l}{\textit{Classical sketch, projected response $\U\U^\intercal\b$}}\\
Ionosphere ($s=256$) & $\mathrm{RelBias}$ & $5.484{\times}10^{-3}$ & $2.007{\times}10^{-3}$ & $\boldsymbol{7.538{\times}10^{-4}}$ & $5.818{\times}10^{-3}$ & $3.391{\times}10^{-3}$\\
 & $\mathrm{RelObjGap}$ & $1.537{\times}10^{-6}$ & $2.493{\times}10^{-7}$ & $\boldsymbol{5.086{\times}10^{-8}}$ & $2.075{\times}10^{-6}$ & $1.071{\times}10^{-6}$\\
 & Time ($\mu$s) & 43.0 & 289.6 & 541.5 & 73.7 & 91.7\\[2pt]
Abalone ($s=256$) & $\mathrm{RelBias}$ & $5.151{\times}10^{-2}$ & $4.455{\times}10^{-2}$ & $4.292{\times}10^{-2}$ & $4.312{\times}10^{-2}$ & $\boldsymbol{3.983{\times}10^{-2}}$\\
 & $\mathrm{RelObjGap}$ & $6.454{\times}10^{-6}$ & $4.544{\times}10^{-6}$ & $4.210{\times}10^{-6}$ & $4.249{\times}10^{-6}$ & $\boldsymbol{3.625{\times}10^{-6}}$\\
 & Time ($\mu$s) & 9.2 & 26.8 & 42.0 & 20.0 & 23.3\\[2pt]
\shortstack[l]{California Housing\\($s=512$)} & $\mathrm{RelBias}$ & $2.537{\times}10^{-5}$ & $6.314{\times}10^{-6}$ & $4.683{\times}10^{-6}$ & $2.073{\times}10^{-6}$ & $\boldsymbol{2.009{\times}10^{-6}}$\\
 & $\mathrm{RelObjGap}$ & $2.213{\times}10^{-11}$ & $2.427{\times}10^{-12}$ & $1.623{\times}10^{-12}$ & $8.484{\times}10^{-13}$ & $\boldsymbol{6.209{\times}10^{-13}}$\\
 & Time ($\mu$s) & 22.4 & 65.4 & 108.8 & 33.0 & 42.9\\
\midrule
\multicolumn{7}{@{}l}{\textit{Hessian sketch, original response $\b$}}\\
Ionosphere ($s=256$) & $\mathrm{RelBias}$ & $1.560{\times}10^{-1}$ & $5.392{\times}10^{-2}$ & $\boldsymbol{3.386{\times}10^{-2}}$ & $1.199{\times}10^{-1}$ & $1.528{\times}10^{-1}$\\
 & $\mathrm{RelObjGap}$ & $7.979{\times}10^{-3}$ & $8.044{\times}10^{-4}$ & $\boldsymbol{2.048{\times}10^{-4}}$ & $4.180{\times}10^{-3}$ & $4.110{\times}10^{-3}$\\
 & Time ($\mu$s) & 40.7 & 284.0 & 525.7 & 68.8 & 89.2\\[2pt]
Abalone ($s=256$) & $\mathrm{RelBias}$ & $7.765{\times}10^{-1}$ & $6.723{\times}10^{-1}$ & $6.641{\times}10^{-1}$ & $\boldsymbol{6.565{\times}10^{-1}}$ & $6.596{\times}10^{-1}$\\
 & $\mathrm{RelObjGap}$ & $2.803{\times}10^{-3}$ & $1.036{\times}10^{-3}$ & $1.009{\times}10^{-3}$ & $\boldsymbol{1.001{\times}10^{-3}}$ & $1.003{\times}10^{-3}$\\
 & Time ($\mu$s) & 15.7 & 34.7 & 53.0 & 26.8 & 35.8\\[2pt]
\shortstack[l]{California Housing\\($s=512$)} & $\mathrm{RelBias}$ & $1.542{\times}10^{-1}$ & $2.218{\times}10^{-2}$ & $\boldsymbol{2.099{\times}10^{-2}}$ & $3.770{\times}10^{-2}$ & $2.148{\times}10^{-2}$\\
 & $\mathrm{RelObjGap}$ & $1.665{\times}10^{-3}$ & $8.383{\times}10^{-5}$ & $7.257{\times}10^{-5}$ & $9.279{\times}10^{-5}$ & $\boldsymbol{6.058{\times}10^{-5}}$\\
 & Time ($\mu$s) & 56.8 & 136.3 & 214.3 & 105.8 & 151.7\\
\bottomrule
\end{tabular}%
}
\endgroup
\end{table}

\begin{table}[htbp]
\centering
\caption{Debiasing comparison under leverage-score sampling.}
\label{tab:app-primary-leverage}
\begingroup
\small
\setlength{\tabcolsep}{3pt}
\renewcommand{\arraystretch}{1.12}
\resizebox{\linewidth}{!}{%
\begin{tabular}{@{}llrrrrr@{}}
\toprule
Dataset & Metric & Single & EB-2 & EB-3 & LA-2 & LA-3\\
\midrule
\multicolumn{7}{@{}l}{\textit{Classical sketch, original response $\b$}}\\
Ionosphere ($s=256$) & $\mathrm{RelBias}$ & $2.423{\times}10^{-3}$ & $\boldsymbol{1.971{\times}10^{-3}}$ & $3.945{\times}10^{-3}$ & $2.928{\times}10^{-3}$ & $6.100{\times}10^{-3}$\\
 & $\mathrm{RelObjGap}$ & $\boldsymbol{4.138{\times}10^{-7}}$ & $5.201{\times}10^{-7}$ & $1.046{\times}10^{-6}$ & $1.381{\times}10^{-6}$ & $3.470{\times}10^{-6}$\\
 & Time ($\mu$s) & 75.4 & 116.4 & 157.3 & 107.0 & 129.1\\[2pt]
Abalone ($s=256$) & $\mathrm{RelBias}$ & $1.175{\times}10^{-3}$ & $\boldsymbol{1.007{\times}10^{-3}}$ & $2.508{\times}10^{-3}$ & $2.324{\times}10^{-3}$ & $2.810{\times}10^{-3}$\\
 & $\mathrm{RelObjGap}$ & $1.000{\times}10^{-8}$ & $\boldsymbol{7.386{\times}10^{-9}}$ & $2.829{\times}10^{-8}$ & $3.658{\times}10^{-8}$ & $1.846{\times}10^{-7}$\\
 & Time ($\mu$s) & 25.1 & 34.7 & 43.6 & 41.5 & 57.5\\[2pt]
\shortstack[l]{California Housing\\($s=512$)} & $\mathrm{RelBias}$ & $3.054{\times}10^{-4}$ & $\boldsymbol{1.495{\times}10^{-4}}$ & $3.564{\times}10^{-4}$ & $4.352{\times}10^{-4}$ & $1.295{\times}10^{-3}$\\
 & $\mathrm{RelObjGap}$ & $\boldsymbol{1.064{\times}10^{-8}}$ & $1.865{\times}10^{-8}$ & $4.148{\times}10^{-8}$ & $5.946{\times}10^{-8}$ & $2.924{\times}10^{-7}$\\
 & Time ($\mu$s) & 81.6 & 96.3 & 109.5 & 127.8 & 173.6\\
\midrule
\multicolumn{7}{@{}l}{\textit{Classical sketch, projected response $\U\U^\intercal\b$}}\\
Ionosphere ($s=256$) & $\mathrm{RelBias}$ & $1.806{\times}10^{-3}$ & $4.011{\times}10^{-4}$ & $\boldsymbol{1.191{\times}10^{-4}}$ & $8.028{\times}10^{-4}$ & $6.190{\times}10^{-4}$\\
 & $\mathrm{RelObjGap}$ & $1.535{\times}10^{-7}$ & $8.434{\times}10^{-9}$ & $\boldsymbol{1.244{\times}10^{-9}}$ & $3.303{\times}10^{-8}$ & $2.292{\times}10^{-8}$\\
 & Time ($\mu$s) & 51.5 & 299.5 & 548.7 & 85.6 & 110.4\\[2pt]
Abalone ($s=256$) & $\mathrm{RelBias}$ & $1.372{\times}10^{-3}$ & $5.835{\times}10^{-5}$ & $\boldsymbol{5.265{\times}10^{-5}}$ & $1.006{\times}10^{-4}$ & $1.034{\times}10^{-4}$\\
 & $\mathrm{RelObjGap}$ & $1.615{\times}10^{-8}$ & $3.616{\times}10^{-11}$ & $\boldsymbol{1.484{\times}10^{-11}}$ & $5.749{\times}10^{-11}$ & $1.873{\times}10^{-10}$\\
 & Time ($\mu$s) & 24.3 & 39.3 & 53.9 & 40.4 & 53.0\\[2pt]
\shortstack[l]{California Housing\\($s=512$)} & $\mathrm{RelBias}$ & $1.778{\times}10^{-6}$ & $7.874{\times}10^{-8}$ & $\boldsymbol{2.696{\times}10^{-8}}$ & $8.688{\times}10^{-8}$ & $2.166{\times}10^{-7}$\\
 & $\mathrm{RelObjGap}$ & $3.707{\times}10^{-13}$ & $9.512{\times}10^{-16}$ & $6.590{\times}10^{-16}$ & $\boldsymbol{4.197{\times}10^{-16}}$ & $5.792{\times}10^{-15}$\\
 & Time ($\mu$s) & 77.3 & 120.6 & 164.4 & 127.5 & 169.4\\
\midrule
\multicolumn{7}{@{}l}{\textit{Hessian sketch, original response $\b$}}\\
Ionosphere ($s=256$) & $\mathrm{RelBias}$ & $1.459{\times}10^{-1}$ & $3.773{\times}10^{-2}$ & $\boldsymbol{2.061{\times}10^{-2}}$ & $6.593{\times}10^{-2}$ & $7.910{\times}10^{-2}$\\
 & $\mathrm{RelObjGap}$ & $2.267{\times}10^{-2}$ & $1.993{\times}10^{-3}$ & $\boldsymbol{3.649{\times}10^{-4}}$ & $6.977{\times}10^{-3}$ & $9.914{\times}10^{-3}$\\
 & Time ($\mu$s) & 73.5 & 116.6 & 177.3 & 115.1 & 148.2\\[2pt]
Abalone ($s=256$) & $\mathrm{RelBias}$ & $1.905{\times}10^{-2}$ & $7.162{\times}10^{-3}$ & $9.384{\times}10^{-3}$ & $\boldsymbol{5.245{\times}10^{-3}}$ & $2.739{\times}10^{-2}$\\
 & $\mathrm{RelObjGap}$ & $4.819{\times}10^{-4}$ & $\boldsymbol{7.070{\times}10^{-7}}$ & $1.371{\times}10^{-6}$ & $1.098{\times}10^{-6}$ & $5.990{\times}10^{-6}$\\
 & Time ($\mu$s) & 31.6 & 45.3 & 60.0 & 52.3 & 71.1\\[2pt]
\shortstack[l]{California Housing\\($s=512$)} & $\mathrm{RelBias}$ & $1.589{\times}10^{-2}$ & $\boldsymbol{9.219{\times}10^{-4}}$ & $1.115{\times}10^{-3}$ & $1.651{\times}10^{-3}$ & $3.609{\times}10^{-3}$\\
 & $\mathrm{RelObjGap}$ & $2.562{\times}10^{-4}$ & $2.778{\times}10^{-7}$ & $\boldsymbol{2.776{\times}10^{-7}}$ & $6.571{\times}10^{-7}$ & $1.107{\times}10^{-6}$\\
 & Time ($\mu$s) & 119.8 & 173.6 & 227.1 & 206.8 & 291.9\\
\bottomrule
\end{tabular}%
}
\endgroup
\end{table}

%% file: appendix_larger_single_sketch_table.tex
\begin{table}[htbp]
\centering
\caption{Iterative bootstrap and linear aggregation versus larger single sketches on Abalone.
The corresponding single-sketch size $s_{\mathrm{S}}$ equals the sum of the component sizes.
Bold marks the smaller relative bias and relative objective gap in each row.}
\label{tab:app-larger-single}
\begingroup
\small
\setlength{\tabcolsep}{2pt}
\renewcommand{\arraystretch}{1.16}
\begin{tabular}{@{}llrrr@{\hspace{8pt}}rrrr@{}}
\toprule
Method & Component sizes
& \multicolumn{3}{c}{Iterative method}
& \multicolumn{4}{c}{Corresponding single sketch}\\
\cmidrule(lr){3-5}\cmidrule(lr){6-9}
& & $\mathrm{RelBias}$ & $\mathrm{RelObjGap}$ & \shortstack{Time\\($\mu$s)}
& $s_{\mathrm{S}}$ & $\mathrm{RelBias}$ & $\mathrm{RelObjGap}$ & \shortstack{Time\\($\mu$s)}\\
\midrule
EB-2 & $(64,64)$ & $\boldsymbol{2.843{\times}10^{-4}}$ & $\boldsymbol{1.302{\times}10^{-9}}$ & 9.17 & $128$ & $3.794{\times}10^{-3}$ & $1.207{\times}10^{-7}$ & 8.62\\
EB-3 & $(64,64,64)$ & $\boldsymbol{1.767{\times}10^{-4}}$ & $\boldsymbol{1.750{\times}10^{-10}}$ & 12.86 & $192$ & $2.451{\times}10^{-3}$ & $5.173{\times}10^{-8}$ & 12.01\\
LA-2 & $(64,32)$ & $\boldsymbol{1.468{\times}10^{-3}}$ & $\boldsymbol{2.230{\times}10^{-8}}$ & 8.71 & $96$ & $5.017{\times}10^{-3}$ & $2.162{\times}10^{-7}$ & 6.71\\
LA-3 & $(64,32,21)$ & $\boldsymbol{7.259{\times}10^{-4}}$ & $\boldsymbol{2.608{\times}10^{-9}}$ & 10.76 & $117$ & $3.954{\times}10^{-3}$ & $1.355{\times}10^{-7}$ & 8.41\\
\midrule
EB-2 & $(128,128)$ & $\boldsymbol{1.511{\times}10^{-4}}$ & $\boldsymbol{1.884{\times}10^{-10}}$ & 15.78 & $256$ & $1.842{\times}10^{-3}$ & $2.833{\times}10^{-8}$ & 15.12\\
EB-3 & $(128,128,128)$ & $\boldsymbol{4.662{\times}10^{-4}}$ & $\boldsymbol{7.913{\times}10^{-10}}$ & 22.82 & $384$ & $1.220{\times}10^{-3}$ & $1.241{\times}10^{-8}$ & 21.83\\
LA-2 & $(128,64)$ & $\boldsymbol{5.653{\times}10^{-4}}$ & $\boldsymbol{1.977{\times}10^{-9}}$ & 14.22 & $192$ & $2.451{\times}10^{-3}$ & $5.173{\times}10^{-8}$ & 12.01\\
LA-3 & $(128,64,42)$ & $\boldsymbol{7.100{\times}10^{-4}}$ & $\boldsymbol{3.469{\times}10^{-9}}$ & 17.82 & $234$ & $2.008{\times}10^{-3}$ & $3.317{\times}10^{-8}$ & 14.13\\
\midrule
EB-2 & $(256,256)$ & $\boldsymbol{1.196{\times}10^{-4}}$ & $\boldsymbol{1.241{\times}10^{-10}}$ & 30.32 & $512$ & $9.339{\times}10^{-4}$ & $7.289{\times}10^{-9}$ & 28.19\\
EB-3 & $(256,256,256)$ & $\boldsymbol{2.772{\times}10^{-4}}$ & $\boldsymbol{1.587{\times}10^{-10}}$ & 44.82 & $768$ & $5.845{\times}10^{-4}$ & $3.023{\times}10^{-9}$ & 45.50\\
LA-2 & $(256,128)$ & $\boldsymbol{1.764{\times}10^{-4}}$ & $\boldsymbol{9.255{\times}10^{-11}}$ & 24.23 & $384$ & $1.220{\times}10^{-3}$ & $1.241{\times}10^{-8}$ & 21.83\\
LA-3 & $(256,128,85)$ & $\boldsymbol{2.472{\times}10^{-4}}$ & $\boldsymbol{7.785{\times}10^{-10}}$ & 30.56 & $469$ & $1.040{\times}10^{-3}$ & $8.914{\times}10^{-9}$ & 26.18\\
\bottomrule
\end{tabular}
\endgroup
\end{table}

%% file: appendix_bootstrap_comparison_table.tex
\begin{table}[htbp]
\centering
\caption{Empirical versus parametric bootstrap on Abalone.
Bold marks the smallest relative bias and relative objective gap in each row.}
\label{tab:app-bootstrap-variants}
\begingroup
\small
\setlength{\tabcolsep}{4pt}
\renewcommand{\arraystretch}{1.15}
\begin{tabular}{@{}lrrrrr@{}}
\toprule
Metric & Single & EB-2 & PB-2 & EB-3 & PB-3\\
\midrule
\multicolumn{6}{@{}l}{\textit{Classical sketch, original response $\b$}}\\
$\mathrm{RelBias}$ & $2.200{\times}10^{-3}$ & $\boldsymbol{1.130{\times}10^{-3}}$ & $1.497{\times}10^{-3}$ & $1.173{\times}10^{-3}$ & $2.024{\times}10^{-3}$\\
$\mathrm{RelObjGap}$ & $4.018{\times}10^{-8}$ & $\boldsymbol{9.072{\times}10^{-9}}$ & $1.288{\times}10^{-8}$ & $1.382{\times}10^{-8}$ & $4.184{\times}10^{-8}$\\
Time ($\mu$s) & 15.77 & 30.63 & 55.57 & 44.91 & 95.07\\
\midrule
\multicolumn{6}{@{}l}{\textit{Classical sketch, projected response $\U\U^\intercal\b$}}\\
$\mathrm{RelBias}$ & $1.826{\times}10^{-3}$ & $\boldsymbol{3.373{\times}10^{-5}}$ & $3.766{\times}10^{-5}$ & $4.102{\times}10^{-5}$ & $5.335{\times}10^{-5}$\\
$\mathrm{RelObjGap}$ & $2.882{\times}10^{-8}$ & $1.298{\times}10^{-11}$ & $\boldsymbol{1.032{\times}10^{-11}}$ & $2.353{\times}10^{-11}$ & $1.977{\times}10^{-11}$\\
Time ($\mu$s) & 18.12 & 36.19 & 63.51 & 52.33 & 114.27\\
\midrule
\multicolumn{6}{@{}l}{\textit{Hessian sketch, original response $\b$}}\\
$\mathrm{RelBias}$ & $2.727{\times}10^{-2}$ & $8.812{\times}10^{-3}$ & $\boldsymbol{3.297{\times}10^{-3}}$ & $1.509{\times}10^{-2}$ & $7.547{\times}10^{-3}$\\
$\mathrm{RelObjGap}$ & $8.534{\times}10^{-4}$ & $7.583{\times}10^{-7}$ & $1.273{\times}10^{-6}$ & $1.726{\times}10^{-6}$ & $\boldsymbol{4.556{\times}10^{-7}}$\\
Time ($\mu$s) & 23.05 & 43.57 & 73.75 & 63.25 & 122.99\\
\bottomrule
\end{tabular}
\endgroup
\end{table}

%% file: bias_decay_nine_panels.tex
\begin{figure}[t]
\centering
\begingroup

\definecolor{bnpColor1}{HTML}{151E2B}
\definecolor{bnpColor2}{HTML}{8B4BB4}
\definecolor{bnpColor3}{HTML}{1671C4}
\definecolor{bnpColor4}{HTML}{C64F91}
\definecolor{bnpColor5}{HTML}{164795}
\definecolor{bnpColor6}{HTML}{D47414}
\definecolor{bnpColor7}{HTML}{16836E}
\begin{tikzpicture}[mark size=1.25pt,line width=0.6pt]
\begin{groupplot}[
    group style={group size=3 by 3,horizontal sep=0.078\linewidth,vertical sep=0.155\linewidth},
    scale only axis,
    width=0.238\linewidth,
    height=0.205\linewidth,
    xmode=log,log basis x=2,ymode=log,
    xlabel={Sketch size $s$},
    grid=major,
    major grid style={solid,black!12,line width=0.25pt},
    axis line style={black!55,line width=0.45pt},
    tick style={black!55},tick align=outside,
    xtick pos=left,ytick pos=left,minor tick style={draw=none},
    tick label style={font=\scriptsize},
    label style={font=\scriptsize},
    title style={font=\scriptsize,align=center,yshift=1pt},
    minor x tick num=0,minor y tick num=0,
    clip=true,
]

\nextgroupplot[title={(a) Abalone\\Classic (projected response)},
xmin=13.559322033898306,xmax=2416.64,ymin=1e-6,ymax=1e-1,
xtick={16,64,256,2048},xticklabels={16,64,256,2048},ytick={1e-6,1e-4,1e-2,1e-1},
ylabel={Relative bias},
legend to name=biasDecayNinePanelLegend,legend columns=7,legend cell align=left,
legend style={draw=none,fill=none,font=\scriptsize,
    /tikz/every even column/.append style={column sep=0.12cm}},
]
\addplot[bnpColor1,solid,mark=*,mark options={solid,fill=bnpColor1},forget plot] coordinates {
    (16, 0.04202328108566408)
    (32, 0.017268486807148785)
    (64, 0.00783301189931209)
    (128, 0.0037225562819577927)
    (256, 0.001801428235250558)
    (512, 0.0009094273401073375)
    (1024, 0.00044938739530812774)
    (2048, 0.00022356937829354158)
};
\addplot[bnpColor2,solid,mark=square*,mark options={solid,fill=bnpColor2},forget plot] coordinates {
    (16, 0.011396626152443084)
    (32, 0.0019995047089265447)
    (64, 0.00044078402792911665)
    (128, 7.71139141256224e-5)
    (256, 3.934189013364523e-5)
    (512, 3.598832359363981e-5)
    (1024, 1.754710144502347e-5)
    (2048, 8.613680035738657e-6)
};
\addplot[bnpColor3,solid,mark=triangle*,mark options={solid,fill=bnpColor3},forget plot] coordinates {
    (16, 0.005173547334491393)
    (32, 0.0016169075804496572)
    (64, 0.0004672763285800798)
    (128, 0.00010513243381149544)
    (256, 3.7662887117620126e-5)
    (512, 3.3858444980131064e-5)
    (1024, 1.4552678767939464e-5)
    (2048, 4.52474190502744e-6)
};
\addplot[bnpColor4,solid,mark=diamond*,mark options={solid,fill=bnpColor4},forget plot] coordinates {
    (16, 0.005867445711736704)
    (32, 0.001041181956928233)
    (64, 0.000250759601526556)
    (128, 6.7744670812729e-5)
    (256, 8.469264651367825e-5)
    (512, 4.447504195235976e-5)
    (1024, 4.431409801655263e-5)
    (2048, 2.1708015199906278e-5)
};
\addplot[bnpColor5,solid,mark=pentagon*,mark options={solid,fill=bnpColor5},forget plot] coordinates {
    (16, 0.0048509579920828225)
    (32, 0.00041323489288542804)
    (64, 5.9746615512845655e-5)
    (128, 3.82165771075599e-5)
    (256, 5.334824025983271e-5)
    (512, 2.3770620957039624e-5)
    (1024, 2.2735915952255684e-5)
    (2048, 9.203412967267868e-6)
};
\addplot[bnpColor6,solid,mark=triangle*,mark options={solid,fill=bnpColor6,rotate=180},forget plot] coordinates {
    (16, 0.04010637203684714)
    (32, 0.007469918726027615)
    (64, 0.0014624431885918502)
    (128, 0.00041768925911289873)
    (256, 3.963585520777416e-5)
    (512, 8.472201936274824e-5)
    (1024, 1.8730910299924447e-5)
    (2048, 2.914487298819598e-5)
};
\addplot[bnpColor7,solid,mark=otimes*,mark options={solid,fill=bnpColor7},forget plot] coordinates {
    (16, 0.02918149468703767)
    (32, 0.003213443378911869)
    (64, 0.0007561208239020173)
    (128, 0.0002522980603486548)
    (256, 0.0001061198460319389)
    (512, 0.0001594471886796178)
    (1024, 6.37275806138669e-5)
    (2048, 6.312210117777517e-5)
};
\addlegendimage{bnpColor1,solid,mark=*,mark options={solid,fill=bnpColor1}}
\addlegendentry{Single}
\addlegendimage{bnpColor2,solid,mark=square*,mark options={solid,fill=bnpColor2}}
\addlegendentry{EB-2}
\addlegendimage{bnpColor3,solid,mark=triangle*,mark options={solid,fill=bnpColor3}}
\addlegendentry{PB-2}
\addlegendimage{bnpColor4,solid,mark=diamond*,mark options={solid,fill=bnpColor4}}
\addlegendentry{EB-3}
\addlegendimage{bnpColor5,solid,mark=pentagon*,mark options={solid,fill=bnpColor5}}
\addlegendentry{PB-3}
\addlegendimage{bnpColor6,solid,mark=triangle*,mark options={solid,fill=bnpColor6,rotate=180}}
\addlegendentry{LA-2}
\addlegendimage{bnpColor7,solid,mark=otimes*,mark options={solid,fill=bnpColor7}}
\addlegendentry{LA-3}

\nextgroupplot[title={(b) Ionosphere\\Classic (projected response)},
xmin=54.237288135593225,xmax=2416.64,ymin=1e-6,ymax=1e0,
xtick={64,256,2048},xticklabels={64,256,2048},ytick={1e-6,1e-4,1e-2,1e0},
]
\addplot[bnpColor1,solid,mark=*,mark options={solid,fill=bnpColor1},forget plot] coordinates {
    (64, 0.012981930071506382)
    (128, 0.004348936747827867)
    (256, 0.0018646138227775473)
    (512, 0.0008727556368735427)
    (1024, 0.00042422836191255116)
    (2048, 0.00020835704536195348)
};
\addplot[bnpColor2,solid,mark=square*,mark options={solid,fill=bnpColor2},forget plot] coordinates {
    (64, 0.024799308855409607)
    (128, 0.001762278874648487)
    (256, 0.000291153604341045)
    (512, 5.6220652550665833e-5)
    (1024, 1.2566586527438007e-5)
    (2048, 2.555833423889174e-6)
};
\addplot[bnpColor3,solid,mark=triangle*,mark options={solid,fill=bnpColor3},forget plot] coordinates {
    (64, 0.009442953102124924)
    (128, 0.001297062781045487)
    (256, 0.00024467546828811957)
    (512, 4.896127019059509e-5)
    (1024, 1.6991672839889053e-5)
    (2048, 3.3709261274961095e-6)
};
\addplot[bnpColor4,solid,mark=diamond*,mark options={solid,fill=bnpColor4},forget plot] coordinates {
    (64, 0.0419219840545407)
    (128, 0.0015744564584159494)
    (256, 9.414062749009994e-5)
    (512, 2.3127602685315727e-5)
    (1024, 1.1376461409804311e-5)
    (2048, 4.3562251118752265e-6)
};
\addplot[bnpColor5,solid,mark=pentagon*,mark options={solid,fill=bnpColor5},forget plot] coordinates {
    (64, 0.0017080409242523542)
    (128, 0.00026381365897858926)
    (256, 3.241717119718504e-5)
    (512, 1.0089227359224471e-5)
    (1024, 8.470264560306585e-6)
    (2048, 6.736801632399515e-6)
};
\addplot[bnpColor6,solid,mark=triangle*,mark options={solid,fill=bnpColor6,rotate=180},forget plot] coordinates {
    (64, 0.08452913819453448)
    (128, 0.004270465968935602)
    (256, 0.0006402417419338077)
    (512, 0.00013033983455367702)
    (1024, 3.2598734728463456e-5)
    (2048, 9.129650365972306e-6)
};
\addplot[bnpColor7,solid,mark=otimes*,mark options={solid,fill=bnpColor7},forget plot] coordinates {
    (64, 0.07186905069808724)
    (128, 0.010544364505062843)
    (256, 0.00033930301408178007)
    (512, 3.9299418465570996e-5)
    (1024, 2.3294769327775643e-5)
    (2048, 1.3517844526124525e-5)
};

\nextgroupplot[title={(c) California Housing\\Classic (projected response)},
xmin=27.118644067796613,xmax=2416.64,ymin=1e-8,ymax=1e-2,
xtick={32,128,512,2048},xticklabels={32,128,512,2048},ytick={1e-8,1e-6,1e-4,1e-2},
]
\addplot[bnpColor1,solid,mark=*,mark options={solid,fill=bnpColor1},forget plot] coordinates {
    (32, 4.969210970611343e-5)
    (64, 2.0016138974526463e-5)
    (128, 9.190526986517238e-6)
    (256, 4.353156358439956e-6)
    (512, 2.184783726848672e-6)
    (1024, 1.0814701022273105e-6)
    (2048, 5.218462633344565e-7)
};
\addplot[bnpColor2,solid,mark=square*,mark options={solid,fill=bnpColor2},forget plot] coordinates {
    (32, 3.152475204331869e-5)
    (64, 4.068882120642618e-6)
    (128, 6.362525915636647e-7)
    (256, 1.341254024337451e-7)
    (512, 2.5137073297445783e-8)
    (1024, 2.4512413861411958e-8)
    (2048, 3.4709522142973316e-8)
};
\addplot[bnpColor3,solid,mark=triangle*,mark options={solid,fill=bnpColor3},forget plot] coordinates {
    (32, 2.227082457783926e-5)
    (64, 3.7362115989208733e-6)
    (128, 6.862110292728454e-7)
    (256, 1.6795566421766515e-7)
    (512, 8.941663041444664e-8)
    (1024, 4.3182547293903864e-8)
    (2048, 5.152201476237653e-8)
};
\addplot[bnpColor4,solid,mark=diamond*,mark options={solid,fill=bnpColor4},forget plot] coordinates {
    (32, 9.04581803007114e-5)
    (64, 2.2294939872093373e-6)
    (128, 3.4956039089187937e-7)
    (256, 4.592796538898349e-8)
    (512, 1.610260125884828e-7)
    (1024, 3.0209019708910996e-8)
    (2048, 2.7823271143211756e-8)
};
\addplot[bnpColor5,solid,mark=pentagon*,mark options={solid,fill=bnpColor5},forget plot] coordinates {
    (32, 8.81825443063126e-6)
    (64, 7.721799763024582e-7)
    (128, 1.165479960214747e-7)
    (256, 4.6161034237236746e-8)
    (512, 4.261999361744568e-8)
    (1024, 4.044505831960684e-8)
    (2048, 5.862694109047299e-8)
};
\addplot[bnpColor6,solid,mark=triangle*,mark options={solid,fill=bnpColor6,rotate=180},forget plot] coordinates {
    (32, 8.179207042263976e-5)
    (64, 9.387987107657e-6)
    (128, 1.5632656721885776e-6)
    (256, 5.020216796103869e-7)
    (512, 1.30787754761114e-7)
    (1024, 4.317584010233345e-8)
    (2048, 2.7790842076157126e-8)
};
\addplot[bnpColor7,solid,mark=otimes*,mark options={solid,fill=bnpColor7},forget plot] coordinates {
    (32, 0.0014479316933802757)
    (64, 9.431589589027258e-6)
    (128, 5.318088899989502e-7)
    (256, 2.5240086659252234e-7)
    (512, 1.912609512383782e-7)
    (1024, 1.1172709178159358e-7)
    (2048, 6.100706552463152e-8)
};

\nextgroupplot[title={(d) Abalone\\Classic (original response)},
xmin=13.559322033898306,xmax=2416.64,ymin=1e-4,ymax=1e-1,
xtick={16,64,256,2048},xticklabels={16,64,256,2048},ytick={1e-4,1e-3,1e-2,1e-1},
ylabel={Relative bias},
]
\addplot[bnpColor1,solid,mark=*,mark options={solid,fill=bnpColor1},forget plot] coordinates {
    (16, 0.042824727990448105)
    (32, 0.016801283196631128)
    (64, 0.008829706829225153)
    (128, 0.003712201991863946)
    (256, 0.0018287534322182792)
    (512, 0.0015128040366064737)
    (1024, 0.0007354814691930929)
    (2048, 0.0005070931565378976)
};
\addplot[bnpColor2,solid,mark=square*,mark options={solid,fill=bnpColor2},forget plot] coordinates {
    (16, 0.014858455787801999)
    (32, 0.00874259332421039)
    (64, 0.002909927976239509)
    (128, 0.0027125620995460107)
    (256, 0.0006848130851923605)
    (512, 0.0014546697264613793)
    (1024, 0.0009023107675000979)
    (2048, 0.0004963325140649154)
};
\addplot[bnpColor3,solid,mark=triangle*,mark options={solid,fill=bnpColor3},forget plot] coordinates {
    (16, 0.00964418779387547)
    (32, 0.00419448231720891)
    (64, 0.0010245983495684955)
    (128, 0.0015285593071659644)
    (256, 0.0014966323127183036)
    (512, 0.00094540711787893)
    (1024, 0.000715511265004858)
    (2048, 0.00044568809457831883)
};
\addplot[bnpColor4,solid,mark=diamond*,mark options={solid,fill=bnpColor4},forget plot] coordinates {
    (16, 0.005838162588316977)
    (32, 0.007835341949405803)
    (64, 0.001962655709680536)
    (128, 0.003687916562005316)
    (256, 0.0027477997565725225)
    (512, 0.0010515030231983992)
    (1024, 0.0006770310860858552)
    (2048, 0.0006100956401561964)
};
\addplot[bnpColor5,solid,mark=pentagon*,mark options={solid,fill=bnpColor5},forget plot] coordinates {
    (16, 0.011491189901354637)
    (32, 0.007773814446535504)
    (64, 0.005909105476943275)
    (128, 0.0026731453955724763)
    (256, 0.002023884174861391)
    (512, 0.0014868687416819204)
    (1024, 0.001290573060868592)
    (2048, 0.001022225138492592)
};
\addplot[bnpColor6,solid,mark=triangle*,mark options={solid,fill=bnpColor6,rotate=180},forget plot] coordinates {
    (16, 0.04015120145946963)
    (32, 0.012062166533759713)
    (64, 0.005846481657125188)
    (128, 0.0016901735092082013)
    (256, 0.001376666718586442)
    (512, 0.0016903713184856022)
    (1024, 0.0011419830659472276)
    (2048, 0.0007641141739234033)
};
\addplot[bnpColor7,solid,mark=otimes*,mark options={solid,fill=bnpColor7},forget plot] coordinates {
    (16, 0.021783001753674442)
    (32, 0.008666927798332164)
    (64, 0.008145375811833905)
    (128, 0.004663510408189211)
    (256, 0.0025332759986019344)
    (512, 0.0032968981123135182)
    (1024, 0.0017120006901768423)
    (2048, 0.0012282753865436888)
};

\nextgroupplot[title={(e) Ionosphere\\Classic (original response)},
xmin=54.237288135593225,xmax=2416.64,ymin=1e-4,ymax=1e-1,
xtick={64,256,2048},xticklabels={64,256,2048},ytick={1e-4,1e-3,1e-2,1e-1},
]
\addplot[bnpColor1,solid,mark=*,mark options={solid,fill=bnpColor1},forget plot] coordinates {
    (64, 0.012833761418695423)
    (128, 0.004690573575363257)
    (256, 0.0020595803992884522)
    (512, 0.0011378610682447096)
    (1024, 0.0006057522294174993)
    (2048, 0.0005343592534425586)
};
\addplot[bnpColor2,solid,mark=square*,mark options={solid,fill=bnpColor2},forget plot] coordinates {
    (64, 0.023974585628853483)
    (128, 0.0024411545125279112)
    (256, 0.0014236128878604816)
    (512, 0.0010908134875199477)
    (1024, 0.0006533311132258295)
    (2048, 0.0005112283397718847)
};
\addplot[bnpColor3,solid,mark=triangle*,mark options={solid,fill=bnpColor3},forget plot] coordinates {
    (64, 0.00886066407227844)
    (128, 0.0026423229688310565)
    (256, 0.00134411784703877)
    (512, 0.001099831770559563)
    (1024, 0.0005517423031511001)
    (2048, 0.00047718503542208175)
};
\addplot[bnpColor4,solid,mark=diamond*,mark options={solid,fill=bnpColor4},forget plot] coordinates {
    (64, 0.04424257312597434)
    (128, 0.004103327787790651)
    (256, 0.0020213347177573)
    (512, 0.0016708016215304588)
    (1024, 0.0011480712844985786)
    (2048, 0.0007276096742501772)
};
\addplot[bnpColor5,solid,mark=pentagon*,mark options={solid,fill=bnpColor5},forget plot] coordinates {
    (64, 0.005667720298797353)
    (128, 0.004204929627262491)
    (256, 0.002048234789246949)
    (512, 0.001689262111298134)
    (1024, 0.0015035056304835953)
    (2048, 0.0010065585642788882)
};
\addplot[bnpColor6,solid,mark=triangle*,mark options={solid,fill=bnpColor6,rotate=180},forget plot] coordinates {
    (64, 0.08079445913392812)
    (128, 0.004496154148059513)
    (256, 0.0024755555461774773)
    (512, 0.0014863052916945986)
    (1024, 0.0009087260564771794)
    (2048, 0.0007932341067451648)
};
\addplot[bnpColor7,solid,mark=otimes*,mark options={solid,fill=bnpColor7},forget plot] coordinates {
    (64, 0.07261360201523706)
    (128, 0.0133511642318263)
    (256, 0.006515585060870052)
    (512, 0.0036146461018135103)
    (1024, 0.002970676982870579)
    (2048, 0.0021116725585900937)
};

\nextgroupplot[title={(f) California Housing\\Classic (original response)},
xmin=27.118644067796613,xmax=2416.64,ymin=1e-5,ymax=1e-2,
xtick={32,128,512,2048},xticklabels={32,128,512,2048},ytick={1e-5,1e-4,1e-3,1e-2},
]
\addplot[bnpColor1,solid,mark=*,mark options={solid,fill=bnpColor1},forget plot] coordinates {
    (32, 0.0011795605446531577)
    (64, 0.0007123764718836093)
    (128, 0.0003151676808194252)
    (256, 0.000517575366981992)
    (512, 0.0002222981669380684)
    (1024, 0.0002444739375656365)
    (2048, 0.00023325674176668382)
};
\addplot[bnpColor2,solid,mark=square*,mark options={solid,fill=bnpColor2},forget plot] coordinates {
    (32, 0.0013215862176013044)
    (64, 0.0013655204721837472)
    (128, 0.000817146507137028)
    (256, 0.00023143852684581395)
    (512, 0.0003515496537995692)
    (1024, 0.00013632269445327397)
    (2048, 0.0001633939014849373)
};
\addplot[bnpColor3,solid,mark=triangle*,mark options={solid,fill=bnpColor3},forget plot] coordinates {
    (32, 0.0010864397222218821)
    (64, 0.0005471109677417849)
    (128, 0.0008292575429184753)
    (256, 0.00043047218484609166)
    (512, 0.00040483997346657134)
    (1024, 7.924971931786635e-5)
    (2048, 0.0001048478680796198)
};
\addplot[bnpColor4,solid,mark=diamond*,mark options={solid,fill=bnpColor4},forget plot] coordinates {
    (32, 0.006727192365733183)
    (64, 0.0011047382352765167)
    (128, 0.0013210489619426896)
    (256, 0.0002600770078388653)
    (512, 0.00043886552887736744)
    (1024, 0.0001944956433211413)
    (2048, 0.0001751073451202703)
};
\addplot[bnpColor5,solid,mark=pentagon*,mark options={solid,fill=bnpColor5},forget plot] coordinates {
    (32, 0.0028077859091056945)
    (64, 0.0007104437403013618)
    (128, 0.0010162084870269192)
    (256, 0.00031834035119780897)
    (512, 0.0005002530130487511)
    (1024, 0.00016752689025192227)
    (2048, 0.0001789028680001207)
};
\addplot[bnpColor6,solid,mark=triangle*,mark options={solid,fill=bnpColor6,rotate=180},forget plot] coordinates {
    (32, 0.0033726755091377)
    (64, 0.0006947187574000446)
    (128, 0.00037011490573975333)
    (256, 0.0014841578731592728)
    (512, 0.001187326698414585)
    (1024, 0.00020046025712746723)
    (2048, 0.00024952994066444425)
};
\addplot[bnpColor7,solid,mark=otimes*,mark options={solid,fill=bnpColor7},forget plot] coordinates {
    (32, 0.0031831753618049804)
    (64, 0.006018894254488112)
    (128, 0.0007562151969519867)
    (256, 0.0017027906602249383)
    (512, 0.0006705030923554488)
    (1024, 0.0009050575934544939)
    (2048, 0.0006314707878457953)
};

\nextgroupplot[title={(g) Abalone\\Hessian (original response)},
xmin=13.559322033898306,xmax=2416.64,ymin=1e-3,ymax=1e1,
xtick={16,64,256,2048},xticklabels={16,64,256,2048},ytick={1e-3,1e-1,1e1},
ylabel={Relative bias},
]
\addplot[bnpColor1,solid,mark=*,mark options={solid,fill=bnpColor1},forget plot] coordinates {
    (16, 0.6943773566636074)
    (32, 0.2660625312996277)
    (64, 0.12093800556190573)
    (128, 0.05768923787170696)
    (256, 0.030018030576633097)
    (512, 0.013668471284785377)
    (1024, 0.0073837154991598)
    (2048, 0.0029614887205581526)
};
\addplot[bnpColor2,solid,mark=square*,mark options={solid,fill=bnpColor2},forget plot] coordinates {
    (16, 0.42214372308783293)
    (32, 0.06930759803224225)
    (64, 0.013277409212180456)
    (128, 0.014723601200397568)
    (256, 0.007245887662860467)
    (512, 0.003467846365139092)
    (1024, 0.0022516843973819805)
    (2048, 0.001727725801769151)
};
\addplot[bnpColor3,solid,mark=triangle*,mark options={solid,fill=bnpColor3},forget plot] coordinates {
    (16, 0.22981581414290306)
    (32, 0.04390339373948851)
    (64, 0.018423727462899372)
    (128, 0.013886552029330956)
    (256, 0.0032969309404853224)
    (512, 0.0035994564342700947)
    (1024, 0.0020127912237608452)
    (2048, 0.0019197183829185926)
};
\addplot[bnpColor4,solid,mark=diamond*,mark options={solid,fill=bnpColor4},forget plot] coordinates {
    (16, 0.49654897075010784)
    (32, 0.04458497393548327)
    (64, 0.03448511831454454)
    (128, 0.01876148757062377)
    (256, 0.011284701193269795)
    (512, 0.008011292777779764)
    (1024, 0.004634546912491204)
    (2048, 0.0015472532232992543)
};
\addplot[bnpColor5,solid,mark=pentagon*,mark options={solid,fill=bnpColor5},forget plot] coordinates {
    (16, 0.10796857905733843)
    (32, 0.027958054925039787)
    (64, 0.03848324909519264)
    (128, 0.008838435447553271)
    (256, 0.007546602444354489)
    (512, 0.00577127264903921)
    (1024, 0.0038820758350684476)
    (2048, 0.002789915051431397)
};
\addplot[bnpColor6,solid,mark=triangle*,mark options={solid,fill=bnpColor6,rotate=180},forget plot] coordinates {
    (16, 1.119693997534591)
    (32, 0.15244213343992505)
    (64, 0.03223854574214898)
    (128, 0.015833796075562238)
    (256, 0.0057216232788186265)
    (512, 0.01060002133835839)
    (1024, 0.005149448352078913)
    (2048, 0.005179273280578834)
};
\addplot[bnpColor7,solid,mark=otimes*,mark options={solid,fill=bnpColor7},forget plot] coordinates {
    (16, 1.9602032861399592)
    (32, 0.13077925849511762)
    (64, 0.06541426867073921)
    (128, 0.020458705125843867)
    (256, 0.02163102139773917)
    (512, 0.013127095879698462)
    (1024, 0.01112229262391642)
    (2048, 0.006575207686170614)
};

\nextgroupplot[title={(h) Ionosphere\\Hessian (original response)},
xmin=54.237288135593225,xmax=2416.64,ymin=1e-4,ymax=1e2,
xtick={64,256,2048},xticklabels={64,256,2048},ytick={1e-4,1e-2,1e0,1e2},
]
\addplot[bnpColor1,solid,mark=*,mark options={solid,fill=bnpColor1},forget plot] coordinates {
    (64, 1.080300693836193)
    (128, 0.35223785858557455)
    (256, 0.15258006471393099)
    (512, 0.07065311251861597)
    (1024, 0.034343905652416365)
    (2048, 0.01633533780341206)
};
\addplot[bnpColor2,solid,mark=square*,mark options={solid,fill=bnpColor2},forget plot] coordinates {
    (64, 2.670032974046082)
    (128, 0.15574470670035612)
    (256, 0.02482177835097104)
    (512, 0.005581456539495986)
    (1024, 0.0022425552236559187)
    (2048, 0.000869042083042875)
};
\addplot[bnpColor3,solid,mark=triangle*,mark options={solid,fill=bnpColor3},forget plot] coordinates {
    (64, 0.9826138866169938)
    (128, 0.1156763199127012)
    (256, 0.02072097490816348)
    (512, 0.004695584142431035)
    (1024, 0.001505697233890084)
    (2048, 0.0014197335225828894)
};
\addplot[bnpColor4,solid,mark=diamond*,mark options={solid,fill=bnpColor4},forget plot] coordinates {
    (64, 9.902729589255229)
    (128, 0.15740137064028714)
    (256, 0.012050303287458297)
    (512, 0.005858363308006176)
    (1024, 0.0038465741485195545)
    (2048, 0.0019822087621057445)
};
\addplot[bnpColor5,solid,mark=pentagon*,mark options={solid,fill=bnpColor5},forget plot] coordinates {
    (64, 0.4708168781490155)
    (128, 0.03641590345166325)
    (256, 0.006299513952533553)
    (512, 0.004727995237844262)
    (1024, 0.003372639199055999)
    (2048, 0.0015689328712525526)
};
\addplot[bnpColor6,solid,mark=triangle*,mark options={solid,fill=bnpColor6,rotate=180},forget plot] coordinates {
    (64, 10.870273873758503)
    (128, 0.37316484443952624)
    (256, 0.05142600816589277)
    (512, 0.01121513991751026)
    (1024, 0.0030689932665457775)
    (2048, 0.002171635918901307)
};
\addplot[bnpColor7,solid,mark=otimes*,mark options={solid,fill=bnpColor7},forget plot] coordinates {
    (64, 34.36385812310005)
    (128, 1.06953425701437)
    (256, 0.05143946657894919)
    (512, 0.01599333030293159)
    (1024, 0.007397682502627847)
    (2048, 0.004484723723424208)
};

\nextgroupplot[title={(i) California Housing\\Hessian (original response)},
xmin=27.118644067796613,xmax=2416.64,ymin=1e-4,ymax=1e2,
xtick={32,128,512,2048},xticklabels={32,128,512,2048},ytick={1e-4,1e-2,1e0,1e2},
]
\addplot[bnpColor1,solid,mark=*,mark options={solid,fill=bnpColor1},forget plot] coordinates {
    (32, 0.4539015322550914)
    (64, 0.18575963848924387)
    (128, 0.08519389810254285)
    (256, 0.04054201329343837)
    (512, 0.019514431348761036)
    (1024, 0.009671969155245937)
    (2048, 0.004851856285474812)
};
\addplot[bnpColor2,solid,mark=square*,mark options={solid,fill=bnpColor2},forget plot] coordinates {
    (32, 0.3014202472694357)
    (64, 0.03543526569070219)
    (128, 0.006293201856285196)
    (256, 0.0034924749416108875)
    (512, 0.00047593986741590683)
    (1024, 0.0011917845053544974)
    (2048, 0.0004955846931207703)
};
\addplot[bnpColor3,solid,mark=triangle*,mark options={solid,fill=bnpColor3},forget plot] coordinates {
    (32, 0.20628403034678774)
    (64, 0.03512068967050342)
    (128, 0.0070533213065863765)
    (256, 0.0011715700353180656)
    (512, 0.0012618775061364806)
    (1024, 0.0009414940977170431)
    (2048, 0.0005606295851304709)
};
\addplot[bnpColor4,solid,mark=diamond*,mark options={solid,fill=bnpColor4},forget plot] coordinates {
    (32, 1.0239428166960922)
    (64, 0.022920020577309404)
    (128, 0.0049686800681399464)
    (256, 0.0015209725513850507)
    (512, 0.001671510124106143)
    (1024, 0.0005491144230278612)
    (2048, 0.000473151185110953)
};
\addplot[bnpColor5,solid,mark=pentagon*,mark options={solid,fill=bnpColor5},forget plot] coordinates {
    (32, 0.09858216137378359)
    (64, 0.004926369730866262)
    (128, 0.0031214452776245063)
    (256, 0.0010959406917223668)
    (512, 0.0009071585407900115)
    (1024, 0.0009159445102547586)
    (2048, 0.0009418499125479468)
};
\addplot[bnpColor6,solid,mark=triangle*,mark options={solid,fill=bnpColor6,rotate=180},forget plot] coordinates {
    (32, 0.761735687600255)
    (64, 0.08487273532499028)
    (128, 0.014556662971524163)
    (256, 0.00449762902312176)
    (512, 0.0009495546523370776)
    (1024, 0.0011638044957299036)
    (2048, 0.0008038335483910622)
};
\addplot[bnpColor7,solid,mark=otimes*,mark options={solid,fill=bnpColor7},forget plot] coordinates {
    (32, 16.12184710456235)
    (64, 0.07582290898234952)
    (128, 0.0064841560525737835)
    (256, 0.0028529174475721984)
    (512, 0.009174433267022136)
    (1024, 0.0030475377954266507)
    (2048, 0.0016442121261231709)
};
\end{groupplot}
\end{tikzpicture}
\par\vspace{1mm}
\pgfplotslegendfromname{biasDecayNinePanelLegend}
\endgroup
\caption{Relative bias versus sketch size for Gaussian sketches across three datasets and three sketch settings.}
\label{fig:bias-decay-nine-panels}
\end{figure}
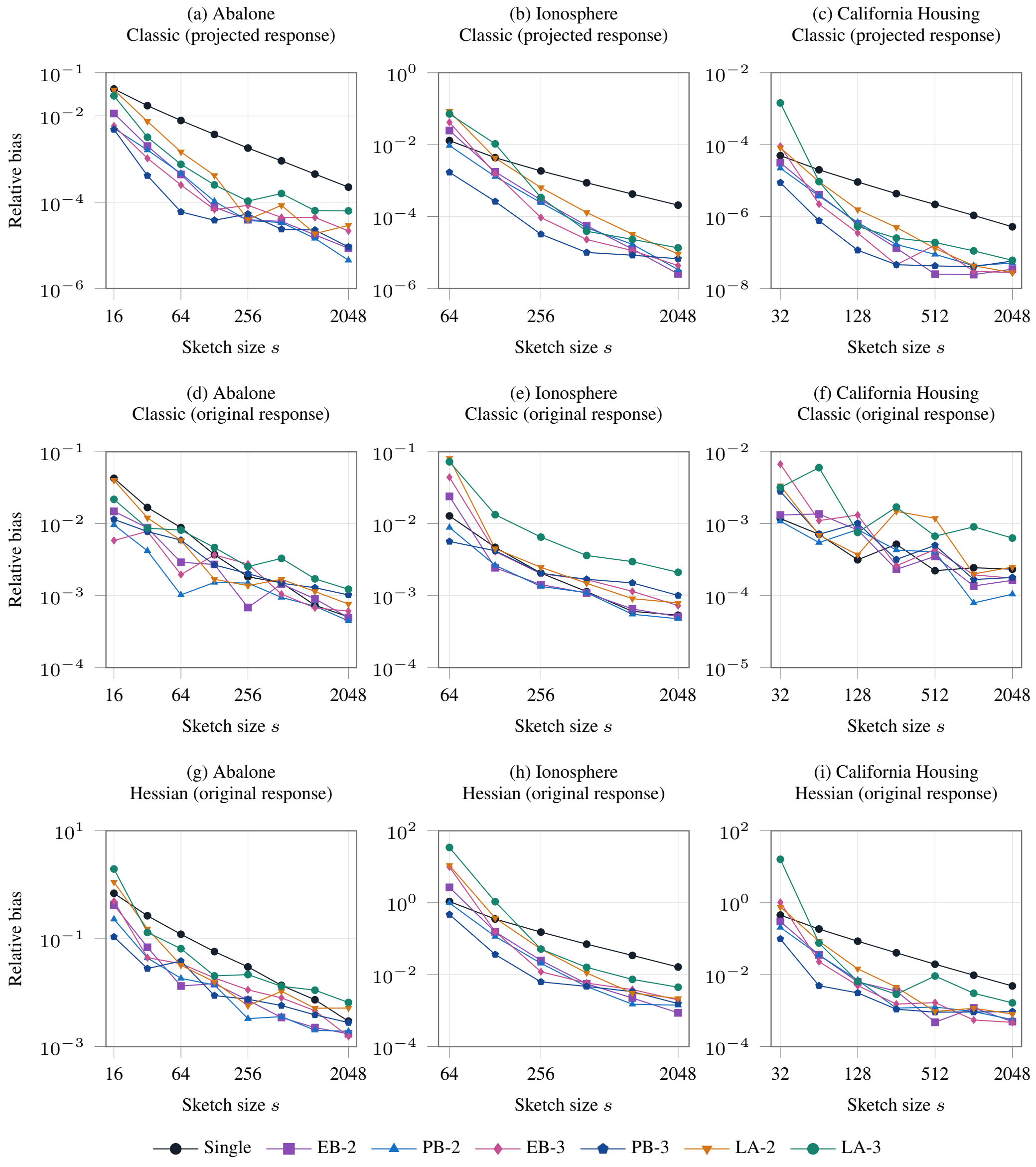

%% file: mean_variance_nine_panels.tex
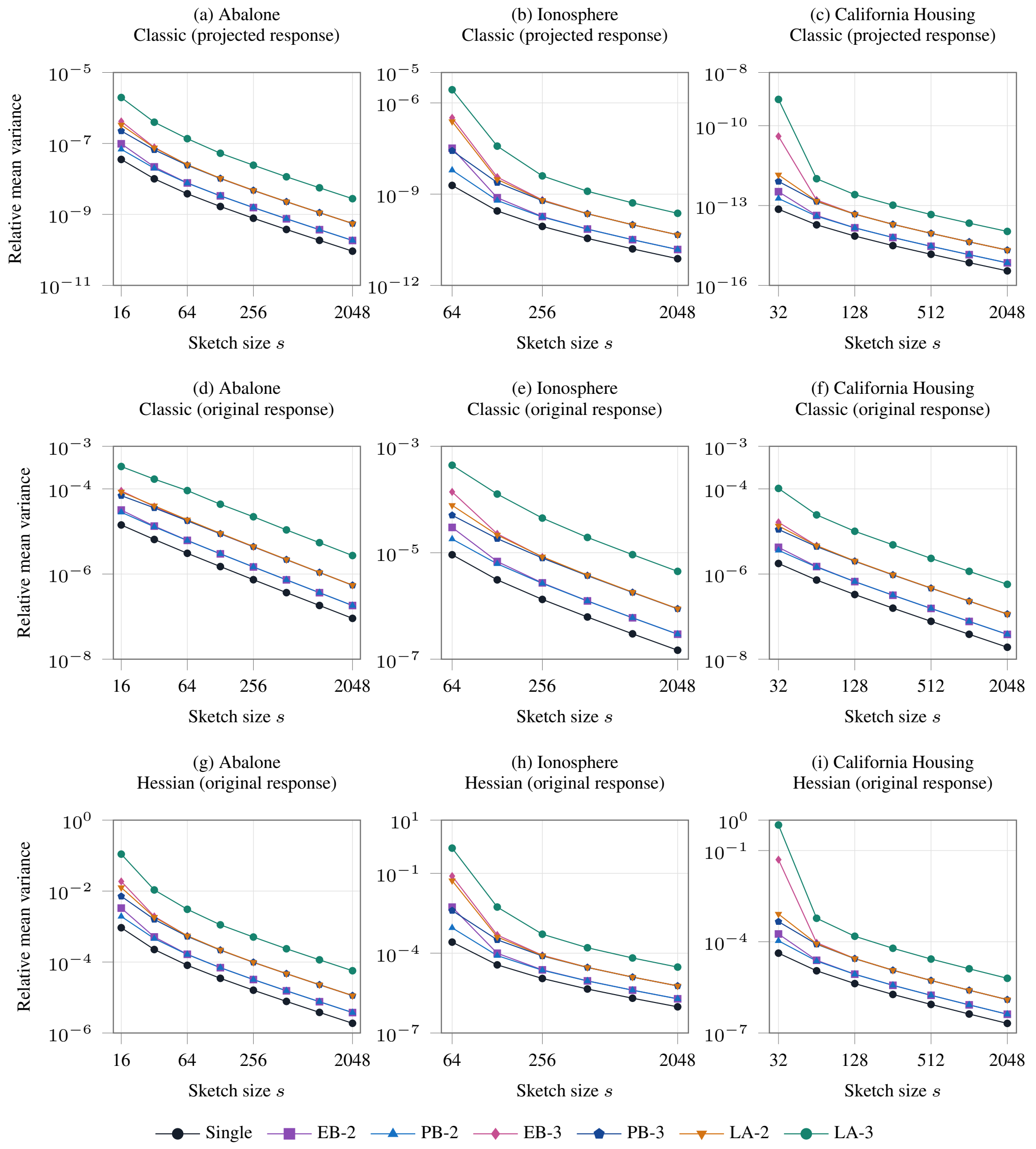
\begin{figure}[t]
\centering
\begingroup

\definecolor{mnpColor1}{HTML}{151E2B}
\definecolor{mnpColor2}{HTML}{8B4BB4}
\definecolor{mnpColor3}{HTML}{1671C4}
\definecolor{mnpColor4}{HTML}{C64F91}
\definecolor{mnpColor5}{HTML}{164795}
\definecolor{mnpColor6}{HTML}{D47414}
\definecolor{mnpColor7}{HTML}{16836E}
\begin{tikzpicture}[mark size=1.25pt,line width=0.6pt]
\begin{groupplot}[
    group style={group size=3 by 3,horizontal sep=0.078\linewidth,vertical sep=0.155\linewidth},
    scale only axis,
    width=0.238\linewidth,
    height=0.205\linewidth,
    xmode=log,log basis x=2,ymode=log,
    xlabel={Sketch size $s$},
    grid=major,
    major grid style={solid,black!12,line width=0.25pt},
    axis line style={black!55,line width=0.45pt},
    tick style={black!55},tick align=outside,
    xtick pos=left,ytick pos=left,minor tick style={draw=none},
    tick label style={font=\scriptsize},
    label style={font=\scriptsize},
    title style={font=\scriptsize,align=center,yshift=1pt},
    minor x tick num=0,minor y tick num=0,
    clip=true,
]

\nextgroupplot[title={(a) Abalone\\Classic (projected response)},
xmin=13.559322033898306,xmax=2416.64,ymin=1e-11,ymax=1e-5,
xtick={16,64,256,2048},xticklabels={16,64,256,2048},ytick={1e-11,1e-9,1e-7,1e-5},
ylabel={Relative mean variance},
legend to name=meanVarianceNinePanelLegend,legend columns=7,legend cell align=left,
legend style={draw=none,fill=none,font=\scriptsize,
    /tikz/every even column/.append style={column sep=0.12cm}},
]
\addplot[mnpColor1,solid,mark=*,mark options={solid,fill=mnpColor1},forget plot] coordinates {
    (16, 3.542435453175886e-8)
    (32, 1.0069464393728766e-8)
    (64, 3.819007607044946e-9)
    (128, 1.6712784757772848e-9)
    (256, 7.79023732819654e-10)
    (512, 3.764880104336536e-10)
    (1024, 1.8574733741387645e-10)
    (2048, 9.208517610101647e-11)
};
\addplot[mnpColor2,solid,mark=square*,mark options={solid,fill=mnpColor2},forget plot] coordinates {
    (16, 9.781992146743472e-8)
    (32, 2.187747402817956e-8)
    (64, 7.719940119781698e-9)
    (128, 3.336258559484738e-9)
    (256, 1.5566681926370334e-9)
    (512, 7.526230985535854e-10)
    (1024, 3.711754479008632e-10)
    (2048, 1.842404833381393e-10)
};
\addplot[mnpColor3,solid,mark=triangle*,mark options={solid,fill=mnpColor3},forget plot] coordinates {
    (16, 6.852149412211297e-8)
    (32, 2.0009531301918983e-8)
    (64, 7.60972428201881e-9)
    (128, 3.3297548678675122e-9)
    (256, 1.5597338692710945e-9)
    (512, 7.560056313512926e-10)
    (1024, 3.7162243978822524e-10)
    (2048, 1.8468993433210918e-10)
};
\addplot[mnpColor4,solid,mark=diamond*,mark options={solid,fill=mnpColor4},forget plot] coordinates {
    (16, 4.185655266502069e-7)
    (32, 7.77019423613664e-8)
    (64, 2.4825098193887185e-8)
    (128, 1.0282099888674454e-8)
    (256, 4.7098840647632755e-9)
    (512, 2.280480824986981e-9)
    (1024, 1.1155987535339093e-9)
    (2048, 5.550923687832155e-10)
};
\addplot[mnpColor5,solid,mark=pentagon*,mark options={solid,fill=mnpColor5},forget plot] coordinates {
    (16, 2.250315339275815e-7)
    (32, 6.622207584012902e-8)
    (64, 2.427561225641904e-8)
    (128, 1.0331098074308462e-8)
    (256, 4.765661976864677e-9)
    (512, 2.2768785393585973e-9)
    (1024, 1.1229393153506733e-9)
    (2048, 5.550112811810629e-10)
};
\addplot[mnpColor6,solid,mark=triangle*,mark options={solid,fill=mnpColor6,rotate=180},forget plot] coordinates {
    (16, 3.3326245065991453e-7)
    (32, 7.551353412769918e-8)
    (64, 2.5353505725668516e-8)
    (128, 1.0467600699853103e-8)
    (256, 4.773207223293906e-9)
    (512, 2.298798298840103e-9)
    (1024, 1.1202752996305587e-9)
    (2048, 5.552850342336678e-10)
};
\addplot[mnpColor7,solid,mark=otimes*,mark options={solid,fill=mnpColor7},forget plot] coordinates {
    (16, 1.971991452283293e-6)
    (32, 3.974158888853901e-7)
    (64, 1.364089914906376e-7)
    (128, 5.294299466617868e-8)
    (256, 2.438569354864139e-8)
    (512, 1.1486878228604888e-8)
    (1024, 5.6288756481482035e-9)
    (2048, 2.7760515618860393e-9)
};
\addlegendimage{mnpColor1,solid,mark=*,mark options={solid,fill=mnpColor1}}
\addlegendentry{Single}
\addlegendimage{mnpColor2,solid,mark=square*,mark options={solid,fill=mnpColor2}}
\addlegendentry{EB-2}
\addlegendimage{mnpColor3,solid,mark=triangle*,mark options={solid,fill=mnpColor3}}
\addlegendentry{PB-2}
\addlegendimage{mnpColor4,solid,mark=diamond*,mark options={solid,fill=mnpColor4}}
\addlegendentry{EB-3}
\addlegendimage{mnpColor5,solid,mark=pentagon*,mark options={solid,fill=mnpColor5}}
\addlegendentry{PB-3}
\addlegendimage{mnpColor6,solid,mark=triangle*,mark options={solid,fill=mnpColor6,rotate=180}}
\addlegendentry{LA-2}
\addlegendimage{mnpColor7,solid,mark=otimes*,mark options={solid,fill=mnpColor7}}
\addlegendentry{LA-3}

\nextgroupplot[title={(b) Ionosphere\\Classic (projected response)},
xmin=54.237288135593225,xmax=2416.64,ymin=1e-12,ymax=1e-5,
xtick={64,256,2048},xticklabels={64,256,2048},ytick={1e-12,1e-9,1e-6,1e-5},
]
\addplot[mnpColor1,solid,mark=*,mark options={solid,fill=mnpColor1},forget plot] coordinates {
    (64, 1.9364529824793667e-9)
    (128, 2.7985407108993795e-10)
    (256, 8.685035403099486e-11)
    (512, 3.508609339076094e-11)
    (1024, 1.5826452142715435e-11)
    (2048, 7.532762306789069e-12)
};
\addplot[mnpColor2,solid,mark=square*,mark options={solid,fill=mnpColor2},forget plot] coordinates {
    (64, 3.2302526198827415e-8)
    (128, 7.610983256216089e-10)
    (256, 1.8338548777418389e-10)
    (512, 7.097948006525258e-11)
    (1024, 3.177935231011522e-11)
    (2048, 1.509341455106754e-11)
};
\addplot[mnpColor3,solid,mark=triangle*,mark options={solid,fill=mnpColor3},forget plot] coordinates {
    (64, 6.13403392848461e-9)
    (128, 6.172158573712599e-10)
    (256, 1.7754391108670701e-10)
    (512, 7.048489048248861e-11)
    (1024, 3.169748053128498e-11)
    (2048, 1.5076159970915413e-11)
};
\addplot[mnpColor4,solid,mark=diamond*,mark options={solid,fill=mnpColor4},forget plot] coordinates {
    (64, 3.234861916958127e-7)
    (128, 3.638318767952673e-9)
    (256, 6.45814908790286e-10)
    (512, 2.28211521458085e-10)
    (1024, 9.839580354640777e-11)
    (2048, 4.5965613204585493e-11)
};
\addplot[mnpColor5,solid,mark=pentagon*,mark options={solid,fill=mnpColor5},forget plot] coordinates {
    (64, 2.667241201773946e-8)
    (128, 2.394752050309452e-9)
    (256, 6.068737422584502e-10)
    (512, 2.25634340674831e-10)
    (1024, 9.820240146903899e-11)
    (2048, 4.594488233726123e-11)
};
\addplot[mnpColor6,solid,mark=triangle*,mark options={solid,fill=mnpColor6,rotate=180},forget plot] coordinates {
    (64, 2.446421294231343e-7)
    (128, 3.0573930083058327e-9)
    (256, 6.278127689687802e-10)
    (512, 2.2671320046922342e-10)
    (1024, 9.843445496764135e-11)
    (2048, 4.599239777264951e-11)
};
\addplot[mnpColor7,solid,mark=otimes*,mark options={solid,fill=mnpColor7},forget plot] coordinates {
    (64, 2.708908845881525e-6)
    (128, 3.793109728289379e-8)
    (256, 3.980891438567744e-9)
    (512, 1.242166492746506e-9)
    (1024, 5.148634776756628e-10)
    (2048, 2.344663602112968e-10)
};

\nextgroupplot[title={(c) California Housing\\Classic (projected response)},
xmin=27.118644067796613,xmax=2416.64,ymin=1e-16,ymax=1e-8,
xtick={32,128,512,2048},xticklabels={32,128,512,2048},ytick={1e-16,1e-13,1e-10,1e-8},
]
\addplot[mnpColor1,solid,mark=*,mark options={solid,fill=mnpColor1},forget plot] coordinates {
    (32, 7.262657805528521e-14)
    (64, 1.884529987878621e-14)
    (128, 7.1423591032754e-15)
    (256, 3.1267372376093784e-15)
    (512, 1.4751381384686891e-15)
    (1024, 7.137176068494406e-16)
    (2048, 3.504094449801254e-16)
};
\addplot[mnpColor2,solid,mark=square*,mark options={solid,fill=mnpColor2},forget plot] coordinates {
    (32, 3.311951496361505e-13)
    (64, 4.2266077102446266e-14)
    (128, 1.4571788433579645e-14)
    (256, 6.283017058155817e-15)
    (512, 2.9360178314188373e-15)
    (1024, 1.4248402024615125e-15)
    (2048, 7.019014840720552e-16)
};
\addplot[mnpColor3,solid,mark=triangle*,mark options={solid,fill=mnpColor3},forget plot] coordinates {
    (32, 1.8463837816412466e-13)
    (64, 3.973014834901553e-14)
    (128, 1.44863223378688e-14)
    (256, 6.294858675474394e-15)
    (512, 2.943247677472799e-15)
    (1024, 1.4294737391777776e-15)
    (2048, 7.008695917379424e-16)
};
\addplot[mnpColor4,solid,mark=diamond*,mark options={solid,fill=mnpColor4},forget plot] coordinates {
    (32, 4.0037000550350176e-11)
    (64, 1.6125222286913712e-13)
    (128, 4.8025283369844046e-14)
    (256, 1.9745358589245018e-14)
    (512, 9.002077449789656e-15)
    (1024, 4.306571721722553e-15)
    (2048, 2.108030499671682e-15)
};
\addplot[mnpColor5,solid,mark=pentagon*,mark options={solid,fill=mnpColor5},forget plot] coordinates {
    (32, 8.041872765111698e-13)
    (64, 1.4226324488905297e-13)
    (128, 4.744063157925743e-14)
    (256, 1.9673716345119303e-14)
    (512, 9.016490949074005e-15)
    (1024, 4.316830043626269e-15)
    (2048, 2.1162634964985683e-15)
};
\addplot[mnpColor6,solid,mark=triangle*,mark options={solid,fill=mnpColor6,rotate=180},forget plot] coordinates {
    (32, 1.4240853188593823e-12)
    (64, 1.476891781754966e-13)
    (128, 4.73927235461878e-14)
    (256, 1.9598419395888006e-14)
    (512, 9.032399326988371e-15)
    (1024, 4.309909577054363e-15)
    (2048, 2.118196288622482e-15)
};
\addplot[mnpColor7,solid,mark=otimes*,mark options={solid,fill=mnpColor7},forget plot] coordinates {
    (32, 9.740278465886887e-10)
    (64, 1.017100058124693e-12)
    (128, 2.5734879234505127e-13)
    (256, 1.0337961470142872e-13)
    (512, 4.585734901779016e-14)
    (1024, 2.1781401509436127e-14)
    (2048, 1.0652716932988293e-14)
};

\nextgroupplot[title={(d) Abalone\\Classic (original response)},
xmin=13.559322033898306,xmax=2416.64,ymin=1e-8,ymax=1e-3,
xtick={16,64,256,2048},xticklabels={16,64,256,2048},ytick={1e-8,1e-6,1e-4,1e-3},
ylabel={Relative mean variance},
]
\addplot[mnpColor1,solid,mark=*,mark options={solid,fill=mnpColor1},forget plot] coordinates {
    (16, 1.4200947253844014e-5)
    (32, 6.483725676303114e-6)
    (64, 3.0700471314601466e-6)
    (128, 1.496606017633987e-6)
    (256, 7.359814489454162e-7)
    (512, 3.65702324972213e-7)
    (1024, 1.8263108456599258e-7)
    (2048, 9.097252080526731e-8)
};
\addplot[mnpColor2,solid,mark=square*,mark options={solid,fill=mnpColor2},forget plot] coordinates {
    (16, 3.1858456442491624e-5)
    (32, 1.3361990656062088e-5)
    (64, 6.1719496814636475e-6)
    (128, 2.9789833401478546e-6)
    (256, 1.4664668355212007e-6)
    (512, 7.310924489891678e-7)
    (1024, 3.645185237857735e-7)
    (2048, 1.8208312115605612e-7)
};
\addplot[mnpColor3,solid,mark=triangle*,mark options={solid,fill=mnpColor3},forget plot] coordinates {
    (16, 2.82807201895233e-5)
    (32, 1.2874778976688783e-5)
    (64, 6.115886841759385e-6)
    (128, 2.9867071489562085e-6)
    (256, 1.4683006590228362e-6)
    (512, 7.311406132435632e-7)
    (1024, 3.6452813324264995e-7)
    (2048, 1.8212542964756123e-7)
};
\addplot[mnpColor4,solid,mark=diamond*,mark options={solid,fill=mnpColor4},forget plot] coordinates {
    (16, 9.002437572593477e-5)
    (32, 3.849314484552153e-5)
    (64, 1.8014806317398037e-5)
    (128, 8.773966151974185e-6)
    (256, 4.364851508078345e-6)
    (512, 2.1779946385432543e-6)
    (1024, 1.089308945826936e-6)
    (2048, 5.45762121233123e-7)
};
\addplot[mnpColor5,solid,mark=pentagon*,mark options={solid,fill=mnpColor5},forget plot] coordinates {
    (16, 6.936450078175834e-5)
    (32, 3.584920822183703e-5)
    (64, 1.7825014295511697e-5)
    (128, 8.82691258585873e-6)
    (256, 4.386339273250123e-6)
    (512, 2.179862034105389e-6)
    (1024, 1.0902574308629965e-6)
    (2048, 5.454857764112795e-7)
};
\addplot[mnpColor6,solid,mark=triangle*,mark options={solid,fill=mnpColor6,rotate=180},forget plot] coordinates {
    (16, 8.506434736899357e-5)
    (32, 4.004323327267931e-5)
    (64, 1.8797435174814478e-5)
    (128, 9.049393384254405e-6)
    (256, 4.432691440163064e-6)
    (512, 2.2038279366387387e-6)
    (1024, 1.0911576471757752e-6)
    (2048, 5.46346699977452e-7)
};
\addplot[mnpColor7,solid,mark=otimes*,mark options={solid,fill=mnpColor7},forget plot] coordinates {
    (16, 0.0003355389338945391)
    (32, 0.0001694410913409635)
    (64, 9.12930920533183e-5)
    (128, 4.3577538561369424e-5)
    (256, 2.2137838542307028e-5)
    (512, 1.0872796092016472e-5)
    (1024, 5.4545067169617486e-6)
    (2048, 2.7187599304410444e-6)
};

\nextgroupplot[title={(e) Ionosphere\\Classic (original response)},
xmin=54.237288135593225,xmax=2416.64,ymin=1e-7,ymax=1e-3,
xtick={64,256,2048},xticklabels={64,256,2048},ytick={1e-7,1e-5,1e-3},
]
\addplot[mnpColor1,solid,mark=*,mark options={solid,fill=mnpColor1},forget plot] coordinates {
    (64, 9.168325968126944e-6)
    (128, 3.095006161449674e-6)
    (256, 1.327227130229279e-6)
    (512, 6.205592559438385e-7)
    (1024, 2.9953527662510614e-7)
    (2048, 1.473602472314786e-7)
};
\addplot[mnpColor2,solid,mark=square*,mark options={solid,fill=mnpColor2},forget plot] coordinates {
    (64, 2.9924655788051463e-5)
    (128, 6.841668962072845e-6)
    (256, 2.716655693716354e-6)
    (512, 1.2447702455003414e-6)
    (1024, 5.993646179982015e-7)
    (2048, 2.9456320090110936e-7)
};
\addplot[mnpColor3,solid,mark=triangle*,mark options={solid,fill=mnpColor3},forget plot] coordinates {
    (64, 1.8092512372591975e-5)
    (128, 6.191947407837221e-6)
    (256, 2.6609957839851434e-6)
    (512, 1.2393564635937408e-6)
    (1024, 5.992714019971475e-7)
    (2048, 2.951274859640065e-7)
};
\addplot[mnpColor4,solid,mark=diamond*,mark options={solid,fill=mnpColor4},forget plot] coordinates {
    (64, 0.00013906645565763437)
    (128, 2.270743648919562e-5)
    (256, 8.315624176863174e-6)
    (512, 3.748534581743037e-6)
    (1024, 1.8002028737832152e-6)
    (2048, 8.844636318749057e-7)
};
\addplot[mnpColor5,solid,mark=pentagon*,mark options={solid,fill=mnpColor5},forget plot] coordinates {
    (64, 5.066733230972648e-5)
    (128, 1.8427200652545186e-5)
    (256, 7.94576138153509e-6)
    (512, 3.713455389799448e-6)
    (1024, 1.7969843477872392e-6)
    (2048, 8.84576152773139e-7)
};
\addplot[mnpColor6,solid,mark=triangle*,mark options={solid,fill=mnpColor6,rotate=180},forget plot] coordinates {
    (64, 7.783515410424944e-5)
    (128, 2.1574250883328686e-5)
    (256, 8.399537849134849e-6)
    (512, 3.8058146023381003e-6)
    (1024, 1.8201018409278308e-6)
    (2048, 8.900830458884386e-7)
};
\addplot[mnpColor7,solid,mark=otimes*,mark options={solid,fill=mnpColor7},forget plot] coordinates {
    (64, 0.00044122370239173405)
    (128, 0.00012674379091347045)
    (256, 4.478636154284849e-5)
    (512, 1.9402856883857786e-5)
    (1024, 9.2109410027143e-6)
    (2048, 4.468664497799913e-6)
};

\nextgroupplot[title={(f) California Housing\\Classic (original response)},
xmin=27.118644067796613,xmax=2416.64,ymin=1e-8,ymax=1e-3,
xtick={32,128,512,2048},xticklabels={32,128,512,2048},ytick={1e-8,1e-6,1e-4,1e-3},
]
\addplot[mnpColor1,solid,mark=*,mark options={solid,fill=mnpColor1},forget plot] coordinates {
    (32, 1.769790588457836e-6)
    (64, 7.224045481749282e-7)
    (128, 3.298786898103518e-7)
    (256, 1.5792022715240148e-7)
    (512, 7.751198449812984e-8)
    (1024, 3.84322946527501e-8)
    (2048, 1.908060367495231e-8)
};
\addplot[mnpColor2,solid,mark=square*,mark options={solid,fill=mnpColor2},forget plot] coordinates {
    (32, 4.241937259631938e-6)
    (64, 1.4975289351599314e-6)
    (128, 6.647615643654767e-7)
    (256, 3.17562039915436e-7)
    (512, 1.55543536908018e-7)
    (1024, 7.694401411916874e-8)
    (2048, 3.818308454279697e-8)
};
\addplot[mnpColor3,solid,mark=triangle*,mark options={solid,fill=mnpColor3},forget plot] coordinates {
    (32, 3.605171330302331e-6)
    (64, 1.4578234471332631e-6)
    (128, 6.627893801410968e-7)
    (256, 3.180838817648424e-7)
    (512, 1.5551113222204992e-7)
    (1024, 7.704262204569987e-8)
    (2048, 3.831913455583316e-8)
};
\addplot[mnpColor4,solid,mark=diamond*,mark options={solid,fill=mnpColor4},forget plot] coordinates {
    (32, 1.6439134202645714e-5)
    (64, 4.639683159438476e-6)
    (128, 2.006057755369697e-6)
    (256, 9.541086864970786e-7)
    (512, 4.6605722406255706e-7)
    (1024, 2.3003191976580998e-7)
    (2048, 1.1435787662978984e-7)
};
\addplot[mnpColor5,solid,mark=pentagon*,mark options={solid,fill=mnpColor5},forget plot] coordinates {
    (32, 1.1136921304848189e-5)
    (64, 4.410236588054826e-6)
    (128, 1.999099469063058e-6)
    (256, 9.544470908972883e-7)
    (512, 4.654259220136125e-7)
    (1024, 2.310514692578986e-7)
    (2048, 1.149940863941076e-7)
};
\addplot[mnpColor6,solid,mark=triangle*,mark options={solid,fill=mnpColor6,rotate=180},forget plot] coordinates {
    (32, 1.3485953673188451e-5)
    (64, 4.63735153080544e-6)
    (128, 2.0329195571666903e-6)
    (256, 9.639188027073132e-7)
    (512, 4.6871417115322546e-7)
    (1024, 2.3098023751444653e-7)
    (2048, 1.1506804070318896e-7)
};
\addplot[mnpColor7,solid,mark=otimes*,mark options={solid,fill=mnpColor7},forget plot] coordinates {
    (32, 0.00010260701473910997)
    (64, 2.4503290348243518e-5)
    (128, 1.0110909770425497e-5)
    (256, 4.8509321965926324e-6)
    (512, 2.330131633272346e-6)
    (1024, 1.1566064379161153e-6)
    (2048, 5.731527516624186e-7)
};

\nextgroupplot[title={(g) Abalone\\Hessian (original response)},
xmin=13.559322033898306,xmax=2416.64,ymin=1e-6,ymax=1e0,
xtick={16,64,256,2048},xticklabels={16,64,256,2048},ytick={1e-6,1e-4,1e-2,1e0},
ylabel={Relative mean variance},
]
\addplot[mnpColor1,solid,mark=*,mark options={solid,fill=mnpColor1},forget plot] coordinates {
    (16, 0.0009280455337581207)
    (32, 0.00022524017702687157)
    (64, 8.097908455827045e-5)
    (128, 3.477482972315234e-5)
    (256, 1.6077114466274508e-5)
    (512, 7.766992054848268e-6)
    (1024, 3.8094547680498254e-6)
    (2048, 1.8806854261732292e-6)
};
\addplot[mnpColor2,solid,mark=square*,mark options={solid,fill=mnpColor2},forget plot] coordinates {
    (16, 0.003314169626093228)
    (32, 0.0005101450361895284)
    (64, 0.00016545016005218283)
    (128, 6.941683609849769e-5)
    (256, 3.2106880228330085e-5)
    (512, 1.546626470832756e-5)
    (1024, 7.619735792991329e-6)
    (2048, 3.7651593907533907e-6)
};
\addplot[mnpColor3,solid,mark=triangle*,mark options={solid,fill=mnpColor3},forget plot] coordinates {
    (16, 0.0019299128460456962)
    (32, 0.0004582737877913341)
    (64, 0.00016294607374620605)
    (128, 6.932840344708248e-5)
    (256, 3.2054745524074876e-5)
    (512, 1.5495978253889976e-5)
    (1024, 7.624126016067588e-6)
    (2048, 3.7623045809302176e-6)
};
\addplot[mnpColor4,solid,mark=diamond*,mark options={solid,fill=mnpColor4},forget plot] coordinates {
    (16, 0.018699589276297918)
    (32, 0.0019329189942864998)
    (64, 0.0005456314856954777)
    (128, 0.00021697083978487524)
    (256, 9.818633759246724e-5)
    (512, 4.696608552314789e-5)
    (1024, 2.2867348591015113e-5)
    (2048, 1.1294818895925828e-5)
};
\addplot[mnpColor5,solid,mark=pentagon*,mark options={solid,fill=mnpColor5},forget plot] coordinates {
    (16, 0.007133184684223174)
    (32, 0.001591143726132705)
    (64, 0.0005285310356922256)
    (128, 0.00021683279037292789)
    (256, 9.845272767416733e-5)
    (512, 4.709581900053958e-5)
    (1024, 2.2919430044214986e-5)
    (2048, 1.13419848315386e-5)
};
\addplot[mnpColor6,solid,mark=triangle*,mark options={solid,fill=mnpColor6,rotate=180},forget plot] coordinates {
    (16, 0.012586465471962306)
    (32, 0.0018420526339888686)
    (64, 0.0005507546631886987)
    (128, 0.0002203288398963717)
    (256, 9.86162170466958e-5)
    (512, 4.702762918199712e-5)
    (1024, 2.2992173154581185e-5)
    (2048, 1.1320964728757978e-5)
};
\addplot[mnpColor7,solid,mark=otimes*,mark options={solid,fill=mnpColor7},forget plot] coordinates {
    (16, 0.11000571658656712)
    (32, 0.010768929148309505)
    (64, 0.003062724638267509)
    (128, 0.0011185258187087768)
    (256, 0.0005066544689107396)
    (512, 0.0002366365380619673)
    (1024, 0.00011541982042028508)
    (2048, 5.687887077116726e-5)
};

\nextgroupplot[title={(h) Ionosphere\\Hessian (original response)},
xmin=54.237288135593225,xmax=2416.64,ymin=1e-7,ymax=1e1,
xtick={64,256,2048},xticklabels={64,256,2048},ytick={1e-7,1e-4,1e-1,1e1},
]
\addplot[mnpColor1,solid,mark=*,mark options={solid,fill=mnpColor1},forget plot] coordinates {
    (64, 0.00026041538517231587)
    (128, 3.634963321043949e-5)
    (256, 1.1178650268470532e-5)
    (512, 4.497281862490921e-6)
    (1024, 2.026380086499702e-6)
    (2048, 9.6393466663647e-7)
};
\addplot[mnpColor2,solid,mark=square*,mark options={solid,fill=mnpColor2},forget plot] coordinates {
    (64, 0.005369549546557383)
    (128, 9.933533027940488e-5)
    (256, 2.3611175902361817e-5)
    (512, 9.09538109896928e-6)
    (1024, 4.066094846344389e-6)
    (2048, 1.92730953595916e-6)
};
\addplot[mnpColor3,solid,mark=triangle*,mark options={solid,fill=mnpColor3},forget plot] coordinates {
    (64, 0.0008977677875267296)
    (128, 8.09990250217143e-5)
    (256, 2.282561626549826e-5)
    (512, 9.034074987916682e-6)
    (1024, 4.066765223553363e-6)
    (2048, 1.926971268049135e-6)
};
\addplot[mnpColor4,solid,mark=diamond*,mark options={solid,fill=mnpColor4},forget plot] coordinates {
    (64, 0.07895216992232919)
    (128, 0.00048214415053266137)
    (256, 8.3351672268897e-5)
    (512, 2.9314573389848904e-5)
    (1024, 1.2629097301862677e-5)
    (2048, 5.878549418215397e-6)
};
\addplot[mnpColor5,solid,mark=pentagon*,mark options={solid,fill=mnpColor5},forget plot] coordinates {
    (64, 0.003994177982988855)
    (128, 0.0003173256381550378)
    (256, 7.838209341197679e-5)
    (512, 2.898597216256877e-5)
    (1024, 1.2574959987284936e-5)
    (2048, 5.8829197534918034e-6)
};
\addplot[mnpColor6,solid,mark=triangle*,mark options={solid,fill=mnpColor6,rotate=180},forget plot] coordinates {
    (64, 0.053074558650594475)
    (128, 0.0004062705761354634)
    (256, 8.093617313728531e-5)
    (512, 2.9181675765999366e-5)
    (1024, 1.2617782880012353e-5)
    (2048, 5.88018427442396e-6)
};
\addplot[mnpColor7,solid,mark=otimes*,mark options={solid,fill=mnpColor7},forget plot] coordinates {
    (64, 0.8868480458625466)
    (128, 0.005414758822082777)
    (256, 0.0005161878154749121)
    (512, 0.00015996447131781373)
    (1024, 6.59664095428451e-5)
    (2048, 3.0007609561237544e-5)
};

\nextgroupplot[title={(i) California Housing\\Hessian (original response)},
xmin=27.118644067796613,xmax=2416.64,ymin=1e-7,ymax=1e0,
xtick={32,128,512,2048},xticklabels={32,128,512,2048},ytick={1e-7,1e-4,1e-1,1e0},
]
\addplot[mnpColor1,solid,mark=*,mark options={solid,fill=mnpColor1},forget plot] coordinates {
    (32, 4.224502228449988e-5)
    (64, 1.1102055610892583e-5)
    (128, 4.215391097926007e-6)
    (256, 1.8534003994505014e-6)
    (512, 8.696163415003966e-7)
    (1024, 4.206466181646588e-7)
    (2048, 2.070980645104946e-7)
};
\addplot[mnpColor2,solid,mark=square*,mark options={solid,fill=mnpColor2},forget plot] coordinates {
    (32, 0.00017997967416786455)
    (64, 2.459738264027286e-5)
    (128, 8.594621698847194e-6)
    (256, 3.7000616663802716e-6)
    (512, 1.7396274727218173e-6)
    (1024, 8.424386667153429e-7)
    (2048, 4.137206307096796e-7)
};
\addplot[mnpColor3,solid,mark=triangle*,mark options={solid,fill=mnpColor3},forget plot] coordinates {
    (32, 0.00010747677416814208)
    (64, 2.3276188115116812e-5)
    (128, 8.51888449270521e-6)
    (256, 3.707005884475647e-6)
    (512, 1.733048985463503e-6)
    (1024, 8.403915612690511e-7)
    (2048, 4.1424726969035116e-7)
};
\addplot[mnpColor4,solid,mark=diamond*,mark options={solid,fill=mnpColor4},forget plot] coordinates {
    (32, 0.05016931979990436)
    (64, 9.305378236079631e-5)
    (128, 2.8091847749147058e-5)
    (256, 1.154870241971352e-5)
    (512, 5.335386478952257e-6)
    (1024, 2.5466789567158475e-6)
    (2048, 1.2480296588727264e-6)
};
\addplot[mnpColor5,solid,mark=pentagon*,mark options={solid,fill=mnpColor5},forget plot] coordinates {
    (32, 0.00045860390077054043)
    (64, 8.315676668077787e-5)
    (128, 2.7851245672735164e-5)
    (256, 1.1592876139726887e-5)
    (512, 5.332935269026664e-6)
    (1024, 2.5463100484239443e-6)
    (2048, 1.2504925469655451e-6)
};
\addplot[mnpColor6,solid,mark=triangle*,mark options={solid,fill=mnpColor6,rotate=180},forget plot] coordinates {
    (32, 0.0008154189004452731)
    (64, 8.651188149799463e-5)
    (128, 2.8074139372436626e-5)
    (256, 1.1587619600022369e-5)
    (512, 5.311315343432478e-6)
    (1024, 2.5542862132690296e-6)
    (2048, 1.2520790048918195e-6)
};
\addplot[mnpColor7,solid,mark=otimes*,mark options={solid,fill=mnpColor7},forget plot] coordinates {
    (32, 0.6930185520367806)
    (64, 0.0005923944941586563)
    (128, 0.00015198320212850936)
    (256, 6.105373245650653e-5)
    (512, 2.7006001124449873e-5)
    (1024, 1.2928363922592832e-5)
    (2048, 6.290973302040056e-6)
};
\end{groupplot}
\end{tikzpicture}
\par\vspace{1mm}
\pgfplotslegendfromname{meanVarianceNinePanelLegend}
\endgroup
\caption{Normalized variance of the Monte Carlo average versus sketch size, with $N=200{,}000$ draws.}
\label{fig:mean-variance-nine-panels}
\end{figure}

%% file: checklist.tex
\section*{NeurIPS Paper Checklist}

\begin{enumerate}

\item {\bf Claims}
    \item[] Question: Do the main claims made in the abstract and introduction accurately reflect the paper's contributions and scope?
    \item[] Answer: \answerYes{} 
    \item[] Justification: The abstract and introduction state the study of bias in sketched ridge regression problem. The paper provides theoretical bias reduction guarantee via iterative methods, with empirical validation.
    \item[] Guidelines:
    \begin{itemize}
        \item The answer \answerNA{} means that the abstract and introduction do not include the claims made in the paper.
        \item The abstract and/or introduction should clearly state the claims made, including the contributions made in the paper and important assumptions and limitations. A \answerNo{} or \answerNA{} answer to this question will not be perceived well by the reviewers. 
        \item The claims made should match theoretical and experimental results, and reflect how much the results can be expected to generalize to other settings. 
        \item It is fine to include aspirational goals as motivation as long as it is clear that these goals are not attained by the paper. 
    \end{itemize}

\item {\bf Limitations}
    \item[] Question: Does the paper discuss the limitations of the work performed by the authors?
    \item[] Answer: \answerYes{} 
    \item[] Justification: The limitation and future work is discussed in the Conclusion section
    \item[] Guidelines:
    \begin{itemize}
        \item The answer \answerNA{} means that the paper has no limitation while the answer \answerNo{} means that the paper has limitations, but those are not discussed in the paper. 
        \item The authors are encouraged to create a separate ``Limitations'' section in their paper.
        \item The paper should point out any strong assumptions and how robust the results are to violations of these assumptions (e.g., independence assumptions, noiseless settings, model well-specification, asymptotic approximations only holding locally). The authors should reflect on how these assumptions might be violated in practice and what the implications would be.
        \item The authors should reflect on the scope of the claims made, e.g., if the approach was only tested on a few datasets or with a few runs. In general, empirical results often depend on implicit assumptions, which should be articulated.
        \item The authors should reflect on the factors that influence the performance of the approach. For example, a facial recognition algorithm may perform poorly when image resolution is low or images are taken in low lighting. Or a speech-to-text system might not be used reliably to provide closed captions for online lectures because it fails to handle technical jargon.
        \item The authors should discuss the computational efficiency of the proposed algorithms and how they scale with dataset size.
        \item If applicable, the authors should discuss possible limitations of their approach to address problems of privacy and fairness.
        \item While the authors might fear that complete honesty about limitations might be used by reviewers as grounds for rejection, a worse outcome might be that reviewers discover limitations that aren't acknowledged in the paper. The authors should use their best judgment and recognize that individual actions in favor of transparency play an important role in developing norms that preserve the integrity of the community. Reviewers will be specifically instructed to not penalize honesty concerning limitations.
    \end{itemize}

\item {\bf Theory assumptions and proofs}
    \item[] Question: For each theoretical result, does the paper provide the full set of assumptions and a complete (and correct) proof?
    \item[] Answer: \answerYes{} 
    \item[] Justification: All theorems include clear assumptions and full proof are provided in Appendix~\ref{appx:bias},Appendix~\ref{appx:bootstrap}, Appendix~\ref{appx:linear_aggragation}
    \item[] Guidelines:
    \begin{itemize}
        \item The answer \answerNA{} means that the paper does not include theoretical results. 
        \item All the theorems, formulas, and proofs in the paper should be numbered and cross-referenced.
        \item All assumptions should be clearly stated or referenced in the statement of any theorems.
        \item The proofs can either appear in the main paper or the supplemental material, but if they appear in the supplemental material, the authors are encouraged to provide a short proof sketch to provide intuition. 
        \item Inversely, any informal proof provided in the core of the paper should be complemented by formal proofs provided in appendix or supplemental material.
        \item Theorems and Lemmas that the proof relies upon should be properly referenced. 
    \end{itemize}

    \item {\bf Experimental result reproducibility}
    \item[] Question: Does the paper fully disclose all the information needed to reproduce the main experimental results of the paper to the extent that it affects the main claims and/or conclusions of the paper (regardless of whether the code and data are provided or not)?
    \item[] Answer: \answerYes{} 
    \item[] Justification: Appendix~\ref{appx:experiment} provides implement details including parameter setting, evaluation metrics.
    \item[] Guidelines:
    \begin{itemize}
        \item The answer \answerNA{} means that the paper does not include experiments.
        \item If the paper includes experiments, a \answerNo{} answer to this question will not be perceived well by the reviewers: Making the paper reproducible is important, regardless of whether the code and data are provided or not.
        \item If the contribution is a dataset and\slash or model, the authors should describe the steps taken to make their results reproducible or verifiable. 
        \item Depending on the contribution, reproducibility can be accomplished in various ways. For example, if the contribution is a novel architecture, describing the architecture fully might suffice, or if the contribution is a specific model and empirical evaluation, it may be necessary to either make it possible for others to replicate the model with the same dataset, or provide access to the model. In general. releasing code and data is often one good way to accomplish this, but reproducibility can also be provided via detailed instructions for how to replicate the results, access to a hosted model (e.g., in the case of a large language model), releasing of a model checkpoint, or other means that are appropriate to the research performed.
        \item While NeurIPS does not require releasing code, the conference does require all submissions to provide some reasonable avenue for reproducibility, which may depend on the nature of the contribution. For example
        \begin{enumerate}
            \item If the contribution is primarily a new algorithm, the paper should make it clear how to reproduce that algorithm.
            \item If the contribution is primarily a new model architecture, the paper should describe the architecture clearly and fully.
            \item If the contribution is a new model (e.g., a large language model), then there should either be a way to access this model for reproducing the results or a way to reproduce the model (e.g., with an open-source dataset or instructions for how to construct the dataset).
            \item We recognize that reproducibility may be tricky in some cases, in which case authors are welcome to describe the particular way they provide for reproducibility. In the case of closed-source models, it may be that access to the model is limited in some way (e.g., to registered users), but it should be possible for other researchers to have some path to reproducing or verifying the results.
        \end{enumerate}
    \end{itemize}

\item {\bf Open access to data and code}
    \item[] Question: Does the paper provide open access to the data and code, with sufficient instructions to faithfully reproduce the main experimental results, as described in supplemental material?
    \item[] Answer: \answerYes{}{} 
    \item[] Justification: The data is public via UC Irvine Machine Learning Repository~(\url{https://archive.ics.uci.edu}). The code is provided as in the supplemental material.
    \item[] Guidelines:
    \begin{itemize}
        \item The answer \answerNA{} means that paper does not include experiments requiring code.
        \item Please see the NeurIPS code and data submission guidelines (\url{https://neurips.cc/public/guides/CodeSubmissionPolicy}) for more details.
        \item While we encourage the release of code and data, we understand that this might not be possible, so \answerNo{} is an acceptable answer. Papers cannot be rejected simply for not including code, unless this is central to the contribution (e.g., for a new open-source benchmark).
        \item The instructions should contain the exact command and environment needed to run to reproduce the results. See the NeurIPS code and data submission guidelines (\url{https://neurips.cc/public/guides/CodeSubmissionPolicy}) for more details.
        \item The authors should provide instructions on data access and preparation, including how to access the raw data, preprocessed data, intermediate data, and generated data, etc.
        \item The authors should provide scripts to reproduce all experimental results for the new proposed method and baselines. If only a subset of experiments are reproducible, they should state which ones are omitted from the script and why.
        \item At submission time, to preserve anonymity, the authors should release anonymized versions (if applicable).
        \item Providing as much information as possible in supplemental material (appended to the paper) is recommended, but including URLs to data and code is permitted.
    \end{itemize}

\item {\bf Experimental setting/details}
    \item[] Question: Does the paper specify all the training and test details (e.g., data splits, hyperparameters, how they were chosen, type of optimizer) necessary to understand the results?
    \item[] Answer: \answerYes{} 
    \item[] Justification: Appendix~\ref{appx:experiment} describes the detail of all experiment setting.
    \item[] Guidelines:
    \begin{itemize}
        \item The answer \answerNA{} means that the paper does not include experiments.
        \item The experimental setting should be presented in the core of the paper to a level of detail that is necessary to appreciate the results and make sense of them.
        \item The full details can be provided either with the code, in appendix, or as supplemental material.
    \end{itemize}

\item {\bf Experiment statistical significance}
    \item[] Question: Does the paper report error bars suitably and correctly defined or other appropriate information about the statistical significance of the experiments?
    \item[] Answer: \answerYes{} 
    \item[] Justification: We report the errors across repeated trials and present the results with clear plots and captions.
    \item[] Guidelines:
    \begin{itemize}
        \item The answer \answerNA{} means that the paper does not include experiments.
        \item The authors should answer \answerYes{} if the results are accompanied by error bars, confidence intervals, or statistical significance tests, at least for the experiments that support the main claims of the paper.
        \item The factors of variability that the error bars are capturing should be clearly stated (for example, train/test split, initialization, random drawing of some parameter, or overall run with given experimental conditions).
        \item The method for calculating the error bars should be explained (closed form formula, call to a library function, bootstrap, etc.)
        \item The assumptions made should be given (e.g., Normally distributed errors).
        \item It should be clear whether the error bar is the standard deviation or the standard error of the mean.
        \item It is OK to report 1-sigma error bars, but one should state it. The authors should preferably report a 2-sigma error bar than state that they have a 96\% CI, if the hypothesis of Normality of errors is not verified.
        \item For asymmetric distributions, the authors should be careful not to show in tables or figures symmetric error bars that would yield results that are out of range (e.g., negative error rates).
        \item If error bars are reported in tables or plots, the authors should explain in the text how they were calculated and reference the corresponding figures or tables in the text.
    \end{itemize}

\item {\bf Experiments compute resources}
    \item[] Question: For each experiment, does the paper provide sufficient information on the computer resources (type of compute workers, memory, time of execution) needed to reproduce the experiments?
    \item[] Answer: \answerYes{} 
    \item[] Justification: Our experiments are not computationally heavy and are run on a commercial laptop. There is no specific resources required to reproduce the experiments.
    \item[] Guidelines:
    \begin{itemize}
        \item The answer \answerNA{} means that the paper does not include experiments.
        \item The paper should indicate the type of compute workers CPU or GPU, internal cluster, or cloud provider, including relevant memory and storage.
        \item The paper should provide the amount of compute required for each of the individual experimental runs as well as estimate the total compute. 
        \item The paper should disclose whether the full research project required more compute than the experiments reported in the paper (e.g., preliminary or failed experiments that didn't make it into the paper). 
    \end{itemize}
    
\item {\bf Code of ethics}
    \item[] Question: Does the research conducted in the paper conform, in every respect, with the NeurIPS Code of Ethics \url{https://neurips.cc/public/EthicsGuidelines}?
    \item[] Answer: \answerYes{} 
    \item[] Justification: The research is mainly theoretical and uses only public datasets, and conforms to ethical standards.
    \item[] Guidelines:
    \begin{itemize}
        \item The answer \answerNA{} means that the authors have not reviewed the NeurIPS Code of Ethics.
        \item If the authors answer \answerNo, they should explain the special circumstances that require a deviation from the Code of Ethics.
        \item The authors should make sure to preserve anonymity (e.g., if there is a special consideration due to laws or regulations in their jurisdiction).
    \end{itemize}

\item {\bf Broader impacts}
    \item[] Question: Does the paper discuss both potential positive societal impacts and negative societal impacts of the work performed?
    \item[] Answer: \answerNA{} 
    \item[] Justification: This is a theoretical work, with no direct societal impact.
    \item[] Guidelines:
    \begin{itemize}
        \item The answer \answerNA{} means that there is no societal impact of the work performed.
        \item If the authors answer \answerNA{} or \answerNo, they should explain why their work has no societal impact or why the paper does not address societal impact.
        \item Examples of negative societal impacts include potential malicious or unintended uses (e.g., disinformation, generating fake profiles, surveillance), fairness considerations (e.g., deployment of technologies that could make decisions that unfairly impact specific groups), privacy considerations, and security considerations.
        \item The conference expects that many papers will be foundational research and not tied to particular applications, let alone deployments. However, if there is a direct path to any negative applications, the authors should point it out. For example, it is legitimate to point out that an improvement in the quality of generative models could be used to generate Deepfakes for disinformation. On the other hand, it is not needed to point out that a generic algorithm for optimizing neural networks could enable people to train models that generate Deepfakes faster.
        \item The authors should consider possible harms that could arise when the technology is being used as intended and functioning correctly, harms that could arise when the technology is being used as intended but gives incorrect results, and harms following from (intentional or unintentional) misuse of the technology.
        \item If there are negative societal impacts, the authors could also discuss possible mitigation strategies (e.g., gated release of models, providing defenses in addition to attacks, mechanisms for monitoring misuse, mechanisms to monitor how a system learns from feedback over time, improving the efficiency and accessibility of ML).
    \end{itemize}
    
\item {\bf Safeguards}
    \item[] Question: Does the paper describe safeguards that have been put in place for responsible release of data or models that have a high risk for misuse (e.g., pre-trained language models, image generators, or scraped datasets)?
    \item[] Answer: \answerNA{} 
    \item[] Justification: This paper poses no risk for misuse.
    \item[] Guidelines:
    \begin{itemize}
        \item The answer \answerNA{} means that the paper poses no such risks.
        \item Released models that have a high risk for misuse or dual-use should be released with necessary safeguards to allow for controlled use of the model, for example by requiring that users adhere to usage guidelines or restrictions to access the model or implementing safety filters. 
        \item Datasets that have been scraped from the Internet could pose safety risks. The authors should describe how they avoided releasing unsafe images.
        \item We recognize that providing effective safeguards is challenging, and many papers do not require this, but we encourage authors to take this into account and make a best faith effort.
    \end{itemize}

\item {\bf Licenses for existing assets}
    \item[] Question: Are the creators or original owners of assets (e.g., code, data, models), used in the paper, properly credited and are the license and terms of use explicitly mentioned and properly respected?
    \item[] Answer: \answerYes{} 
    \item[] Justification: This paper cites datasets properly. 
    \item[] Guidelines:
    \begin{itemize}
        \item The answer \answerNA{} means that the paper does not use existing assets.
        \item The authors should cite the original paper that produced the code package or dataset.
        \item The authors should state which version of the asset is used and, if possible, include a URL.
        \item The name of the license (e.g., CC-BY 4.0) should be included for each asset.
        \item For scraped data from a particular source (e.g., website), the copyright and terms of service of that source should be provided.
        \item If assets are released, the license, copyright information, and terms of use in the package should be provided. For popular datasets, \url{paperswithcode.com/datasets} has curated licenses for some datasets. Their licensing guide can help determine the license of a dataset.
        \item For existing datasets that are re-packaged, both the original license and the license of the derived asset (if it has changed) should be provided.
        \item If this information is not available online, the authors are encouraged to reach out to the asset's creators.
    \end{itemize}

\item {\bf New assets}
    \item[] Question: Are new assets introduced in the paper well documented and is the documentation provided alongside the assets?
    \item[] Answer: \answerNA{} 
    \item[] Justification: This research does not release new assets.
    \item[] Guidelines:
    \begin{itemize}
        \item The answer \answerNA{} means that the paper does not release new assets.
        \item Researchers should communicate the details of the dataset\slash code\slash model as part of their submissions via structured templates. This includes details about training, license, limitations, etc. 
        \item The paper should discuss whether and how consent was obtained from people whose asset is used.
        \item At submission time, remember to anonymize your assets (if applicable). You can either create an anonymized URL or include an anonymized zip file.
    \end{itemize}

\item {\bf Crowdsourcing and research with human subjects}
    \item[] Question: For crowdsourcing experiments and research with human subjects, does the paper include the full text of instructions given to participants and screenshots, if applicable, as well as details about compensation (if any)? 
    \item[] Answer: \answerNA{} 
    \item[] Justification: This research does not involve crowdsourcing nor research with human subjects.
    \item[] Guidelines:
    \begin{itemize}
        \item The answer \answerNA{} means that the paper does not involve crowdsourcing nor research with human subjects.
        \item Including this information in the supplemental material is fine, but if the main contribution of the paper involves human subjects, then as much detail as possible should be included in the main paper. 
        \item According to the NeurIPS Code of Ethics, workers involved in data collection, curation, or other labor should be paid at least the minimum wage in the country of the data collector. 
    \end{itemize}

\item {\bf Institutional review board (IRB) approvals or equivalent for research with human subjects}
    \item[] Question: Does the paper describe potential risks incurred by study participants, whether such risks were disclosed to the subjects, and whether Institutional Review Board (IRB) approvals (or an equivalent approval/review based on the requirements of your country or institution) were obtained?
    \item[] Answer: \answerNA{} 
    \item[] Justification: This research does not involve crowdsourcing nor research with human subjects.
    \item[] Guidelines:
    \begin{itemize}
        \item The answer \answerNA{} means that the paper does not involve crowdsourcing nor research with human subjects.
        \item Depending on the country in which research is conducted, IRB approval (or equivalent) may be required for any human subjects research. If you obtained IRB approval, you should clearly state this in the paper. 
        \item We recognize that the procedures for this may vary significantly between institutions and locations, and we expect authors to adhere to the NeurIPS Code of Ethics and the guidelines for their institution. 
        \item For initial submissions, do not include any information that would break anonymity (if applicable), such as the institution conducting the review.
    \end{itemize}

\item {\bf Declaration of LLM usage}
    \item[] Question: Does the paper describe the usage of LLMs if it is an important, original, or non-standard component of the core methods in this research? Note that if the LLM is used only for writing, editing, or formatting purposes and does \emph{not} impact the core methodology, scientific rigor, or originality of the research, declaration is not required.
    \item[] Answer: \answerNA{} 
    \item[] Justification: The he core method development in this research does not involve LLMs.
    \item[] Guidelines:
    \begin{itemize}
        \item The answer \answerNA{} means that the core method development in this research does not involve LLMs as any important, original, or non-standard components.
        \item Please refer to our LLM policy in the NeurIPS handbook for what should or should not be described.
    \end{itemize}

\end{enumerate}

%% file: ref.bib
@article{Wang2018,
  author  = {Shusen Wang and Alex Gittens and Michael W. Mahoney},
  title   = {Sketched Ridge Regression: Optimization Perspective, Statistical Perspective, and Model Averaging},
  journal = {Journal of Machine Learning Research},
  year    = {2018},
  volume  = {18},
  number  = {218},
  pages   = {1--50}
}

@incollection {Johnson_Lindenstrauss,
    AUTHOR = {Johnson, William B. and Lindenstrauss, Joram},
     TITLE = {Extensions of {L}ipschitz mappings into a {H}ilbert space},
 BOOKTITLE = {Conference in modern analysis and probability ({N}ew {H}aven,
              {C}onn., 1982)},
    SERIES = {Contemp. Math.},
    VOLUME = {26},
     PAGES = {189--206},
 PUBLISHER = {Amer. Math. Soc., Providence, RI},
      YEAR = {1984}
}

@article {Jiao2020,
    AUTHOR = {Jiao, Jiantao and Han, Yanjun},
     TITLE = {Bias correction with jackknife, bootstrap, and {T}aylor
              series},
   JOURNAL = {IEEE Trans. Inform. Theory},
  FJOURNAL = {Institute of Electrical and Electronics Engineers.
              Transactions on Information Theory},
    VOLUME = {66},
      YEAR = {2020},
    NUMBER = {7},
     PAGES = {4392--4418}
}

@article{dobriban2019asymptotics,
  title={Asymptotics for sketching in least squares regression},
  author={Dobriban, Edgar and Liu, Sifan},
  journal={Advances in Neural Information Processing Systems},
  volume={32},
  year={2019}
}

@inproceedings{
    Liu2020Ridge,
    title={Ridge Regression: Structure, Cross-Validation, and Sketching},
    author={Sifan Liu and Edgar Dobriban},
    booktitle={International Conference on Learning Representations},
    year={2020}
}

@article{Lopes2019,
  author  = {Miles E. Lopes and Shusen Wang and Michael W. Mahoney},
  title   = {A Bootstrap Method for Error Estimation in Randomized Matrix Multiplication},
  journal = {Journal of Machine Learning Research},
  year    = {2019},
  volume  = {20},
  number  = {39},
  pages   = {1--40}
}

@InProceedings{Lopes2020,
  title = 	 {Error Estimation for Sketched {SVD} via the Bootstrap},
  author =       {Lopes, Miles and Erichson, N. Benjamin and Mahoney, Michael},
  booktitle = 	 {Proceedings of the 37th International Conference on Machine Learning},
  pages = 	 {6382--6392},
  year = 	 {2020},
  editor = 	 {III, Hal Daumé and Singh, Aarti},
  volume = 	 {119},
  series = 	 {Proceedings of Machine Learning Research},
  month = 	 {13--18 Jul},
  publisher =    {PMLR}
}

@article {MR995110,
    AUTHOR = {Hall, Peter and Martin, Michael A.},
     TITLE = {On bootstrap resampling and iteration},
   JOURNAL = {Biometrika},
  FJOURNAL = {Biometrika},
    VOLUME = {75},
      YEAR = {1988},
    NUMBER = {4},
     PAGES = {661--671}
}

@article {MR81040,
    AUTHOR = {Quenouille, M. H.},
     TITLE = {Notes on bias in estimation},
   JOURNAL = {Biometrika},
  FJOURNAL = {Biometrika},
    VOLUME = {43},
      YEAR = {1956},
     PAGES = {353--360}
}

@article {MR32176,
    AUTHOR = {Quenouille, M. H.},
     TITLE = {Approximate tests of correlation in time-series},
   JOURNAL = {J. Roy. Statist. Soc. Ser. B},
  FJOURNAL = {Journal of the Royal Statistical Society. Series B.
              Methodological},
    VOLUME = {11},
      YEAR = {1949},
     PAGES = {68--84}
}

@article{tukey1958bias,
  title={Bias and confidence in not quite large samples},
  author={Tukey, John},
  journal={Ann. Math. Statist.},
  volume={29},
  pages={614},
  year={1958}
}

@InProceedings{koltchinskii2026trace,
    author="Koltchinskii, Vladimir",
    title="Estimation of Trace Functionals and Spectral Measures of Covariance Operators in Gaussian Models",
    booktitle="High Dimensional Probability X",
    year="2026",
    publisher="Springer Nature Switzerland",
    address="Cham",
    pages="473--550",
    isbn="978-3-032-06057-0"
}

@article {Koltchinskii_Lounici,
    AUTHOR = {Koltchinskii, Vladimir and Lounici, Karim},
     TITLE = {Concentration inequalities and moment bounds for sample
              covariance operators},
   JOURNAL = {Bernoulli},
  FJOURNAL = {Bernoulli. Official Journal of the Bernoulli Society for
              Mathematical Statistics and Probability},
    VOLUME = {23},
      YEAR = {2017},
    NUMBER = {1},
     PAGES = {110--133}
}

@article {Koltchinskii_Lounici_bilinear,
    AUTHOR = {Koltchinskii, Vladimir and Lounici, Karim},
     TITLE = {Asymptotics and concentration bounds for bilinear forms of
              spectral projectors of sample covariance},
   JOURNAL = {Ann. Inst. Henri Poincar\'e{} Probab. Stat.},
  FJOURNAL = {Annales de l'Institut Henri Poincar\'e{} Probabilit\'es et
              Statistiques},
    VOLUME = {52},
      YEAR = {2016},
    NUMBER = {4},
     PAGES = {1976--2013}
}

@article {Koltchinskii_Nickl,
    AUTHOR = {Koltchinskii, Vladimir and L\"offler, Matthias and Nickl,
              Richard},
     TITLE = {Efficient estimation of linear functionals of principal
              components},
   JOURNAL = {Ann. Statist.},
  FJOURNAL = {The Annals of Statistics},
    VOLUME = {48},
      YEAR = {2020},
    NUMBER = {1},
     PAGES = {464--490}
}

@inproceedings {Koltchinskii2018,
    AUTHOR = {Koltchinskii, Vladimir},
     TITLE = {Asymptotic efficiency in high-dimensional covariance
              estimation},
 BOOKTITLE = {Proceedings of the {I}nternational {C}ongress of
              {M}athematicians---{R}io de {J}aneiro 2018. {V}ol. {IV}.
              {I}nvited lectures},
     PAGES = {2903--2923},
 PUBLISHER = {World Sci. Publ., Hackensack, NJ},
      YEAR = {2018}
}

@article {Koltchinskii2021,
    AUTHOR = {Koltchinskii, Vladimir},
     TITLE = {Asymptotically efficient estimation of smooth functionals of
              covariance operators},
   JOURNAL = {J. Eur. Math. Soc. (JEMS)},
  FJOURNAL = {Journal of the European Mathematical Society (JEMS)},
    VOLUME = {23},
      YEAR = {2021},
    NUMBER = {3},
     PAGES = {765--843}
}

@article {Koltchinskii_Zhilova,
    AUTHOR = {Koltchinskii, Vladimir and Zhilova, Mayya},
     TITLE = {Estimation of smooth functionals in normal models: bias
              reduction and asymptotic efficiency},
   JOURNAL = {Ann. Statist.},
  FJOURNAL = {The Annals of Statistics},
    VOLUME = {49},
      YEAR = {2021},
    NUMBER = {5},
     PAGES = {2577--2610}
}

@article {Koltchinskii2025,
    AUTHOR = {Koltchinskii, Vladimir},
     TITLE = {Estimation of smooth functionals of covariance operators:
              jackknife bias reduction and bounds in terms of effective
              rank},
   JOURNAL = {Ann. Inst. Henri Poincar\'e{} Probab. Stat.},
  FJOURNAL = {Annales de l'Institut Henri Poincar\'e{} Probabilit\'es et
              Statistiques},
    VOLUME = {61},
      YEAR = {2025},
    NUMBER = {1},
     PAGES = {665--712}
}

@article{drineas2016randnla,
  title={RandNLA: randomized numerical linear algebra},
  author={Drineas, Petros and Mahoney, Michael W},
  journal={Communications of the ACM},
  volume={59},
  number={6},
  pages={80--90},
  year={2016},
  publisher={ACM New York, NY, USA}
}

@article {woodruff2014,
    AUTHOR = {Woodruff, David P.},
     TITLE = {Sketching as a tool for numerical linear algebra},
   JOURNAL = {Found. Trends Theor. Comput. Sci.},
  FJOURNAL = {Foundations and Trends${}^\circledR$ in Theoretical Computer
              Science},
    VOLUME = {10},
      YEAR = {2014},
    NUMBER = {1-2},
     PAGES = {iv+157}
}

@article {martinsson2020,
    AUTHOR = {Martinsson, Per-Gunnar and Tropp, Joel A.},
     TITLE = {Randomized numerical linear algebra: foundations and
              algorithms},
   JOURNAL = {Acta Numer.},
  FJOURNAL = {Acta Numerica},
    VOLUME = {29},
      YEAR = {2020},
     PAGES = {403--572}
}

@article {kannan2017,
    AUTHOR = {Kannan, Ravindran and Vempala, Santosh},
     TITLE = {Randomized algorithms in numerical linear algebra},
   JOURNAL = {Acta Numer.},
  FJOURNAL = {Acta Numerica},
    VOLUME = {26},
      YEAR = {2017},
     PAGES = {95--135}
}

@article{murray2023randomized,
  title={Randomized numerical linear algebra: A perspective on the field with an eye to software},
  author={Murray, Riley and Demmel, James and Mahoney, Michael W and Erichson, N Benjamin and Melnichenko, Maksim and Malik, Osman Asif and Grigori, Laura and Luszczek, Piotr and Derezi{\'n}ski, Micha{\l} and Lopes, Miles E and others},
  journal={arXiv preprint arXiv:2302.11474},
  year={2023}
}

@article {Lepski1999,
    AUTHOR = {Lepski, O. and Nemirovski, A. and Spokoiny, V.},
     TITLE = {On estimation of the {$L_r$} norm of a regression function},
   JOURNAL = {Probab. Theory Related Fields},
  FJOURNAL = {Probability Theory and Related Fields},
    VOLUME = {113},
      YEAR = {1999},
    NUMBER = {2},
     PAGES = {221--253}
}

@incollection {Robins2008,
    AUTHOR = {Robins, James and Li, Lingling and Tchetgen, Eric and van der
              Vaart, Aad},
     TITLE = {Higher order influence functions and minimax estimation of
              nonlinear functionals},
 BOOKTITLE = {Probability and statistics: essays in honor of {D}avid {A}.
              {F}reedman},
    SERIES = {Inst. Math. Stat. (IMS) Collect.},
    VOLUME = {2},
     PAGES = {335--421},
 PUBLISHER = {Inst. Math. Statist., Beachwood, OH},
      YEAR = {2008}
}

@article {Donoho1991,
    AUTHOR = {Donoho, David L. and Liu, Richard C.},
     TITLE = {Geometrizing rates of convergence. {II}, {III}},
   JOURNAL = {Ann. Statist.},
  FJOURNAL = {The Annals of Statistics},
    VOLUME = {19},
      YEAR = {1991},
    NUMBER = {2},
     PAGES = {633--667, 668--701}
}

@article {Birge1995,
    AUTHOR = {Birg\'e, Lucien and Massart, Pascal},
     TITLE = {Estimation of integral functionals of a density},
   JOURNAL = {Ann. Statist.},
  FJOURNAL = {The Annals of Statistics},
    VOLUME = {23},
      YEAR = {1995},
    NUMBER = {1},
     PAGES = {11--29}
}

@article {Laurent1996,
    AUTHOR = {Laurent, B\'eatrice},
     TITLE = {Efficient estimation of integral functionals of a density},
   JOURNAL = {Ann. Statist.},
  FJOURNAL = {The Annals of Statistics},
    VOLUME = {24},
      YEAR = {1996},
    NUMBER = {2},
     PAGES = {659--681}
}

@inproceedings{derezinski2021sparse,
  title={Sparse sketches with small inversion bias},
  author={Derezinski, Michal and Liao, Zhenyu and Dobriban, Edgar and Mahoney, Michael},
  booktitle={Conference on Learning Theory},
  pages={1467--1510},
  year={2021},
  organization={PMLR}
}

@article {Derezinski2022,
    AUTHOR = {Derezi\'nski, Micha\l{} and Warmuth, Manfred K. and Hsu,
              Daniel},
     TITLE = {Unbiased estimators for random design regression},
   JOURNAL = {J. Mach. Learn. Res.},
  FJOURNAL = {Journal of Machine Learning Research (JMLR)},
    VOLUME = {23},
      YEAR = {2022},
     PAGES = {Paper No. [167], 46}
}

@article{derezinski2017unbiased,
  title={Unbiased estimates for linear regression via volume sampling},
  author={Derezinski, Michal and Warmuth, Manfred KK},
  journal={Advances in Neural Information Processing Systems},
  volume={30},
  year={2017}
}

@inproceedings{
niu2025fundamental,
title={Fundamental Bias in Inverting Random Sampling Matrices with Application to Sub-sampled Newton},
author={Chengmei Niu and Zhenyu Liao and Zenan Ling and Michael W. Mahoney},
booktitle={Forty-second International Conference on Machine Learning},
year={2025}
}

@article {Pilanci2016,
    AUTHOR = {Pilanci, Mert and Wainwright, Martin J.},
     TITLE = {Iterative {H}essian sketch: fast and accurate solution
              approximation for constrained least-squares},
   JOURNAL = {J. Mach. Learn. Res.},
  FJOURNAL = {Journal of Machine Learning Research (JMLR)},
    VOLUME = {17},
      YEAR = {2016},
     PAGES = {Paper No. 53, 38}
}

@inproceedings{wang2017sketching,
  title={Sketching meets random projection in the dual: A provable recovery algorithm for big and high-dimensional data},
  author={Wang, Jialei and Lee, Jason and Mahdavi, Mehrdad and Kolar, Mladen and Srebro, Nati},
  booktitle={Artificial Intelligence and Statistics},
  pages={1150--1158},
  year={2017},
  organization={PMLR}
}

@book {MR1102015,
    AUTHOR = {Ledoux, Michel and Talagrand, Michel},
     TITLE = {Probability in {B}anach spaces},
    SERIES = {Ergebnisse der Mathematik und ihrer Grenzgebiete (3) [Results
              in Mathematics and Related Areas (3)]},
    VOLUME = {23},
      NOTE = {Isoperimetry and processes},
 PUBLISHER = {Springer-Verlag, Berlin},
      YEAR = {1991},
     PAGES = {xii+480}
}

@article {MR3332804,
    AUTHOR = {Cai, T. Tony and Liang, Tengyuan and Zhou, Harrison H.},
     TITLE = {Law of log determinant of sample covariance matrix and optimal
              estimation of differential entropy for high-dimensional
              {G}aussian distributions},
   JOURNAL = {J. Multivariate Anal.},
  FJOURNAL = {Journal of Multivariate Analysis},
    VOLUME = {137},
      YEAR = {2015},
     PAGES = {161--172}
}

@article{ma2015statistical,
  title={A statistical perspective on algorithmic leveraging},
  author={Ma, Ping and Mahoney, Michael W and Yu, Bin},
  journal={The Journal of Machine Learning Research},
  volume={16},
  number={1},
  pages={861--911},
  year={2015},
  publisher={JMLR. org}
}

@article{drineas2012fast,
  title={Fast approximation of matrix coherence and statistical leverage},
  author={Drineas, Petros and Magdon-Ismail, Malik and Mahoney, Michael W and Woodruff, David P},
  journal={The Journal of Machine Learning Research},
  volume={13},
  number={1},
  pages={3475--3506},
  year={2012},
  publisher={JMLR. org}
}

@article {efron1979,
    AUTHOR = {Efron, B.},
     TITLE = {Bootstrap methods: another look at the jackknife},
   JOURNAL = {Ann. Statist.},
  FJOURNAL = {The Annals of Statistics},
    VOLUME = {7},
      YEAR = {1979},
    NUMBER = {1},
     PAGES = {1--26},
      ISSN = {0090-5364,2168-8966},
   MRCLASS = {62E15 (62G05 62H30 62J05)},
  MRNUMBER = {515681}
}

@article {Ahfock2021,
    AUTHOR = {Ahfock, D. C. and Astle, W. J. and Richardson, S.},
     TITLE = {Statistical properties of sketching algorithms},
   JOURNAL = {Biometrika},
  FJOURNAL = {Biometrika},
    VOLUME = {108},
      YEAR = {2021},
    NUMBER = {2},
     PAGES = {283--297},
      ISSN = {0006-3444,1464-3510},
   MRCLASS = {62-08 (60F05 62J05 62R07)}
}

@article{10.1561/2200000035,
author = {Mahoney, Michael W.},
title = {Randomized Algorithms for Matrices and Data},
year = {2011},
issue_date = {February 2011},
publisher = {Now Publishers Inc.},
address = {Hanover, MA, USA},
volume = {3},
number = {2},
journal = {Found. Trends Mach. Learn.},
month = feb,
pages = {123–224},
numpages = {102}
}

@book {Vershynin2018,
    AUTHOR = {Vershynin, Roman},
     TITLE = {High-dimensional probability},
    SERIES = {Cambridge Series in Statistical and Probabilistic Mathematics},
    VOLUME = {47},
      NOTE = {An introduction with applications in data science},
 PUBLISHER = {Cambridge University Press, Cambridge},
      YEAR = {2018},
     PAGES = {xiv+284},
      ISBN = {978-1-108-41519-4},
   MRCLASS = {60-01 (60B05 60B20 60E15 60Fxx 62H25)},
  MRNUMBER = {3837109},
MRREVIEWER = {Sasha\ Sodin}
}
